\documentclass[twoside]{article}
\usepackage{fullpage}
\usepackage{algorithm}
\usepackage{algorithmic}
\usepackage{microtype}
\usepackage{graphicx}
\usepackage{booktabs} % for professional tables
\usepackage{amsfonts}
\usepackage{amsmath}
\usepackage{amsthm}
\usepackage{hyperref}
\usepackage{natbib}
\usepackage{pgfplots}
\usepackage{pgfplotstable}
\usepackage{blindtext}
\usepackage{subcaption}
\usepgfplotslibrary{units}
\pgfplotsset{width=7cm,compat=1.13}
\pgfplotsset{major grid style={dotted,gray!50!black}}
\newcommand{\figurewidth}{0.9}
\newcommand{\figuregap}{-2.8em}
\newcommand{\figurepercentperrow}{0.288}
\pgfplotsset{every axis/.append style={
		label style={font=\tiny},
		tick label style={font=\tiny}  
}}

\newtheorem{theorem}{Theorem}
\newtheorem{lemma}{Lemma}
\newtheorem{c-approximate}{Definition}
\newtheorem{lsh}[c-approximate]{Definition}

\begin{document}

% If your paper is accepted and the title of your paper is very long,
% the style will print as headings an error message. Use the following
% command to supply a shorter title of your paper so that it can be
% used as headings.
%
%\runningtitle{I use this title instead because the last one was very long}

% If your paper is accepted and the number of authors is large, the
% style will print as headings an error message. Use the following
% command to supply a shorter version of the authors names so that
% they can be used as headings (for example, use only the surnames)
%
%\runningauthor{Surname 1, Surname 2, Surname 3, ...., Surname n}

\twocolumn[
\title{Norm-Range Partition: A Universal Catalyst for LSH based Maximum Inner Product Search (MIPS)}

\author{
	Xiao Yan,\ \ Xinyan Dai,\ \ Jie Liu,\ \ Kaiwen Zhou,\ \ James Cheng\\
	Department of Computer Science\\
	The Chinese University of Hong Kong\\
	Shatin, Hong Kong \\
	\texttt{\{xyan, xydai, jliu, kwzhou, jcheng\}@cse.cuhk.edu.hk} \\}
\date{\today}
\maketitle
]
%\twocolumn[

%\aistatstitle{Norm-range Partition: A Univiseral Catalyst for LSH based Maximum Inner Product Search (MIPS)}

%\aistatsauthor{ Author 1 \And Author 2 \And  Author 3 }
%
%\aistatsaddress{ Institution 1 \And  Institution 2 \And Institution 3 } 
%]
\begin{abstract}
	
  Recently, locality sensitive hashing (LSH) was shown to be effective for MIPS and several algorithms including $L_2$-ALSH, Sign-ALSH and Simple-LSH have been proposed. In this paper, we introduce the norm-range partition technique, which partitions the original dataset into sub-datasets containing items with similar 2-norms and builds hash index independently for each sub-dataset. We prove that norm-range partition reduces the query processing complexity for all existing LSH based MIPS algorithms under mild conditions. The key to performance improvement is that norm-range partition allows to use smaller normalization factor most sub-datasets. For efficient query processing, we also formulate a unified framework to rank the buckets from the hash indexes of different sub-datasets. Experiments on real datasets show that norm-range partition significantly reduces the number of probed for LSH based MIPS algorithms when achieving the same recall. 
         
\end{abstract}

\section{Introduction}
The problem of maximum inner product search (MIPS) is defined as follows~\citep{shrivastava:alsh}: given a dataset $\mathcal{S}\subset\mathbb{R}^d$ containing $n$ vectors (also called items) and a query $q\in\mathbb{R}^d$, find the item that has the maximum inner product with the query,
\begin{equation}\label{equ:mips}
	p=\arg \max_{x\in\mathcal{S}}{q^{\top}x}.
\end{equation}
Ties are broken randomly and the definition of MIPS can be trivially extended to top-$k$ inner product search. MIPS has many important applications such as recommendation based on user and item embeddings~\citep{koren:mf}, multi-class classification with linear classifier~\citep{dean:obj} and object matching in computer vision~\citep{felzens:obj}. Please refer to~\citep{shrivastava:alsh} for a more detailed discussion of the applications of MIPS. In these applications, it usually suffices to find approximate MIPS.

When the size of the dataset is large, naive linear scan often fails to meet the delay requirement of on-line query processing. Although there are several tree based methods~\citep{ram:cone, koenigstein:retrival} for MIPS, they suffer from the cruse of dimensionality and can perform even worse than linear scan with a moderate number of dimensions (e.g., 20). LSH based methods are appealing as they provide provably sub-linear query processing complexity for approximate nearest neighbor search (NNS) and their complexity does not depend on the number of dimension~\citep{indyk:lsh}. However, constructing an LSH for MIPS is generally considered challenging~\citep{koenigstein:retrival}. The main difficulty is that self-similarity is not the highest for inner product, which means an LSH for MIPS needs to satisfy $\mathbb{P}_{\mathcal{H}} \left[h(x)=h(y)\right]>\mathbb{P}_{\mathcal{H}} \left[h(x)=h(x)\right]=1$ when $y^{\top}x>x^{\top}x$.

%Although MIPS is equivalent to angular NNS or $L_2$ NNS when the 2-norm of the items are identical, existing LSHs for angular-NNS and $L_2$-NNS cannot be used directly for MIPS as there can be variations in the 2-norm of the items.  Euclidean distance    

In their seminal work~\citeyearpar{shrivastava:alsh}, Shrivastava and Li formulated the first LSH for MIPS ($L_2$-ALSH) utilizing the fact that the LSH framework does not require to use the same hash function for item and query. They apply a pair of asymmetric~\footnote{Asymmetric means different transformations/hash functions are used for the query and item. While symmetric means the same transformation is applied to both item and query.} transformations $P(x)$ and $Q(q)$ on item and query, and transform the problem of MIPS into Euclidean distance similarity search, which can be solved by existing LSH. Later, they improved $L_2$-ALSH with Sign-ALSH~\citep{shrivastava:signalsh}, which transforms MIPS into angular similarity search using another pair of asymmetric transformations. However, Neyshabur and Srebro showed that asymmetric transformations are not necessary~\citeyearpar{neyshabur:simple-lsh}. They proposed Simple-LSH, which uses a symmetric transformation to transform MIPS into angular similarity search. In both Sign-ALSH and Simple-LSH, the resulting angular similarity search problem is using an existing family of hash function called sign random projection. A more detailed introduction to these algorithms will be provided in Section~\ref{sec:LSH for MIPS}.    

The aforementioned LSH based MIPS algorithms follow a two-step procedure: first transform MIPS into angular/Euclidean similarity search, then use existing LSHs for angular/Euclidean similarity to solve the transformed problem. This observation opens up two directions for performance improvement, i.e., using better LSH functions and developing better transformations. Connecting the transformation of Simple-LSH with cross-polytope LSH~\citep{andoni:cross-lsh,terasawa:cross-lsh}, a state-of-the-art LSH family for angular similarity, we develop a new LSH based algorithm for MIPS called Cross-LSH. Cross-LSH outperforms all existing LSH based MIPS algorithms and the reason is that cross-polytope LSH solves angular similarity search more efficiently than sign random projection.

For better transformations, we propose the norm-range partition technique as a universal catalyst for LSH based MIPS algorithms. Norm-range partition divides the entire datasets into sub-datasets according to the percentiles of the 2-norm distribution of the items and builds hash index independently for each sub-dataset using an existing MIPS algorithm as sub-routine. The insight is that all existing algorithms need to normalize the items by the maximum Euclidean norm in the dataset and query processing complexity is an increasing function of the normalization constant. By dividing the entire dataset into sub-datasets, norm-range partition can use smaller normalization constant for most sub-datasets. We also prove that norm-range partition reduces the query processing complexity for all LSH based MIPS algorithms under mild conditions. To facilitate practical query processing, we also formulate a general framework to rank the buckets across the hash indexes of different sub-datasets.
     
We conduct experiments on real datasets and the results show that norm-range partition consistently improves the performance for all LSH based MIPS algorithms.

\textbf{Notations}: We use $\Vert . \Vert$ to denote the Euclidean norm of vectors. Euclidean norm is also called norm for conciseness. As the norm of the query does not affect the result of MIPS, we assume the query have unit norm, i.e., $\Vert q \Vert =1$ throughout the paper.        

\section{LSH based MIPS Algorithms}\label{sec:LSH for MIPS}
A widely used formalism of approximate near neighbor search is $c$-approximate near neighbor search, which is defined as follows~\footnote{The original $c$-approximate near neighbor search problem is define in terms of distance, we adopt the adaption of~\citep{shrivastava:alsh} in terms of similarity, which is more suitable for MIPS.}:

\begin{c-approximate}\label{def:c-approximate}
($c$-approximate near neighbor search or $c$-NN) Given a set $\mathcal{S}$ of items in d-dimensional space $\mathbb{R}^d$, and parameters $S>0$, $0<c<1$ and $\delta>0$, construct a data structure which, given a query $q\in\mathbb{R}^d$, does the following with probability $1-\delta$: if there exist an $S$-near neighbor of $q$ in $\mathcal{S}$, it returns some $cS$-near neighbor of $q$ in $\mathcal{S}$. 	
\end{c-approximate}

The definition only concerns $S>0$, which is not very restrictive as we are interested only in items having positive inner product with the query in most cases. Locality sensitive hashing (LSH)~\citep{indyk:lsh, andoni:lsh, datar:lsh} is a family of hash functions with the property that more similar items are hashed to the same value with higher probability. For a similarity function $sim$, if there exist an LSH family, $c$-NN can be conducted in with sub-linear complexity.         

\begin{lsh}\label{def:lsh}
	(Locality Sensitive Hashing) A family $\mathcal{H}$ is said to be a $(S, cS, p_1, p_2)$-LSH for similarity function $sim$ if, for any $x, y\in\mathbb{R}^d$, $h$ chosen uniformly random from $\mathcal{H}$ satisfies the following:
%\vspace{-4mm}
	\begin{itemize}
		\item if $sim(x,y)\geq S$, then $\mathbb{P}_{\mathcal{H}} \left[h(x)=h(y)\right]\ge p_1$,   
		\item if $sim(x,y)\leq cS$, then $\mathbb{P}_{\mathcal{H}} \left[h(x)=h(y)\right]\le p_2$.
	\end{itemize}  
\end{lsh}
%\vspace{-3mm}
For an LSH to be useful, it is required that $p_1>p_2$. Given a family of $(S, cS, p_1, p_2)$-LSH, one can construct a data structure for $c$-NN with $O(n^\rho \log n)$ query time, where $\rho=\frac{\log p_1}{\log p_2}$. We call $\rho$ the quality of an LSH and smaller $\rho$ means lower query processing complexity, thus higher quality. For a family of LSH, $\rho$ is a function of $S$ and $c$, we also call $(S,c)$ the condition number, which decides the difficulty of the $c$-NN problem.    

There exist well-known LSHs for Euclidean distance and angular similarity. For Euclidean distance, one LSH and its collision probability are given as:

\begin{equation}\label{equ:l2lsh}
h_{a,b}^{L_2}(x)=\left \lfloor \frac{a^{\top}x+b}{r} \right \rfloor
\end{equation}
\begin{equation}\label{equ:l2lsh_prob2}
F_r(d)=1-2\Phi(-\frac{r}{d})-\frac{2d}{\sqrt{2\pi}r}(1-e^ {-(r/d)^2/2})
\end{equation}

in which $a$ is a random vector whose entries follow i.i.d. standard normal distribution, $b$ is generated from a uniform distribution over $[0, r]$, $\Phi(x)$ is the cumulative density function of standard normal distribution and $d=\Vert x-y \Vert$ is the Euclidean distance between $x$ and $y$. For angular similarity, sign random projection (SRP) is an LSH:
\begin{equation}\label{equ:sign}
h_a(x)=\mathrm{sign} (a^{\top}x)
\end{equation}  
\begin{equation}\label{equ:sign_prob}
P\left[h_a(x)=h_a(y)\right]=1-\frac{1}{\pi} \cos^{-1} \left( \frac{x^{\top} y}{\Vert x \Vert \Vert y \Vert}\right)
\end{equation}
where the entries of $a$ follow i.i.d. standard normal distribution.     

\subsection{$L_2$-ASLH}
Shrivastava and Li~\citeyearpar{shrivastava:alsh} formulated the first LSH for MIPS by applying different transformations $P(x)$ and $Q(q)$ to the items and the query, respectively.
\begin{equation}\label{equ:asy}
\begin{aligned}
&P(x)=[\frac{Ux}{M};\Vert \frac{Ux}{M} \Vert^2;\Vert \frac{Ux}{M} \Vert^4;...;\Vert \frac{Ux}{M} \Vert^{2^m} ] \\
&Q(q)=[q;1/2;1/2;...;1/2] 
\end{aligned}
\end{equation}
in which $M=\max_{x\in\mathcal{S}}\Vert x \Vert$ is the maximum norm in the dataset and $0<U<1$ is a shrinking factor. After transformation, we have:
\begin{equation}\label{equ:res}
\Vert P(x)-Q(q) \Vert^2=1+\frac{m}{4}-2\frac{U}{M}x^{\top}q+\Vert \frac{Ux}{M} \Vert^{2^{m+1}}.
\end{equation}   
As $\Vert \frac{Ux}{M} \Vert<1$ and the $\Vert \frac{Ux}{M} \Vert^{2^{m+1}}$ term vanishes with $m$ at tower rate, the problem of finding the maximum inner product of $q$ is transformed into finding the nearest neighbor of $Q(q)$ in Euclidean distance, which can solved by the LSH in~\eqref{equ:l2lsh}. The quality $p$ of $L_2$-ASLH is given as:
\begin{equation}\label{equ:alsh_rho}
\rho=\frac{\log F_r\left(\sqrt{1+\frac{m}{4}-2\frac{US}{M}+(\frac{US}{M})^{2^{m+1}}}\right)}{\log F_r\left(\sqrt{1+\frac{m}{4}-2\frac{cUS}{M}}\right)}
\end{equation}  

\subsection{Sign-ALSH}
Shrivastava and Li~\citeyearpar{shrivastava:signalsh} found that SRP for angular similarity provides better $\rho$ than the Euclidean distance LSH in~\eqref{equ:l2lsh}. Therefore, they improved $L_2$-ASLH by transforming MIPS into angular similarity search with another pair of transformations: 
\begin{equation}\label{equ:sign_alsh}
\begin{aligned}
&P(x)=[\frac{Ux}{M};\frac{1}{2}-\Vert \frac{Ux}{M} \Vert^2;\frac{1}{2}-\Vert \frac{Ux}{M} \Vert^4;...;\frac{1}{2}-\Vert \frac{Ux}{M} \Vert^{2^m} ] \\
&Q(q)=[q;0;0;...;0] 
\end{aligned}
\end{equation}
in which the definition of $M$ and $U$ are similar to that of $L_2$-ASLH. After transformation, we have:
\begin{equation}\label{equ:Sign-ALSH}
\frac{Q(q)^{\top}P(x)}{\Vert Q(q) \Vert \Vert P(x) \Vert }=\frac{U}{M}\frac{q^{\top}x}{\sqrt{\frac{m}{4}+\Vert \frac{Ux}{M} \Vert^{2^{m+1}}}}
\end{equation}
As the $\Vert \frac{Ux}{M} \Vert^{2^{m+1}}$ term vanishes at tower rate with $m$, larger inner product leads to higher angular similarity. Please refer to (REF: Sign-ALSH) for the hash quality $\rho$ of Sign-ALSH. 

% If we assume the $\Vert \frac{Ux}{M} \Vert^{2^{m+1}}$ term can be ignored, the hash quality of sign-LSH can be given as~\footnote{For the case that the $\Vert \frac{Ux}{M} \Vert^{2^{m+1}}$ term cannot be ignored, please refer to xxx for more accurate expression}:
%\begin{equation}\label{equ:sign_rho}
%\rho=\frac{\log (1-\frac{\mathrm{cos}^{-1}(\frac{2US}{M\sqrt{m}})}{\pi})}{\log (1-\frac{\mathrm{cos}^{-1}(\frac{2cUS}{M\sqrt{m}})}{\pi})}
%\end{equation} 

\subsection{Simple-LSH}
%Neyshabur and Srebro proved that $L_2$-ALSH and sign-ALSH are not universal LSH for MIPS, that is, for any setting of $m$, $U$ (and $r$ ), there always exists a pair of $S$ and $c$ such that $x^{\top} q=S$ and $y^{\top} q=cS$ but $\mathbb{P}_{\mathcal{H}} \left[h(P(y))=h(Q(q))\right]>\mathbb{P}_{\mathcal{H}} \left[h(P(x))=h(Q(q))\right]$.
Neyshabur and Srebro~\citeyearpar{neyshabur:simple-lsh} argued that asymmetric transformations are not necessary if the items have bounded norm and the query has unit norm. Assuming the items are normalized by the maximum norm $M$ in the dataset, i.e., $x\mapsto\frac{x}{M}$, they proposed to use the same transformation $P(x)$ for both the items and the query.    
\begin{equation}\label{equ:simpleLSH}
P(x)=[x;\sqrt{1-\Vert x \Vert^2}]
\end{equation} 
After the transformation, we have
\begin{equation}
\frac{P(q)^{\top}P(x)}{\Vert P(q) \Vert \Vert P(x) \Vert}=\frac{q^{\top}x}{M}
\end{equation}
which shows that larger inner product leads to higher angular similarity. Simple-LSH uses SRP to solve the resulting angular similarity search problem and its quality $\rho$ is given as:
\begin{equation}\label{equ:simprho}
\rho=\frac{\log \left(1-\frac{\mathrm{cos}^{-1}(\frac{S}{M})}{\pi}\right)}{\log \left(1-\frac{\mathrm{cos}^{-1}(\frac{cS}{M})}{\pi}\right)}.
\end{equation}

We remark that existing LSH based MIPS algorithms can be viewed as a composition of two components: (1) a transformation from the original $(S, c)$-MIPS problem to a $(\tilde{S}, \tilde{c})$-Euclidean distance~\footnote{For $\tilde{c}$-NN in Euclidean distance, we have $\tilde{c}>1$, which means if there exists an item with distance $\tilde{S}$ from the query, the algorithm can return an item with larger distance.} or angular similarity search problem; (2) an existing LSH to solve the transformed problem. For example, Simple-LSH transforms the original $(S, c)$-MIPS problem into a $(\frac{S}{M}, c)$-angular similarity search problem and uses SRP for angular similarity search. Moreover, the two components are relatively independent, which means two directions are possible for performance improvement. Firstly, reusing existing transformations, we can switch to better LSH~\footnote{We say an LSH family $\mathcal{H}_1$ is better than another LSH family $\mathcal{H}_2$, if for every valid configuration of $(S, c)$, $\mathcal{H}_1$ provides smaller $\rho$, i.e., $\rho_{\mathcal{H}_1}(S, c)<\rho_{\mathcal{H}_2}(S, c)$.} for the resulting Euclidean/angular similarity search problem. Secondly, we can design better transformations~\footnote{Given an $(S, c)$-MIPS problem and an LSH family $\mathcal{H}$, we say a transformation $T_1$ is better than transformation $T_2$ if $\rho_{\mathcal{H}}(S_{T_1}, c_{T_1})<\rho_{\mathcal{H}}(S_{T_2}, c_{T_2})$, in which $(S_T, c_T)$ is the condition number of the original MIPS problem after applying transformation $T$.} while using the same LSH family as existing algorithm. In the subsequent sections, we show that both directions can be leveraged to devise better LSH based MIPS algorithms.

\section{Cross-LSH} 
In this section, we formulate an LSH based MIPS algorithm that outperforms existing ones by connecting the transformation of Simple-LSH in~\eqref{equ:simpleLSH} with Cross-polytope LSH, a more advanced LSH family for angular similarity.

\begin{figure} 
	\centering 
	\includegraphics[width=0.45\linewidth]{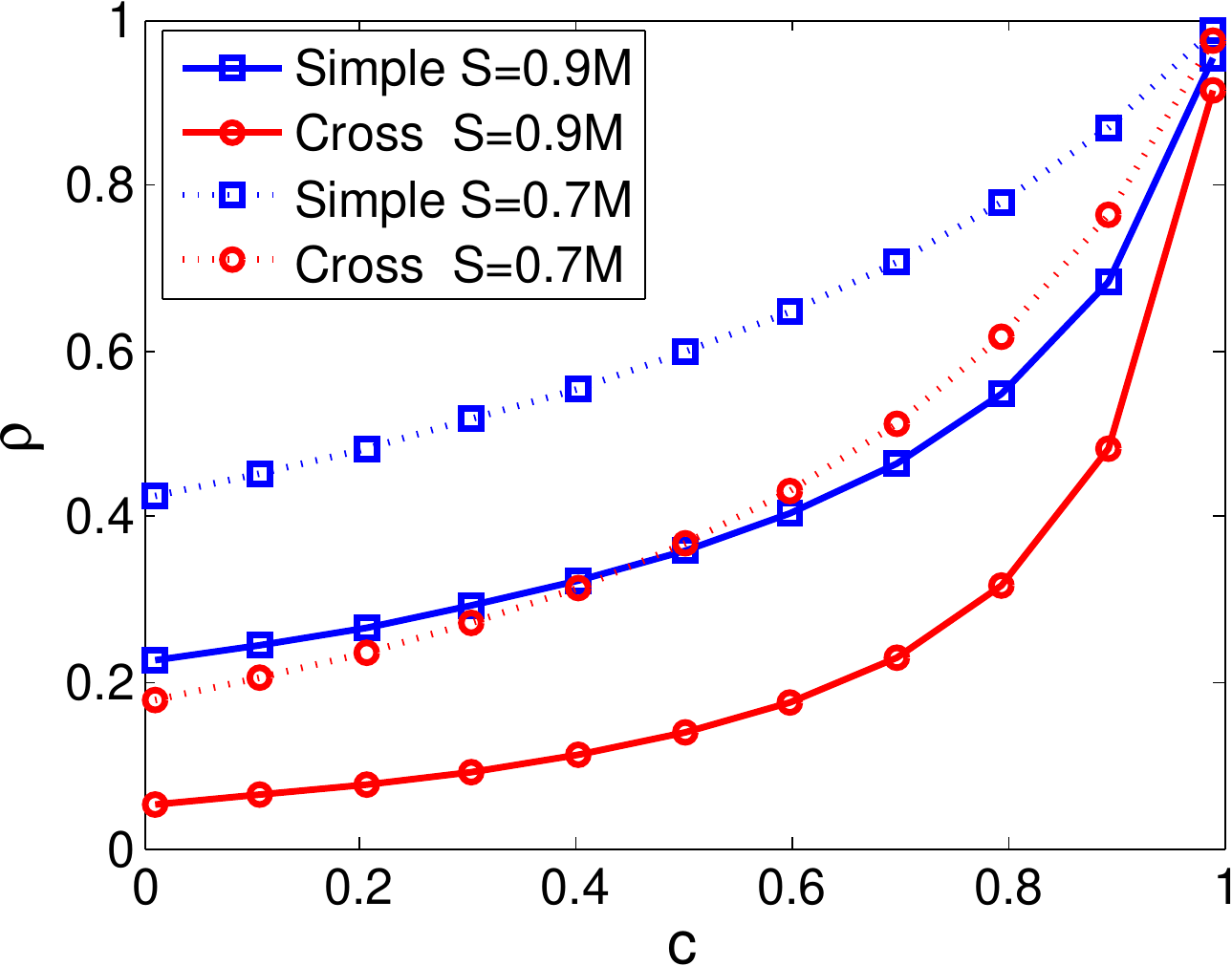} 
	\includegraphics[width=0.45\linewidth]{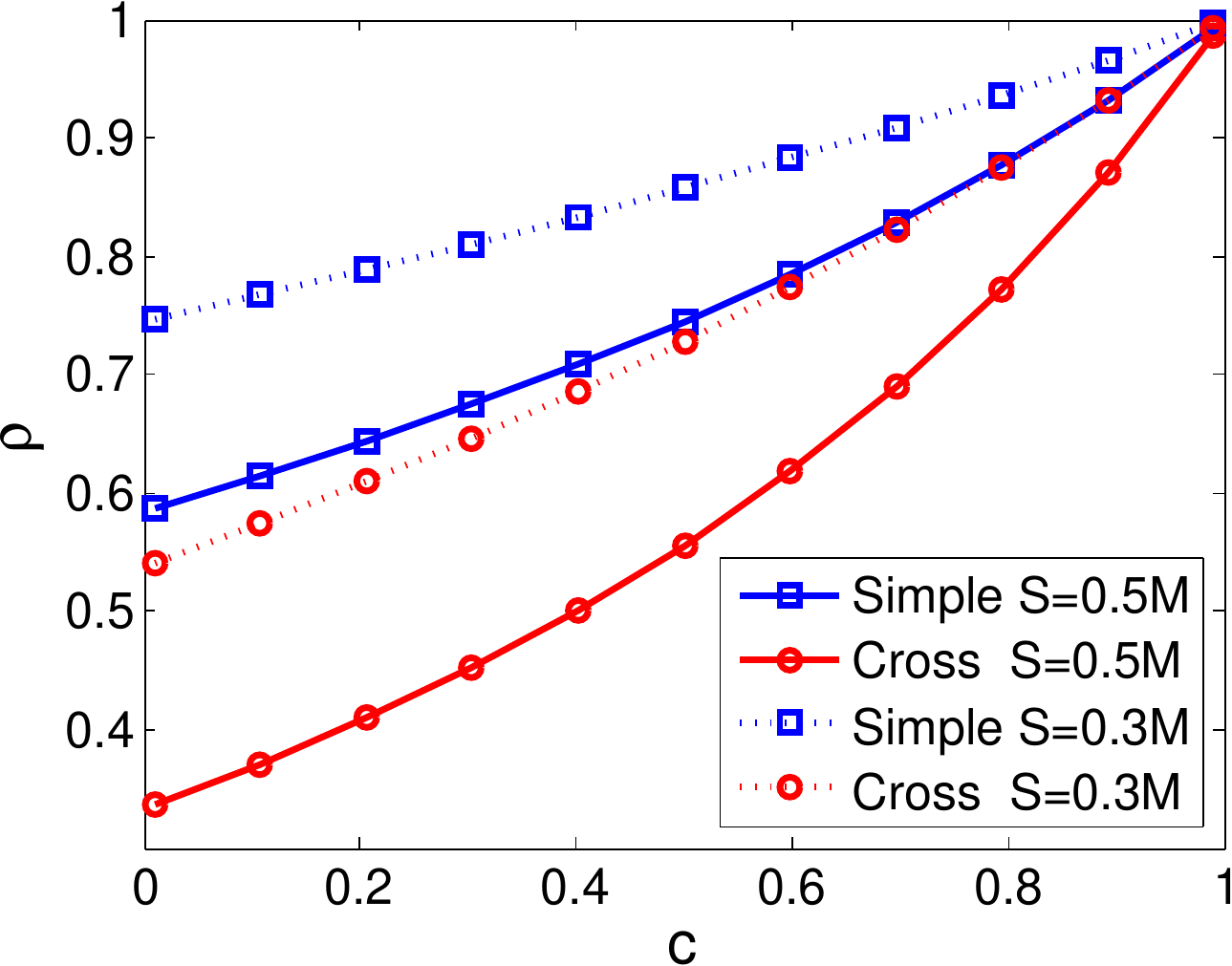}

	\caption{Hash quality comparison between Simple-LSH and Cross-LSH. } 
	\label{fig:rho comparision}
\end{figure}

Cross-polytope LSH targets Euclidean distance similarity search on the unit sphere, which is equivalent to angular similarity search. It is shown that cross-polytope LSH not only achieves the asymptotically optimal running time exponent theoretically but also significantly outperforms SRP in experiments~\citep{andoni:cross-lsh}. A hash function in the cross-polytope family is defined by a random matrix $A\in\mathbb{R}^{d'\times d}$ whose entires follow i.i.d. standard Gaussian distribution, and maps a vector $x$ on $d$-dimensional unit sphere to an alphabet of size $2d'$ using two steps: first calculate the normalized projection as $y=\frac{Ax}{\Vert Ax \Vert}$ and then find the closest point to $y$ in $\left\{\pm e_i \right\}_{ 1\le i\le d'}$, where $e_i$ is the $i$-th standard basis vector of $\mathbb{R}^{d'}$. For $(d,c)$-NN in Euclidean distance, the hash quality $\rho$ of Cross-polytope LSH is given as~\footnote{$\rho$ is given approximately as there are approximations in the derivation in~\citep{andoni:cross-lsh}.}:
%However, the approximations are shown to be tight.        
\begin{equation}\label{equ:cross quality}
\rho \approx \frac{1}{c^2}.\frac{4-c^2d^2}{4-d^2}.
\end{equation}       
Cross-polytope LSH combines naturally with the transformation in~\eqref{equ:simpleLSH} as it maps both query and item to the unit sphere. After the transformation, the Euclidean distance between item $x$ and query $q$ is:
\begin{equation}
\Vert P(x)-P(q) \Vert = \sqrt{2-2\frac{q^{\top}x}{M}},
\end{equation}
which shows that larger inner product results in smaller Euclidean distance. We can show that an $(S,c)$-MIPS is transformed into a $(d, c)$-Euclidean distance similarity search with parameters      
\begin{equation}\label{equ:cross parameter}
d = \sqrt{2-2S/M}; c=\sqrt{(M-cS)/(M-S)}.
\end{equation}  
Combine~\eqref{equ:cross parameter} with the hash quality of cross-ploytope LSH in~\eqref{equ:cross quality}, we can get the hash quality of Cross-LSH for $(S,c)$-MIPS as:
\begin{equation}\label{equ:cross rho}
\rho \approx \frac{(M+cS)(M-S)}{(M+S)(M-cS)}.
\end{equation}  
We plot the theoretical $\rho$ values of Simple-LSH and Cross-polytope LSH in Figure~\ref{fig:rho comparision}. We do not include $L_2$-ALSH and Sign-ALSH in the comparison as it has been shown that Simple-LSH achieves better hash quality than them. The results clearly show that Cross-LSH outperforms Simple-LSH theoretically. We will also show that Cross-LSH outperforms all existing LSH based MIPS algorithms in experiments in Section~\ref{sec:experiment}.   

\section{Norm-range Partition}\label{sec:norm-range}
In this section, we first introduce the norm-range partition technique which achieves better transformation by using smaller normalization factor for most sub-datasets. Then we discuss how to apply the norm-range partition technique efficiently for MIPS in practice.    
    
\subsection{The Norm-range Partition Technique}

The index building and query processing procedure of norm-rang partition are described in Algorithm~\ref{alg:index building} and Algorithm~\ref{alg:query}, respectively. 

\begin{algorithm}
	\caption{Norm-range Partition: Index Building}
	\label{alg:index building}
	\begin{algorithmic}[1]
		\STATE {\bfseries Input:} Dataset $\mathcal{S}$, size $n$, sub-dataset number $w$
		\STATE {\bfseries Output:} A hash index $\mathcal{I}_j$ for each sub-dataset
		\STATE Rank the items in $\mathcal{S}$ according to their norms
		\STATE Partition $\mathcal{S}$ into $w$ sub-datasets \{${\mathcal{S}_1,\mathcal{S}_2,...,\mathcal{S}_w}$\} such that $\mathcal{S}_j$ holds items whose norms ranked in the range $[\frac{(j-1)n}{w}, \frac{jn}{w}]$;
		\FOR{every sub-dataset $\mathcal{S}_j$}
		\STATE Use a meta algorithm to build index $\mathcal{I}_j$ for $\mathcal{S}_j$;
		\ENDFOR
	\end{algorithmic}
\end{algorithm}

\begin{algorithm}
	\caption{Norm-range Partition: Query Processing}
	\label{alg:query}
	\begin{algorithmic}[1]
		\STATE {\bfseries Input:} Hash indexes \{${\mathcal{I}_1,\mathcal{I}_2,...,\mathcal{I}_w}$\}, query $q$
		\STATE {\bfseries Output:} A $c$-approximate MIPS $x^{\star}$ to $q$
		\FOR{every hash index $\mathcal{I}_j$}
		\STATE Conduct MIPS with $q$ to get $x^{\star}_j$;
		\ENDFOR
		\STATE Select the item in \{$x^{\star}_1$, $x^{\star}_2$, ..., $x^{\star}_w$\} that has the maximum inner product with $q$ as the answer.
	\end{algorithmic}
\end{algorithm}

For index building, norm-range partition divides the dataset into sub-datasets according to percentiles in the norm distribution, which ensures that items in the same sub-dataset have similar norms. Then an arbitrary existing LSH based MIPS algorithm ($L_2$-ALSH, Sign-ALSH, Simple-LSH, Cross-LSH) is used as meta algorithm to build index for each sub-dataset independently. Note that for the ranking in the third line of Algorithm~\ref{alg:index building}, ties are broken randomly. If the norm distribution of the dataset is not very special~\footnote{For example, all items have the same norm.}, most sub-datasets will have a local maximum norm $M_j=\max_{x\in\mathcal{S}_j}\Vert x \Vert$ that is smaller than the global maximum norm $M=\max_{x\in\mathcal{S}}\Vert x \Vert$ in the entire dataset. Observe that all transformations in Section~\ref{sec:LSH for MIPS} involve a normalization process, i.e., scaling all items by the maximum norm in the dataset~\footnote{Normalization is implicit in the transformation of Simple-LSH in~\eqref{equ:simpleLSH} as it requires $\Vert x \Vert\le 1$ before applying the transformation.}. The normalization process is necessary to counter the problem that self-similarity is not the highest for inner product. However, by partitioning the dataset into sub-datasets, most sub-datasets can use smaller normalization factors $M_j<M$. We will show that the ability to reduce the normalization factors is the source of performance improvement.      

For query processing, norm-range partition conducts MIPS on the hash index of each sub-dataset independently and gets a local result $x^{\star}_j$ from sub-dataset $\mathcal{S}_j$. Then, the optimal one is selected from the local results as the final answer $x^{\star}$. We show that norm-range partition is a valid LSH for MIPS, which is stated in Theorem~\ref{theorem:corretness}.

\begin{theorem}\label{theorem:corretness}
	Given parameters $S>0$, $0<c<1$, if there exists an item having inner product $S$ with query $q$ in the dataset, norm-range partition returns an item with inner product at least $cS$ with probability $1-\delta$.   
\end{theorem}

\begin{proof}
If there exists an item with inner product $S$, it is contained in one of the sub-datasets after partition. Denote that sub-dataset as $\mathcal{S}_j$, query processing on its hash index $I_j$ is guaranteed to return some item having inner product $cS$ with probability $1-\delta$. This is because the meta algorithm used for index building in Algorithm~\ref{alg:index building} is a valid LSH for MIPS. For query processing in Algorithm~\ref{alg:query}, the final answer $x^{\star}$ is obtained by choosing the optimal one from the local answers generated by the sub-datasets. This ensures that $x^{\star}$ has an inner product at least $cS$ with probability $1-\delta$.         
\end{proof}

Now we analyze the query processing complexity of norm-range partition. Denote the hash quality $\rho$ of an LSH based MIPS algorithm on the entire dataset and sub-dataset $\mathcal{S}_j$ as $\rho$ and $\rho_j$, respectively. From Section~\ref{sec:LSH for MIPS}, we know that both $\rho$ and $\rho_j$ are functions of $S$, $c$ and the normalization factor. The following Lemma gives the relation between $\rho$ and $\rho_j$.  
 
\begin{lemma}\label{lemma:rho}
	For a sub-dataset $\mathcal{S}_j$, if it has $M_j<M$, where $M_j=\max_{x\in\mathcal{S}_j}\Vert x \Vert$ and $M=\max_{x\in\mathcal{S}}\Vert x \Vert$, then $\rho_j<\rho$.    
\end{lemma}

We provide the detailed proof of Lemma~\ref{lemma:rho} in the supplementary material. For Simple-LSH and Cross-LSH, we have $\rho'(M)>0$, which means larger $M$ results in higher query processing complexity. For $L_2$-ALSH and Sign-ALSH, we prove that Lemma~\ref{lemma:rho} holds when assuming the $\Vert \frac{Ux}{M} \Vert^{2^{m+1}}$ term can be ignored. This assumption is not very restrictively as $L_2$-ALSH and Sign-ALSH also need it to be valid LSH for MIPS. Moreover, $U$ and $m$ are usually chosen to make the term very small in practice. We also provide plots of the theoretically in the supplementary material $\rho$ of $L_2$-ALSH and Sign-ALSH without ignoring the $\Vert \frac{Ux}{M} \Vert^{2^{m+1}}$ term under various setting of $(S,c)$, which show that $\rho$ is an increasing function of $M$.

The result of Lemma~\ref{lemma:rho} can also be interrupted intuitively. Observe that the transformations in Section~\ref{sec:LSH for MIPS} all introduce dummy terms, such as the $\sqrt{1-\Vert x \Vert^2}$ the term in Simple-LSH and the $\frac{1}{2}-\Vert \frac{Ux}{M} \Vert^2$ term in Sign-ALSH. These terms do not affect inner product but are necessary to make the algorithms valid LSH for MIPS. If we process the entire dataset as a whole, for a large number of items whose norm is much smaller than the maximum norm $M$~\footnote{We provide the norm distributions of same real datasets in the supplementary material as examples.}, these terms will be large and have significant impact on the result of hashing. This harms the effectiveness of hashing in discriminating similar items from dissimilar items. By ensuring items in the same sub-dataset have similar norm and reducing the normalization factor, norm-range partition effectively reduces the magnitude of the dummy terms, thus reduces the value of $\rho$.

Denote $\rho^{\star}=\max_{\rho_j<\rho} \rho_j$, which is the maximum $\rho$ value for sub-datasets with $\rho_j<\rho$. We analyze the query processing complexity of norm-range partition in Theorem~\ref{theorem:complexity}.

\begin{theorem}\label{theorem:complexity}
	 Norm-range partition attains lower query processing complexity than the meta algorithm with sufficiently large $n$, if the dataset is divided into $n^{\alpha}$ sub-datasets and there are at most $n^{\beta}$ sub-datasets with $\rho_j=\rho$, where $0<\alpha<\mathrm{min}\{\rho, \frac{\rho-\rho^{\star}}{1-\rho^{\star}}\}$ and $0<\beta<\alpha \rho$.   
\end{theorem}

\begin{proof}
The query processing complexity of norm-range partitioning can be expressed as a function of $n$ as:
\begin{equation}
\begin{aligned}\label{equ:complex}
f(n&)\!=\!n^{\alpha}+\sum_{j=1}^{n^{\alpha}} n^{(1-\alpha)\rho_j} \log n^{1-\alpha}\\ 
&\!<\!n^{\alpha}+\sum_{j=1}^{n^{\alpha}} n^{(1-\alpha)\rho_j} \log n\\
&\!\le\!n^{\alpha}\!+\!\sum_{j=1}^{\!\!n^{\alpha}-n^{\beta}}n^{(1-\alpha)\rho_j} \log n\!+\!n^{\beta}n^{(1-\alpha)\rho} \log n\\
&\!<\!n^{\alpha}+n^{\alpha}n^{(1-\alpha)\rho^{\star}} \log n+n^{\beta}n^{(1-\alpha)\rho} \log n
\end{aligned}	
\end{equation}    
In the first line, $n^{\alpha}$ is the complexity of choosing the optimal answer from the local results from $n^{\alpha}$ sub-datasets while the second term is the complexity of MIPS on the sub-datasets. Due to percentile based partition, all sub-datasets have the same size $n^{1-\alpha}$ and $\rho_j$ is the hash quality on sub-dataset $\mathcal{S}_j$. The "$\le$" in the third line is because there are at most $n^{\beta}$ sub-datasets with $\rho_j=\rho$, for the remaining $n^{\alpha}-n^{\beta}$ sub-datasets, their have $\rho_j\le\rho^{\star}<\rho$. Recall that if we use the meta to process the dataset as whole, the query complexity is $O(n^\rho \log n)$. We can compare the complexity of norm-range partition and the meta algorithm as:
\begin{equation}
\begin{aligned}\label{eqa:res}
\frac{f(n)}{n^\rho \log n}&<\frac{n^{\alpha}\!+\!\big(n^{\alpha}n^{(1-\alpha)\rho^{\star}}\!+\!n^{\beta}n^{(1-\alpha)\rho}\big) \log n}{n^\rho \log n}\\
&=n^{\alpha-\rho}/\log n+n^{\alpha+(1-\alpha)\rho^{\star}-\rho}+n^{\beta-\alpha \rho}
\end{aligned}	
\end{equation}        
~(\ref{eqa:res}) tends to 0 with sufficiently large $n$ when $\alpha\le \rho$, $\alpha+(1-\alpha)\rho^{\star}<\rho$ and $\beta-\alpha \rho<0$, which is satisfied by $\alpha<\mathrm{min}\{\rho,  \frac{\rho-\rho^{\star}}{1-\rho^{\star}}\}$ and $\beta<\alpha \rho$.  
\end{proof} 

We would like to comment that Theorem~\ref{theorem:complexity} holds under mild conditions. Firstly, it requires that there are at most $n^{\beta}$ sub-datasets with $\rho_j=\rho$. Combining with Lemma~\ref{lemma:rho}, this means the number of sub-datasets with $M_j=M$ is no more than $n^{\beta}$. For practical datasets, usually only the sub-dataset containing the items with the largest norm has $M_j=M$~\footnote{See examples in the supplementary material}. Moreover, Theorem~\ref{theorem:complexity} imposes an upper bound $n^{\alpha}$ for the number of sub-datasets, which is more favorable than a lower bound. This is because we need to build a hash index for each sub-dataset and a large number of hash indexes will be costly to manage. In the experiments, we will show that using a small number of sub-datasets already provides significant performance improvement. 

When all items have the same norm, norm-range partition has higher complexity than the original meta algorithm due to the additional selection process. However, MIPS is equivalent to angular/Euclidean similarity search in this case, thus an LSH for MIPS is not needed. In the worst case, one can decide not to use norm-range partition if a portion of items have $\Vert x \Vert=M$.
 
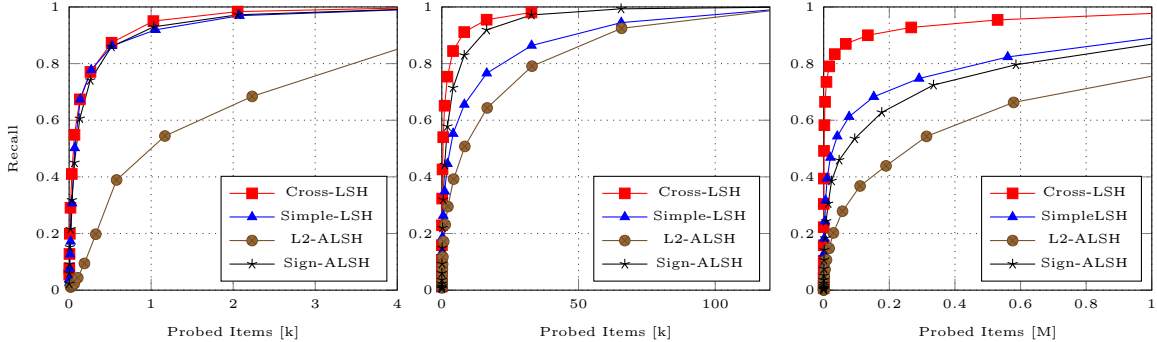
\begin{figure*}[t]
	\centering
		\begin{subfigure}[b]{0.36\textwidth}
			\begin{tikzpicture}
			\begin{axis}[
			height=\figurewidth\linewidth,
			width=\linewidth,
			legend style={font=\fontsize{1}{1}\selectfont},
			grid=major,
			legend pos=south east,
			change x base,
			x SI prefix=kilo,x unit=\-,
			xlabel=Probed Items , ylabel=Recall,
			xmin=0,xmax=4000,
			ymin=0,ymax=1,
			]
					
			\addplot [mark=square*, red]
			coordinates 
			{
				(1.491, 0.0546997)
				(2.295, 0.0765496)
				(4.401, 0.12775)
				(8.493, 0.19905)
				(16.584, 0.2909)
				(32.474, 0.4102)
				(65.11, 0.5486)
				(131.029, 0.673751)
				(260.075, 0.770651)
				(517.461, 0.873702)
				(1027.45, 0.950102)
				(2054.4, 0.983601)
				(4100.19, 0.99655)
				(8193.16, 1.0)
				(16384.3, 1.0)
				(17770.0, 1.0)
			};
			\addlegendentry{Cross-LSH}		
					
			\addplot [mark=triangle*, blue]
			coordinates 
			{
				(2.062, 0.02675)
				(3.755, 0.0396)
				(7.063, 0.07585)
				(11.817, 0.1268)
				(18.355, 0.17215)
				(35.632, 0.3069)
				(68.228, 0.50205)
				(136.115, 0.6739)
				(269.483, 0.777751)
				(521.741, 0.862701)
				(1053.72, 0.919252)
				(2077.03, 0.969201)
				(4155.26, 0.9914)
				(8270.75, 0.99885)
				(16401.9, 1.0)
				(17770.0, 1.0)
			};
			\addlegendentry{Simple-LSH}
			\addplot 
			coordinates 
			{
				(17.84, 0.01095)
				(19.231, 0.0123)
				(32.164, 0.0134)
				(35.82, 0.0178)
				(47.131, 0.02)
				(63.118, 0.0229)
				(105.525, 0.04365)
				(186.873, 0.09435)
				(324.907, 0.19755)
				(578.709, 0.3891)
				(1167.72, 0.54435)
				(2231.86, 0.68415)
				(4211.54, 0.870502)
				(8281.41, 0.972601)
				(16393.4, 1.0)
				(17770.0, 1.0)
			};grid=both
			\addlegendentry{L2-ALSH}
			\addplot 
			coordinates 
			{
				(1.272, 0.0231)
				(2.675, 0.0592999)
				(4.789, 0.09125)
				(9.035, 0.1517)
				(17.344, 0.21555)
				(33.775, 0.3177)
				(65.581, 0.4502)
				(129.918, 0.6059)
				(258.183, 0.7411)
				(515.778, 0.859601)
				(1030.96, 0.929801)
				(2057.75, 0.972351)
				(4109.02, 0.99205)
				(8210.37, 0.99875)
				(16405.4, 1.0)
				(17770.0, 1.0)
			};
			\addlegendentry{Sign-ALSH}
			\end{axis}
			\end{tikzpicture}
	\end{subfigure}%
	\hspace{\figuregap}	
	\begin{subfigure}[b]{0.36\textwidth}
			\begin{tikzpicture}
			\begin{axis}[
			height=\figurewidth\linewidth,
			width=\linewidth,
			legend style={font=\fontsize{1}{1}\selectfont},
			grid=major,
			legend pos=south east,
			change x base,
			x SI prefix=kilo,x unit=\-,
			xlabel=Probed Items ,
			xmin=0,xmax=120000,
			ymin=0,ymax=1,
			]
			
			\addplot [mark=square*, red]
			 coordinates
			{
				(1.293, 0.0126)
				(2.462, 0.02235)
				(4.737, 0.0390499)
				(9.107, 0.0655999)
				(17.636, 0.10715)
				(34.098, 0.1586)
				(66.401, 0.2291)
				(131.114, 0.32335)
				(260.59, 0.42625)
				(517.042, 0.5403)
				(1029.51, 0.651251)
				(2054.43, 0.753801)
				(4102.73, 0.844351)
				(8200.81, 0.911351)
				(16396.9, 0.955001)
				(32784.3, 0.98065)
			};
			\addlegendentry{Cross-LSH}
	
			\addplot [mark=triangle*, blue]
			coordinates 
			{
				(2.184, 0.0054)
				(3.985, 0.00895001)
				(7.288, 0.0173)
				(13.866, 0.0263)
				(23.652, 0.0403499)
				(41.99, 0.0613499)
				(78.428, 0.09465)
				(147.699, 0.1352)
				(276.781, 0.18755)
				(536.383, 0.2632)
				(1055.89, 0.3489)
				(2083.34, 0.446149)
				(4135.04, 0.5527)
				(8244.75, 0.655551)
				(16436.3, 0.765901)
				(32835.4, 0.863851)
				(65624.6, 0.944701)
				(131137.0, 0.9985)
				(136736.0, 1.0)
			};
			\addlegendentry{Simple-LSH}
			\addplot coordinates 
			{
				(37.414, 0.00795001)
				(44.937, 0.01215)
				(67.696, 0.0163)
				(85.9, 0.02065)
				(116.934, 0.0342)
				(152.909, 0.0571999)
				(206.332, 0.075)
				(282.601, 0.0920001)
				(404.0, 0.1173)
				(688.146, 0.17155)
				(1248.76, 0.2312)
				(2235.33, 0.2951)
				(4323.25, 0.39205)
				(8413.33, 0.50755)
				(16592.9, 0.6435)
				(32984.1, 0.790901)
				(65758.1, 0.924651)
				(131152.0, 0.9991)
				(136736.0, 1.0)
			};
			\addlegendentry{L2-ALSH}
			\addplot coordinates 
			{
				(1.833, 0.00545)
				(3.359, 0.00825001)
				(5.978, 0.0134)
				(11.077, 0.02315)
				(19.936, 0.0358999)
				(37.063, 0.0600999)
				(71.018, 0.0928001)
				(135.428, 0.14905)
				(264.425, 0.2198)
				(521.718, 0.3195)
				(1034.47, 0.4423)
				(2061.33, 0.57815)
				(4111.58, 0.71435)
				(8210.93, 0.82935)
				(16405.2, 0.919001)
				(32791.7, 0.971801)
				(65569.0, 0.9934)
				(131124.0, 1.0)
				(136736.0, 1.0)
			};
			\addlegendentry{Sign-ALSH}

			\end{axis}
			\end{tikzpicture}
	\end{subfigure}%
	\hspace{\figuregap}	
	\begin{subfigure}[b]{0.36\textwidth}
			\begin{tikzpicture}
			\begin{axis}[
			height=\figurewidth\linewidth,
			width=\linewidth,
			legend style={font=\fontsize{1}{1}\selectfont},
			grid=major,
			legend pos=south east,
			change x base,
			x SI prefix=mega,x unit=\-,
			xlabel=Probed Items ,
			xmin=0,xmax=1000000,
			ymin=0,ymax=1,
			]
			
			\addplot [mark=square*, red] 
			coordinates 
			{
				(1.069, 0.00670001)
				(2.092, 0.01415)
				(4.144, 0.02495)
				(8.193, 0.0420499)
				(16.312, 0.0662999)
				(32.528, 0.10365)
				(64.844, 0.1584)
				(130.554, 0.2224)
				(259.682, 0.30435)
				(523.785, 0.394)
				(1052.96, 0.4917)
				(2109.28, 0.58265)
				(4163.57, 0.6649)
				(8477.82, 0.7353)
				(16998.2, 0.790251)
				(33838.5, 0.833451)
				(67389.9, 0.869601)
				(134601.0, 0.900101)
				(266778.0, 0.9277)
				(530166.0, 0.954251)
				(1052820.0, 0.97975)
			};
			\addlegendentry{Cross-LSH}
			
			\addplot [mark=triangle*, blue]
			coordinates 
			{
				(1.394, 0.00115)
				(33.772, 0.0027)
				(40.304, 0.0045)
				(45.257, 0.00775001)
				(61.221, 0.01295)
				(185.786, 0.0211)
				(247.885, 0.03455)
				(344.632, 0.0543499)
				(580.825, 0.0864501)
				(1051.26, 0.12945)
				(1879.94, 0.1829)
				(3362.95, 0.2433)
				(5743.07, 0.3167)
				(10573.1, 0.39455)
				(20450.1, 0.46835)
				(41469.9, 0.5435)
				(77510.7, 0.6125)
				(152780.0, 0.682951)
				(290793.0, 0.747351)
				(560941.0, 0.823402)
				(1091640.0, 0.904151)
				(2114840.0, 0.988701)
				(2340370.0, 1.0)
			};
			\addlegendentry{SimpleLSH}
			\addplot coordinates 
			{
				(2.886, 0.00045)
				(4.558, 0.0006)
				(11.566, 0.0009)
				(22.402, 0.00195)
				(38.274, 0.0033)
				(75.927, 0.0064)
				(163.772, 0.0102)
				(275.612, 0.0158)
				(517.158, 0.02465)
				(1010.7, 0.03455)
				(1906.82, 0.0475999)
				(4161.62, 0.0730499)
				(8559.89, 0.10695)
				(16703.5, 0.14765)
				(29667.6, 0.202)
				(57484.8, 0.2787)
				(110982.0, 0.3678)
				(190015.0, 0.4388)
				(313390.0, 0.5429)
				(578900.0, 0.6629)
				(1119210.0, 0.7819)
				(2111280.0, 0.955701)
				(2340370.0, 1.0)
			};
			\addlegendentry{L2-ALSH}
			\addplot coordinates 
			{
				(16.57, 0.0019)
				(43.869, 0.00355)
				(66.626, 0.0068)
				(83.99, 0.0092)
				(105.048, 0.0146)
				(130.412, 0.0233)
				(196.002, 0.03575)
				(288.185, 0.0512999)
				(629.776, 0.07575)
				(987.22, 0.1065)
				(1752.06, 0.14195)
				(4442.29, 0.18325)
				(7520.13, 0.2435)
				(13726.0, 0.30555)
				(24169.5, 0.38595)
				(47143.5, 0.4592)
				(94542.7, 0.5349)
				(177046.0, 0.627701)
				(334760.0, 0.724351)
				(585994.0, 0.795951)
				(1108880.0, 0.887601)
				(2131050.0, 0.990451)
				(2340370.0, 1.0)
			};
			\addlegendentry{Sign-ALSH}

			\end{axis}
			\end{tikzpicture}	
	\end{subfigure}%
	\caption{Probed item-recall comparison between Cross-LSH and existing algorithms under a code length of 32 (best viewed in color). From left to right, the datasets are Netflix, Yahoo!Music and ImageNet, respectively. } 
	\label{fig:cross-LSH vs others}
\end{figure*}  

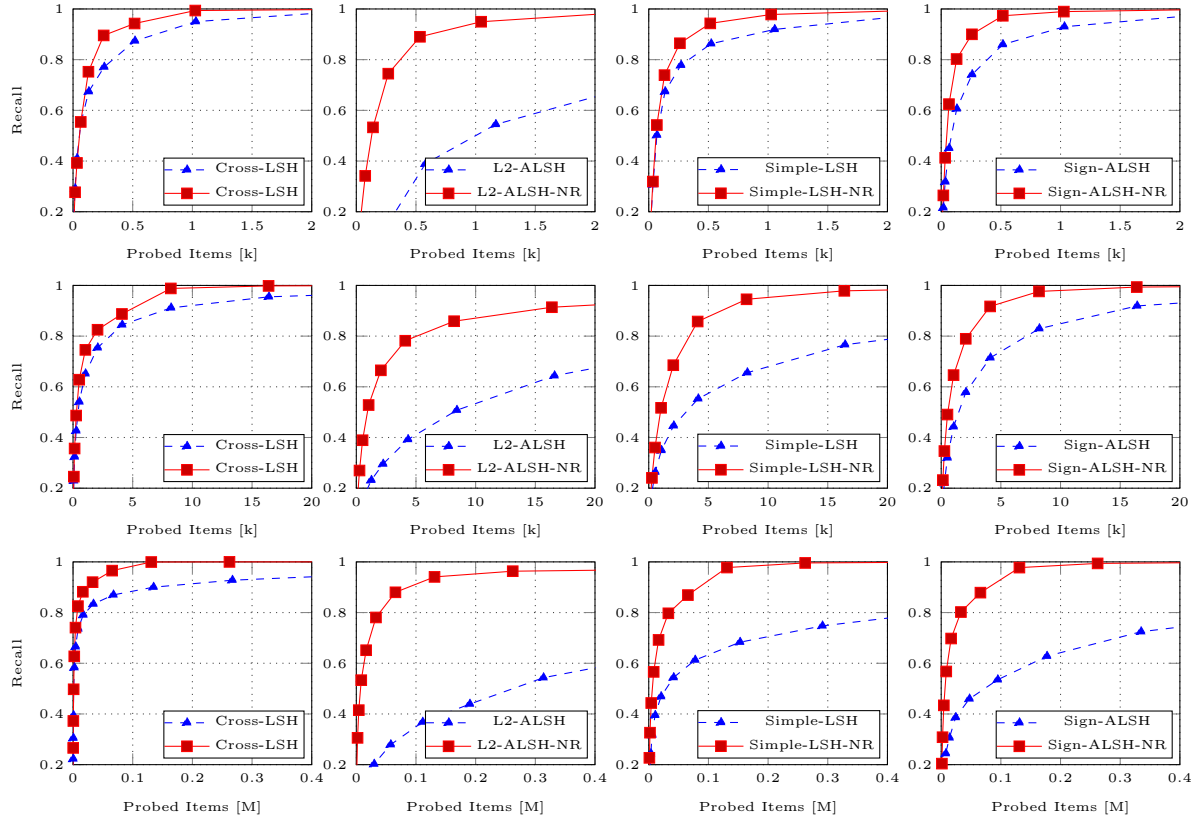
\begin{figure*}[t]
	\centering
		\begin{subfigure}[b]{\figurepercentperrow\textwidth}
			\begin{tikzpicture}
			\begin{axis}[
			height=\figurewidth\linewidth,
			width=\linewidth,
			grid=major,
			legend pos=south east,
			legend style={inner xsep=0pt, inner ysep=0pt, font=\fontsize{1}{1}\selectfont},
			change x base,
			x SI prefix=kilo,x unit=\-,
			xlabel=Probed Items , ylabel=Recall,
			xmin=0,xmax=2000,
			ymin=0.2,ymax=1,
			mark size=2.0pt,
			]
					\addplot  [mark=triangle*,dashed,blue] 
					 coordinates 
					{
						(1.491, 0.0546997)
						(2.295, 0.0765496)
						(4.401, 0.12775)
						(8.493, 0.19905)
						(16.584, 0.2909)
						(32.474, 0.4102)
						(65.11, 0.5486)
						(131.029, 0.673751)
						(260.075, 0.770651)
						(517.461, 0.873702)
						(1027.45, 0.950102)
						(2054.4, 0.983601)
						(4100.19, 0.99655)
						(8193.16, 1.0)
						(16384.3, 1.0)
						(17770.0, 1.0)
					};
					\addlegendentry{Cross-LSH}
					\addplot  
					coordinates 
					{
						(1.428, 0.0464498)
						(2.345, 0.0687497)
						(4.725, 0.1184)
						(8.879, 0.18245)
						(17.105, 0.27685)
						(32.831, 0.3925)
						(64.774, 0.5543)
						(128.937, 0.752301)
						(256.922, 0.895602)
						(515.827, 0.942651)
						(1024.25, 0.9938)
						(2048.26, 0.99735)
						(4096.12, 0.99865)
						(8192.06, 0.99975)
						(16384.0, 1.0)
						(17770.0, 1.0)
					};
					\addlegendentry{Cross-LSH}
					
			\end{axis}
			\end{tikzpicture}
	\end{subfigure}%
	\hspace{\figuregap}
	\begin{subfigure}[b]{\figurepercentperrow\textwidth}
	\begin{tikzpicture}
	\begin{axis}[
	height=\figurewidth\linewidth,
	width=\linewidth,
	legend pos=south east,
	legend style={inner xsep=0pt, inner ysep=0pt, font=\fontsize{1}{1}\selectfont},
	grid=major,
	legend pos=south east,
	change x base,
	x SI prefix=kilo,x unit=\-,
	xlabel=Probed Items ,
	xmin=0,xmax=2000,
	ymin=0.2,ymax=1,
	]
	\addplot  [mark=triangle*,dashed,blue] 
	coordinates 
	{
		(17.84, 0.01095)
		(19.231, 0.0123)
		(32.164, 0.0134)
		(35.82, 0.0178)
		(47.131, 0.02)
		(63.118, 0.0229)
		(105.525, 0.04365)
		(186.873, 0.09435)
		(324.907, 0.19755)
		(578.709, 0.3891)
		(1167.72, 0.54435)
		(2231.86, 0.68415)
		(4211.54, 0.870502)
		(8281.41, 0.972601)
		(16393.4, 1.0)
		(17770.0, 1.0)
	};
	\addlegendentry{L2-ALSH}
	\addplot coordinates 
	{
		(9.596, 0.06005)
		(11.478, 0.0694)
		(13.391, 0.07525)
		(17.439, 0.0986501)
		(25.315, 0.1284)
		(41.422, 0.1979)
		(74.713, 0.3414)
		(139.739, 0.53285)
		(267.869, 0.744701)
		(533.35, 0.890251)
		(1046.19, 0.949601)
		(2071.32, 0.980951)
		(4113.52, 0.9943)
		(8201.93, 0.9989)
		(16386.9, 1.0)
		(17770.0, 1.0)
	};
	\addlegendentry{L2-ALSH-NR}
	\end{axis}
	\end{tikzpicture}
	\end{subfigure}%
	\hspace{\figuregap}	
	\begin{subfigure}[b]{\figurepercentperrow\textwidth}
		\begin{tikzpicture}
		\begin{axis}[
		height=\figurewidth\linewidth,
		width=\linewidth,
		legend pos=south east,
		legend style={inner xsep=0pt, inner ysep=0pt, font=\fontsize{1}{1}\selectfont},
		grid=major,
		legend pos=south east,
		change x base,
		x SI prefix=kilo,x unit=\-,
		xlabel=Probed Items ,
		xmin=0,xmax=2000,
		ymin=0.2,ymax=1,
		]
		\addplot [mark=triangle*,dashed,blue] 
		coordinates 
		{
			(2.062, 0.02675)
			(3.755, 0.0396)
			(7.063, 0.07585)
			(11.817, 0.1268)
			(18.355, 0.17215)
			(35.632, 0.3069)
			(68.228, 0.50205)
			(136.115, 0.6739)
			(269.483, 0.777751)
			(521.741, 0.862701)
			(1053.72, 0.919252)
			(2077.03, 0.969201)
			(4155.26, 0.9914)
			(8270.75, 0.99885)
			(16401.9, 1.0)
			(17770.0, 1.0)
		};
		\addlegendentry{Simple-LSH}
		\addplot
		coordinates 
		{
			(1.979, 0.0121)
			(3.81, 0.03865)
			(6.21, 0.0616999)
			(10.896, 0.1052)
			(18.177, 0.17515)
			(34.861, 0.3191)
			(67.833, 0.54205)
			(132.048, 0.7389)
			(261.093, 0.864701)
			(515.435, 0.943101)
			(1025.89, 0.978201)
			(2050.5, 0.99205)
			(4096.88, 0.99945)
			(8192.73, 0.99995)
			(16384.2, 1.0)
			(17770.0, 1.0)
		};
		\addlegendentry{Simple-LSH-NR}
		\end{axis}
		\end{tikzpicture}
	\end{subfigure}%
	\hspace{\figuregap}
	\begin{subfigure}[b]{\figurepercentperrow\textwidth}
	\begin{tikzpicture}
	\begin{axis}[
	height=\figurewidth\linewidth,
	width=\linewidth,
	legend pos=south east,
	legend style={inner xsep=0pt, inner ysep=0pt, font=\fontsize{1}{1}\selectfont},
	grid=major,
	change x base,
	x SI prefix=kilo,x unit=\-,
	xlabel=Probed Items,
	xmin=0,xmax=2000,
	ymin=0.2,ymax=1,
	]
	\addplot  [mark=triangle*,dashed,blue] 
	coordinates 
	{
		(1.272, 0.0231)
		(2.675, 0.0592999)
		(4.789, 0.09125)
		(9.035, 0.1517)
		(17.344, 0.21555)
		(33.775, 0.3177)
		(65.581, 0.4502)
		(129.918, 0.6059)
		(258.183, 0.7411)
		(515.778, 0.859601)
		(1030.96, 0.929801)
		(2057.75, 0.972351)
		(4109.02, 0.99205)
		(8210.37, 0.99875)
		(16405.4, 1.0)
		(17770.0, 1.0)
	};
	\addlegendentry{Sign-ALSH}
	\addplot coordinates 
	{
		(2.31, 0.0445498)
		(4.967, 0.1014)
		(7.144, 0.135)
		(9.85, 0.16645)
		(18.039, 0.2644)
		(33.438, 0.4128)
		(65.256, 0.62455)
		(129.209, 0.802452)
		(257.659, 0.900052)
		(516.132, 0.973151)
		(1027.52, 0.989601)
		(2051.08, 0.9966)
		(4097.87, 0.99885)
		(8193.42, 0.99975)
		(16385.3, 1.0)
		(17770.0, 1.0)
	};
	\addlegendentry{Sign-ALSH-NR}
	
	\end{axis}
	\end{tikzpicture}
	\end{subfigure}%	
	
	
		\begin{subfigure}[b]{\figurepercentperrow\textwidth}
			\begin{tikzpicture}
			\begin{axis}[
			height=\figurewidth\linewidth,
			width=\linewidth,
			grid=major,
			legend pos=south east,
			legend style={inner xsep=0pt, inner ysep=0pt, font=\fontsize{1}{1}\selectfont},
			change x base,
			x SI prefix=kilo,x unit=\-,
			xmin=0,xmax=20000,
			ymin=0.2,ymax=1,
			xlabel=Probed Items , ylabel=Recall,
			]
	
				\addplot  [mark=triangle*,dashed,blue] 
				coordinates 
				{
					(1.293, 0.0126)
					(2.462, 0.02235)
					(4.737, 0.0390499)
					(9.107, 0.0655999)
					(17.636, 0.10715)
					(34.098, 0.1586)
					(66.401, 0.2291)
					(131.114, 0.32335)
					(260.59, 0.42625)
					(517.042, 0.5403)
					(1029.51, 0.651251)
					(2054.43, 0.753801)
					(4102.73, 0.844351)
					(8200.81, 0.911351)
					(16396.9, 0.955001)
					(32784.3, 0.98065)
					(65537.6, 1.0)
					(131082.0, 1.0)
					(136736.0, 1.0)
				};
				\addlegendentry{Cross-LSH}
				\addplot   
				coordinates 
				{
					(1.555, 0.0131)
					(2.809, 0.02195)
					(5.249, 0.0386999)
					(9.408, 0.0648499)
					(17.469, 0.10295)
					(33.806, 0.16085)
					(66.481, 0.24555)
					(131.856, 0.3566)
					(262.374, 0.486649)
					(518.279, 0.628051)
					(1030.95, 0.745751)
					(2053.25, 0.825101)
					(4097.77, 0.887351)
					(8193.24, 0.988151)
					(16385.0, 0.9981)
					(32768.7, 1.0)
					(65537.6, 1.0)
					(131082.0, 1.0)
					(136736.0, 1.0)
				};
				\addlegendentry{Cross-LSH}
				
			\end{axis}
			\end{tikzpicture}
	\end{subfigure}%
	\hspace{\figuregap}
	\begin{subfigure}[b]{\figurepercentperrow\textwidth}
	\begin{tikzpicture}
	\begin{axis}[
	height=\figurewidth\linewidth,
	width=\linewidth,
	legend pos=south east,
	legend style={inner xsep=0pt, inner ysep=0pt, font=\fontsize{1}{1}\selectfont},
	grid=major,
	change x base,
	x SI prefix=kilo,x unit=\-,
	xlabel=Probed Items,
	xmin=0,xmax=20000,
	ymin=0.2,ymax=1,
	]
	
	\addplot  [mark=triangle*,dashed,blue] 
	coordinates 
	{
		(37.414, 0.00795001)
		(44.937, 0.01215)
		(67.696, 0.0163)
		(85.9, 0.02065)
		(116.934, 0.0342)
		(152.909, 0.0571999)
		(206.332, 0.075)
		(282.601, 0.0920001)
		(404.0, 0.1173)
		(688.146, 0.17155)
		(1248.76, 0.2312)
		(2235.33, 0.2951)
		(4323.25, 0.39205)
		(8413.33, 0.50755)
		(16592.9, 0.6435)
		(32984.1, 0.790901)
		(65758.1, 0.924651)
		(131152.0, 0.9991)
		(136736.0, 1.0)
	};
	\addlegendentry{L2-ALSH}
	\addplot coordinates 
	{
		(3.086, 0.00680001)
		(4.719, 0.0116)
		(7.329, 0.01875)
		(12.796, 0.03055)
		(22.726, 0.0497999)
		(39.009, 0.0782999)
		(73.836, 0.1237)
		(143.679, 0.18925)
		(268.473, 0.26975)
		(525.072, 0.38925)
		(1034.06, 0.5281)
		(2056.99, 0.66505)
		(4101.15, 0.781751)
		(8195.94, 0.858801)
		(16387.7, 0.914001)
		(32771.8, 0.955001)
		(65542.6, 0.986351)
	};
	\addlegendentry{L2-ALSH-NR}

	\end{axis}
	\end{tikzpicture}
\end{subfigure}%
	\hspace{\figuregap}
	\begin{subfigure}[b]{\figurepercentperrow\textwidth}
	\begin{tikzpicture}
	\begin{axis}[
	height=\figurewidth\linewidth,
	width=\linewidth,
	legend pos=south east,
	legend style={inner xsep=0pt, inner ysep=0pt, font=\fontsize{1}{1}\selectfont},
	grid=major,
	change x base,
	x SI prefix=kilo,x unit=\-,
	xlabel=Probed Items ,
	xmin=0,xmax=20000,
	ymin=0.2,ymax=1,
	]
	
	\addplot  [mark=triangle*,dashed,blue] 
	coordinates 
	{
		(2.184, 0.0054)
		(3.985, 0.00895001)
		(7.288, 0.0173)
		(13.866, 0.0263)
		(23.652, 0.0403499)
		(41.99, 0.0613499)
		(78.428, 0.09465)
		(147.699, 0.1352)
		(276.781, 0.18755)
		(536.383, 0.2632)
		(1055.89, 0.3489)
		(2083.34, 0.446149)
		(4135.04, 0.5527)
		(8244.75, 0.655551)
		(16436.3, 0.765901)
		(32835.4, 0.863851)
		(65624.6, 0.944701)
		(131137.0, 0.9985)
		(136736.0, 1.0)
	};
	\addlegendentry{Simple-LSH}
	\addplot coordinates 
	{
		(2.011, 0.00925001)
		(3.571, 0.0139)
		(6.585, 0.02215)
		(11.608, 0.0318499)
		(21.183, 0.0453499)
		(38.136, 0.0674499)
		(73.202, 0.10905)
		(137.3, 0.16035)
		(265.76, 0.2407)
		(523.941, 0.3605)
		(1033.73, 0.5171)
		(2056.5, 0.6852)
		(4100.3, 0.857651)
		(8195.46, 0.945301)
		(16387.8, 0.978701)
		(32772.6, 0.9929)
		(65541.6, 0.9989)
		(131084.0, 1.0)
		(136736.0, 1.0)
	};
	\addlegendentry{Simple-LSH-NR}
	
	\end{axis}
	\end{tikzpicture}
\end{subfigure}%
	\hspace{\figuregap}
	\begin{subfigure}[b]{\figurepercentperrow\textwidth}
	\begin{tikzpicture}
	\begin{axis}[
	height=\figurewidth\linewidth,
	width=\linewidth,
	legend pos=south east,
	legend style={inner xsep=0pt, inner ysep=0pt, font=\fontsize{1}{1}\selectfont},
	grid=major,
	change x base,
	x SI prefix=kilo,x unit=\-,
	xlabel=Probed Items , 
	xmin=0,xmax=20000,
	ymin=0.2,ymax=1,
	]
	
	\addplot  [mark=triangle*,dashed,blue] 
	coordinates 
	{
		(1.833, 0.00545)
		(3.359, 0.00825001)
		(5.978, 0.0134)
		(11.077, 0.02315)
		(19.936, 0.0358999)
		(37.063, 0.0600999)
		(71.018, 0.0928001)
		(135.428, 0.14905)
		(264.425, 0.2198)
		(521.718, 0.3195)
		(1034.47, 0.4423)
		(2061.33, 0.57815)
		(4111.58, 0.71435)
		(8210.93, 0.82935)
		(16405.2, 0.919001)
		(32791.7, 0.971801)
		(65569.0, 0.9934)
		(131124.0, 1.0)
		(136736.0, 1.0)
	};
	\addlegendentry{Sign-ALSH}
	\addplot coordinates 
	{
		(4.039, 0.01275)
		(5.598, 0.0173)
		(8.963, 0.0275)
		(13.773, 0.0383499)
		(23.136, 0.0574999)
		(40.377, 0.0920001)
		(73.602, 0.14665)
		(137.389, 0.2319)
		(267.775, 0.34645)
		(524.387, 0.4906)
		(1037.02, 0.6461)
		(2057.43, 0.78945)
		(4100.62, 0.917151)
		(8195.14, 0.976351)
		(16386.5, 0.99355)
		(32770.0, 0.99855)
		(65537.6, 0.99965)
		(131082.0, 1.0)
		(136736.0, 1.0)
	};
	\addlegendentry{Sign-ALSH-NR}				
	\end{axis}
	\end{tikzpicture}
\end{subfigure}%
	
		\begin{subfigure}[b]{\figurepercentperrow\textwidth}
			\begin{tikzpicture}
			\begin{axis}[
				height=\figurewidth\linewidth,
				width=\linewidth,
				grid=major,
				legend pos=south east,
				legend style={inner xsep=0pt, inner ysep=0pt, font=\fontsize{1}{1}\selectfont},
				change x base,
				x SI prefix=mega,x unit=\-,
				xmin=0,xmax=400000,
				ymin=0.2,ymax=1,
				xlabel=Probed Items , ylabel=Recall,
			]

				\addplot [mark=triangle*,dashed,blue] 
				coordinates 
				{
					(1.069, 0.00670001)
					(2.092, 0.01415)
					(4.144, 0.02495)
					(8.193, 0.0420499)
					(16.312, 0.0662999)
					(32.528, 0.10365)
					(64.844, 0.1584)
					(130.554, 0.2224)
					(259.682, 0.30435)
					(523.785, 0.394)
					(1052.96, 0.4917)
					(2109.28, 0.58265)
					(4163.57, 0.6649)
					(8477.82, 0.7353)
					(16998.2, 0.790251)
					(33838.5, 0.833451)
					(67389.9, 0.869601)
					(134601.0, 0.900101)
					(266778.0, 0.9277)
					(530166.0, 0.954251)
					(1052820.0, 0.97975)
				};
				\addlegendentry{Cross-LSH}
				\addplot  
				coordinates 
				{
					(1.029, 0.0049)
					(2.043, 0.00900001)
					(4.173, 0.01715)
					(8.304, 0.02825)
					(16.412, 0.0461999)
					(32.584, 0.0756)
					(64.621, 0.12045)
					(128.969, 0.18235)
					(257.198, 0.2661)
					(514.132, 0.37225)
					(1025.95, 0.49735)
					(2049.56, 0.62735)
					(4097.35, 0.74065)
					(8192.88, 0.825401)
					(16384.6, 0.882101)
					(32769.0, 0.91995)
					(65536.4, 0.9655)
					(131072.0, 0.99945)
					(262144.0, 0.99995)
					(524288.0, 1.0)
					(1048580.0, 1.0)
					(2097150.0, 1.0)
					(2340370.0, 1.0)
				};
				\addlegendentry{Cross-LSH}
				
			\end{axis}
			\end{tikzpicture}	
	\end{subfigure}%
	\hspace{\figuregap}
	\begin{subfigure}[b]{\figurepercentperrow\textwidth}
		\begin{tikzpicture}
		\begin{axis}[
		height=\figurewidth\linewidth,
		width=\linewidth,
		grid=major,
		legend pos=south east,
		legend style={inner xsep=0pt, inner ysep=0pt, font=\fontsize{1}{1}\selectfont},
		change x base,
		x SI prefix=mega,x unit=\-,
		xlabel=Probed Items ,
		xmin=0,xmax=400000,
		ymin=0.2,ymax=1,
		]
		
		\addplot  [mark=triangle*,dashed,blue]
		coordinates 
		{
			(2.886, 0.00045)
			(4.558, 0.0006)
			(11.566, 0.0009)
			(22.402, 0.00195)
			(38.274, 0.0033)
			(75.927, 0.0064)
			(163.772, 0.0102)
			(275.612, 0.0158)
			(517.158, 0.02465)
			(1010.7, 0.03455)
			(1906.82, 0.0475999)
			(4161.62, 0.0730499)
			(8559.89, 0.10695)
			(16703.5, 0.14765)
			(29667.6, 0.202)
			(57484.8, 0.2787)
			(110982.0, 0.3678)
			(190015.0, 0.4388)
			(313390.0, 0.5429)
			(578900.0, 0.6629)
			(1119210.0, 0.7819)
			(2111280.0, 0.955701)
			(2340370.0, 1.0)
			
		};
		\addlegendentry{L2-ALSH}
		\addplot coordinates 
		{
			(2.72, 0.002)
			(4.605, 0.003)
			(7.125, 0.00425)
			(12.01, 0.0075)
			(19.845, 0.01)
			(37.605, 0.01625)
			(72.71, 0.03125)
			(139.66, 0.051)
			(272.875, 0.07125)
			(529.6, 0.1255)
			(1047.91, 0.19875)
			(2101.01, 0.30575)
			(4145.02, 0.415)
			(8296.3, 0.53325)
			(16469.7, 0.6515)
			(32928.7, 0.78075)
			(65556.0, 0.88)
			(131081.0, 0.94075)
			(262150.0, 0.96325)
			(524290.0, 0.97025)
			(1048580.0, 0.9755)
			(2097170.0, 0.99125)
			(2340370.0, 1.0)
		};
		\addlegendentry{L2-ALSH-NR}
		
		\end{axis}
		\end{tikzpicture}	
	\end{subfigure}%
	\hspace{\figuregap}	
	\begin{subfigure}[b]{\figurepercentperrow\textwidth}
		\begin{tikzpicture}
		\begin{axis}[
		height=\figurewidth\linewidth,
		width=\linewidth,
		legend pos=south east,
		legend style={inner xsep=0pt, inner ysep=0pt, font=\fontsize{1}{1}\selectfont},
		grid=major,
		change x base,
		x SI prefix=mega,x unit=\-,
		xlabel=Probed Items ,
		xmin=0,xmax=400000,
		ymin=0.2,ymax=1,
		]
		\addplot  [mark=triangle*,dashed,blue] 
		coordinates 
		{
			(1.394, 0.00115)
			(33.772, 0.0027)
			(40.304, 0.0045)
			(45.257, 0.00775001)
			(61.221, 0.01295)
			(185.786, 0.0211)
			(247.885, 0.03455)
			(344.632, 0.0543499)
			(580.825, 0.0864501)
			(1051.26, 0.12945)
			(1879.94, 0.1829)
			(3362.95, 0.2433)
			(5743.07, 0.3167)
			(10573.1, 0.39455)
			(20450.1, 0.46835)
			(41469.9, 0.5435)
			(77510.7, 0.6125)
			(152780.0, 0.682951)
			(290793.0, 0.747351)
			(560941.0, 0.823402)
			(1091640.0, 0.904151)
			(2114840.0, 0.988701)
			(2340370.0, 1.0)
		};
		\addlegendentry{Simple-LSH}
		\addplot coordinates 
		{
			(1.825, 0.00105)
			(3.151, 0.00245)
			(5.493, 0.0041)
			(9.848, 0.0067)
			(19.091, 0.01275)
			(37.632, 0.0226)
			(71.679, 0.0384999)
			(139.107, 0.0618499)
			(267.805, 0.10005)
			(527.836, 0.15145)
			(1053.13, 0.22685)
			(2099.87, 0.3261)
			(4155.85, 0.4432)
			(8261.72, 0.56605)
			(16471.4, 0.692101)
			(32853.8, 0.797151)
			(65611.0, 0.869)
			(131075.0, 0.977651)
			(262145.0, 0.99585)
			(524288.0, 0.99935)
			(1048580.0, 0.99995)
			(2097150.0, 1.0)
			(2340370.0, 1.0)
		};
		\addlegendentry{Simple-LSH-NR}
		\end{axis}
		\end{tikzpicture}	
	\end{subfigure}%
	\hspace{\figuregap}	
	\begin{subfigure}[b]{\figurepercentperrow\textwidth}
	\begin{tikzpicture}
	\begin{axis}[
		height=\figurewidth\linewidth,
		width=\linewidth,
		grid=major,
		legend pos=south east,
		legend style={inner xsep=0pt, inner ysep=0pt, font=\fontsize{1}{1}\selectfont},
		change x base,
		x SI prefix=mega,x unit=\-,
		xlabel=Probed Items,
		xmin=0,xmax=400000,
		ymin=0.2,ymax=1,
	]
	\addplot  [mark=triangle*,dashed,blue] 
	coordinates 
	{
		(16.57, 0.0019)
		(43.869, 0.00355)
		(66.626, 0.0068)
		(83.99, 0.0092)
		(105.048, 0.0146)
		(130.412, 0.0233)
		(196.002, 0.03575)
		(288.185, 0.0512999)
		(629.776, 0.07575)
		(987.22, 0.1065)
		(1752.06, 0.14195)
		(4442.29, 0.18325)
		(7520.13, 0.2435)
		(13726.0, 0.30555)
		(24169.5, 0.38595)
		(47143.5, 0.4592)
		(94542.7, 0.5349)
		(177046.0, 0.627701)
		(334760.0, 0.724351)
		(585994.0, 0.795951)
		(1108880.0, 0.887601)
		(2131050.0, 0.990451)
		(2340370.0, 1.0)
	};
	\addlegendentry{Sign-ALSH}
	\addplot coordinates 
	{
		(4.928, 0.0025)
		(7.478, 0.0032)
		(11.97, 0.00535)
		(26.734, 0.0126)
		(36.071, 0.0164)
		(57.596, 0.0238)
		(96.075, 0.0362499)
		(157.194, 0.0523499)
		(288.922, 0.0824501)
		(617.346, 0.14005)
		(1183.32, 0.20405)
		(2279.71, 0.30825)
		(4393.33, 0.434)
		(8498.22, 0.56765)
		(16664.8, 0.69765)
		(33046.8, 0.802151)
		(65729.4, 0.87865)
		(131079.0, 0.97745)
		(262152.0, 0.994)
		(524290.0, 0.9986)
		(1048580.0, 1.0)
		(2097150.0, 1.0)
		(2340370.0, 1.0)
	};
	\addlegendentry{Sign-ALSH-NR}
	\end{axis}
	\end{tikzpicture}	
	\end{subfigure}%
	\caption{Probed item-recall comparison between the original meta algorithms and their norm-range versions under a code length of 32. From top row to bottom row, the datasets are Netflix, Yahoo!Music and ImageNet, respectively.} 
	\label{fig:improvement of norm range}
\end{figure*}  

 Similar ideas was used in~\citep{andoni:optimal}.
                         
\subsection{Practical Considerations}\label{subsec:rank}
Although theoretically norm-range partition provides lower query processing complexity, several issues need to be solved in order to enjoy its performance benefits in practice.

Algorithm~\ref{alg:index building} uses independently generated hash functions to build index for different sub-datasets. Assume there are $w$ sub-datasets and each sub-dataset has $L$ hash functions, the query needs to be hashed $Lw$ times. However, if a single hash index is used for the entire dataset, the query only needs to be hashed $L$ times. To reduce the complexity of query processing, we use the same hash functions for different sub-datasets for index building so that the query only needs to be hashed $L$ times to generate a single hash signature. This hash signature is then used to search then hash index of different sub-datasets.

The theoretical guarantee (i.e., return a good approximate MIPS with high probability) of LSH only holds when using multiple hash tables. However, building multiple hash tables incurs high memory cost~\citep{lv:multi-probe}. In practice, LSH is usually used in a single-table fashion for candidate generation~\citep{li:gqr}. Items are put into buckets according to their hash codes and the buckets are ranked according to the number of identical hashes they have with the query. Then items in top-ranked buckets are retrieved as candidates for further verification. However, norm-range partition builds a hash index for each sub-dataset and how to rank the buckets from different sub-datasets is not straightforward. For example, even if bucket $b_i$ has less identical hashes than $b_j$, $b_i$ could still be more favorable if it is from a sub-dataset with larger $M_j$. Therefore, we formulate a framework that allows to rank the buckets from different sub-datasets.

After the transformations in Section~\ref{sec:LSH for MIPS}, the hash collision probability can be expressed as a function of the inner product between the query and the item, i.e., $P\left[h'(x)=h(q)\right]=g(x^{\top}q)$~\footnote{We use $h'$ and $h$ for the hash functions of the query and the item respectively as there may be asymmetry.}. For example, the collision probability of Simple-LSH is $P=1-\frac{1}{\pi} \cos^{-1} \left( \frac{x^{\top} q}{M_j}\right)$. 
Function $g(.)$ is a monotonically increasing function of inner product as a valid LSH needs to have higher collision probability for larger inner product. This means there exists an inverse function $g^{-1}(.)$ for $g(.)$. Assume there are $L$ hashes in total and a bucket has $l$ identical hashes with the query $q$, we can get an estimate of the collision probability $l/L$. Plug the estimate into $g^{-1}(.)$ we get an estimate of the inner product as $\hat{s}=g^{-1}(l/L)$. For Simple-LSH, we have $\hat{s}=M_j\cos\left[\pi(1-\frac{l}{L})\right]$ for a bucket from sub-dataset $\mathcal{S}_j$. We can use $\hat{s}$ as a similarity metric to rank the buckets from different sub-datasets. This framework is general and we show how to apply it to $L_2$-ALSH, Sign-ALSH and Cross-LSH in the supplementary material. 

Ranking real valued similarity metric $\hat{s}$ is still more complex than ranking the buckets in a single hash index, which can be conducted by efficient radix sort. As the similarity metrics of all LSH algorithms take the form $M_jg^{-1}(l/L)$, we can sort the all $(M_j, l)$ pairs in the index building phase and store a sorted list that is common for all queries. Query can be processed by transversing the sorted list by using $M_j$ to locate the sub-dataset and using $l$ to locate the buckets via standard hash lookup. In this case, sorting is not needed for on-line query processing.                 

\section{Experiment Results}\label{sec:experiment}
For experiment evaluation, we used three popular datasets, i.e., Netflix, Yahoo!Music and ImageNet. Netflix and Yahoo!Music record the ratings users give for items and are commonly used for collaborative filtering. We obtained user and item embeddings from these two datasets using alternating least square (ALS)~\citep{yun:als} based matrix factorization, and set the dimensionality of the embeddings as 300. The item embeddings and user embeddings are regraded as database vectors and queries, respectively. The Netflix dataset contains 17,770 items and the Yahoo!Music dataset contains 136,736 items. The ImageNet dataset contains 2,340,373 SIFT descriptors of the ImageNet images and each descriptor has 150 features. We randomly sampled 1,000 descriptors as queries and used the remaining descriptors as database vectors. The items of the three datasets have very different norm distributions (see the supplementary material), which helps verify the robustness of our methods to norm distribution.

To keep our experiment consistent with existing works~\citep{shrivastava:signalsh,neyshabur:simple-lsh}, we test the performance of our methods when used for single hash table based candidate generation. We report the probed item-recall curve for top-$k$ MIPS, which is obtained as follows: the items are ranked according to the hash index for each query and the average recall of 1,000 queries is calculated when $T$ items are probed. We report the results for top-20 MIPS in the paper and the results for other values of $k$ can be found in the supplementary material. For both Sign-ALSH and $L_2$-ALSH, we used the parameter ($U$, $m$ and $r$) settings recommended by their authors. The datasets are partitioned into 32, 64 and 128 sub-datasets for norm-range partition under a code length of 16, 32 and 64, respectively. For fairness of comparison, we use shorter code for norm-range partition as norm-range partition will generate more buckets than the meta algorithm if the same code length is used. To be more specific, when the meta algorithm uses a code length of 16, 32 and 64, norm-range partition uses code length of 11, 26, 57, respectively. The setting is aimed at ensuring that norm-range partition and the meta algorithm generate a similar number of buckets so that their memory costs are similar. Due to space limit, we report the performance under a code length of 32 in the paper, and the performance under other code lengths can be found in the supplementary material.

We compare the performance of Cross-LSH with existing LSH based MIPS algorithms in Figure~\ref{fig:cross-LSH vs others}. The results show that Cross-LSH consistently outperforms existing algorithms on the three datasets. Moreover, the performance improvement is more significant when the size of the dataset is large.

We compare the LSH based MIPS algorithms with their norm-range versions in Figure~\ref{fig:improvement of norm range}.The results show that norm-range partition provides performance improvement for all three datasets and all algorithms, which provides empirical evidence for Theorem~\ref{theorem:complexity}. The performance improvement is more significant when the size of the dataset is large. Moreover, $L_2$-ALSH attains the most significant improvement while the improvement on Cross-LSH is more moderate. This may be explained by the fact that Cross-LSH already performs very well originally and there is not to much room for improvement.

\section{Conclusion}\label{sec:conclusion}
In this paper, we surveyed existing LSH based MIPS algorithms and made the observation that these algorithms consist of two independent components: a transformation that maps MIPS to similarity search in another metric and an existing LSH to solve the transformed problem. Based on this observation, we improve existing LSH based MIPS algorithms in two directions. By connecting the transformation of Simple-LSH with cross-polytope LSH, we develop a new algorithm named Cross-LSH that outperforms existing ones. More interestingly, we found that query processing complexity can be reduced by using smaller normalization factor. To utilize this fact, we propose the norm-range partition technique, which divides the dataset into sub-datasets according to percentiles in the norm distribution and allows us to use smaller normalization factor for most sub-datasets. We prove that norm-range partition is a universal catalyst for all existing LSH based MIPS algorithms under mild conditions. Experiment results also show that norm-range partition consistently improves the performance of all LSH based MIPS algorithms.

\bibliography{norm-range}
\bibliographystyle{plainnat}
\newpage
\onecolumn
{\centering
	{\Large\bfseries Supplementary Material for Norm-range Partition: A Univiseral Catalyst for LSH based Maximum Inner Product Search (MIPS) \par}}
\vskip 0.2in 
\pagestyle{empty}
\appendix

\section{Norm distributions of some real datasets}

\begin{figure}[H]  
	\centering
	\includegraphics[width=0.31\columnwidth]{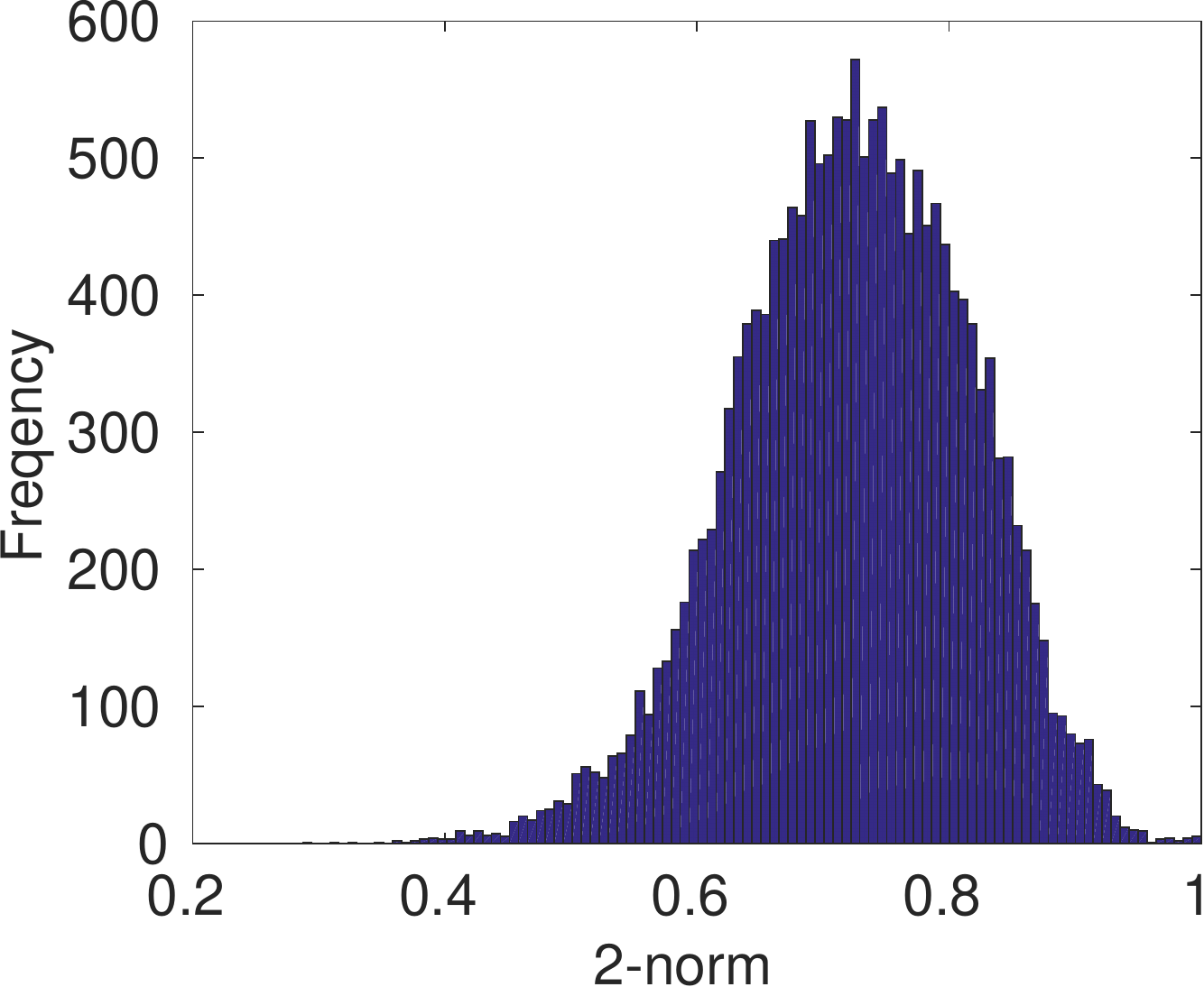} 
	\includegraphics[width=0.31\columnwidth]{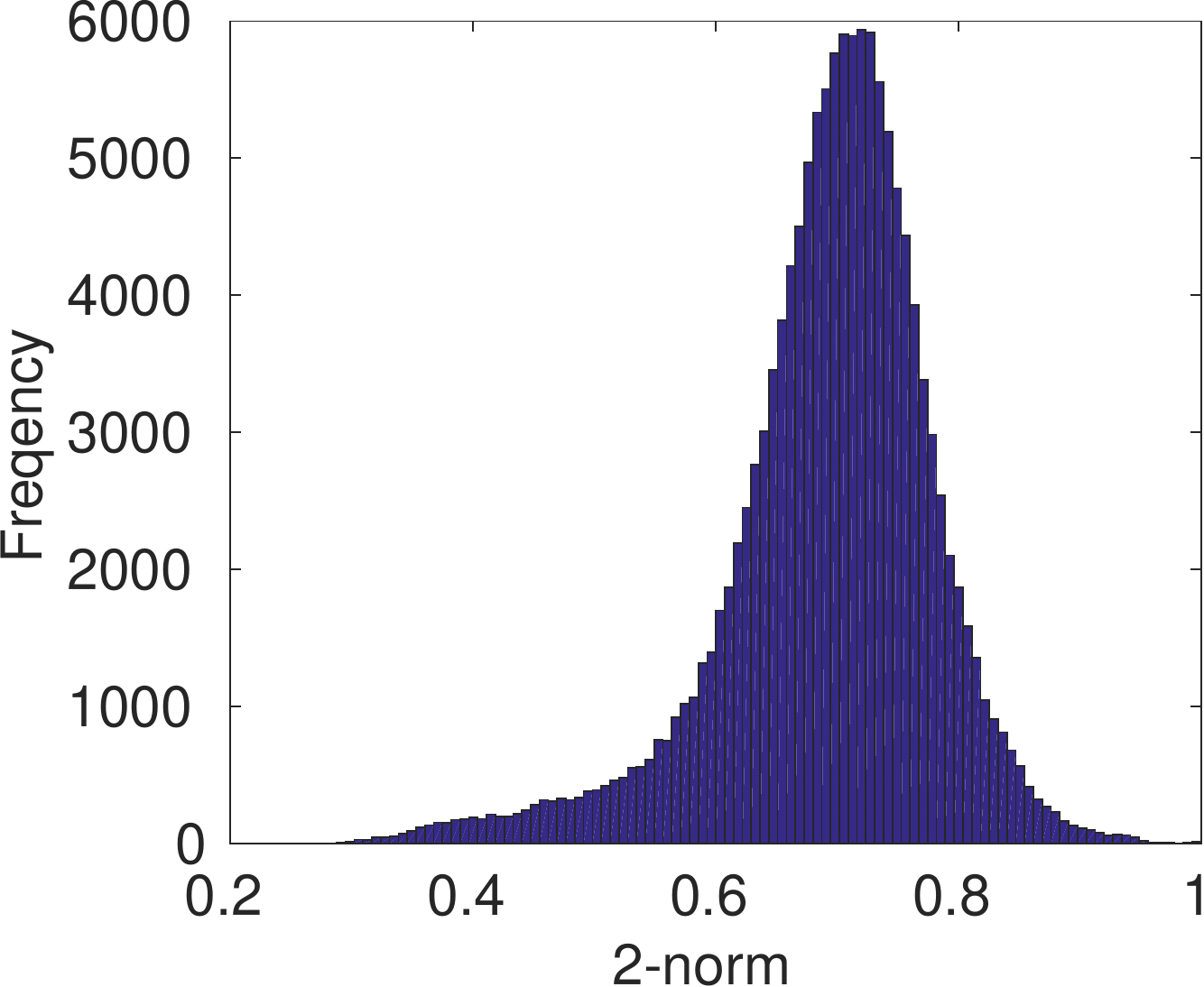}
	\includegraphics[width=0.31\columnwidth]{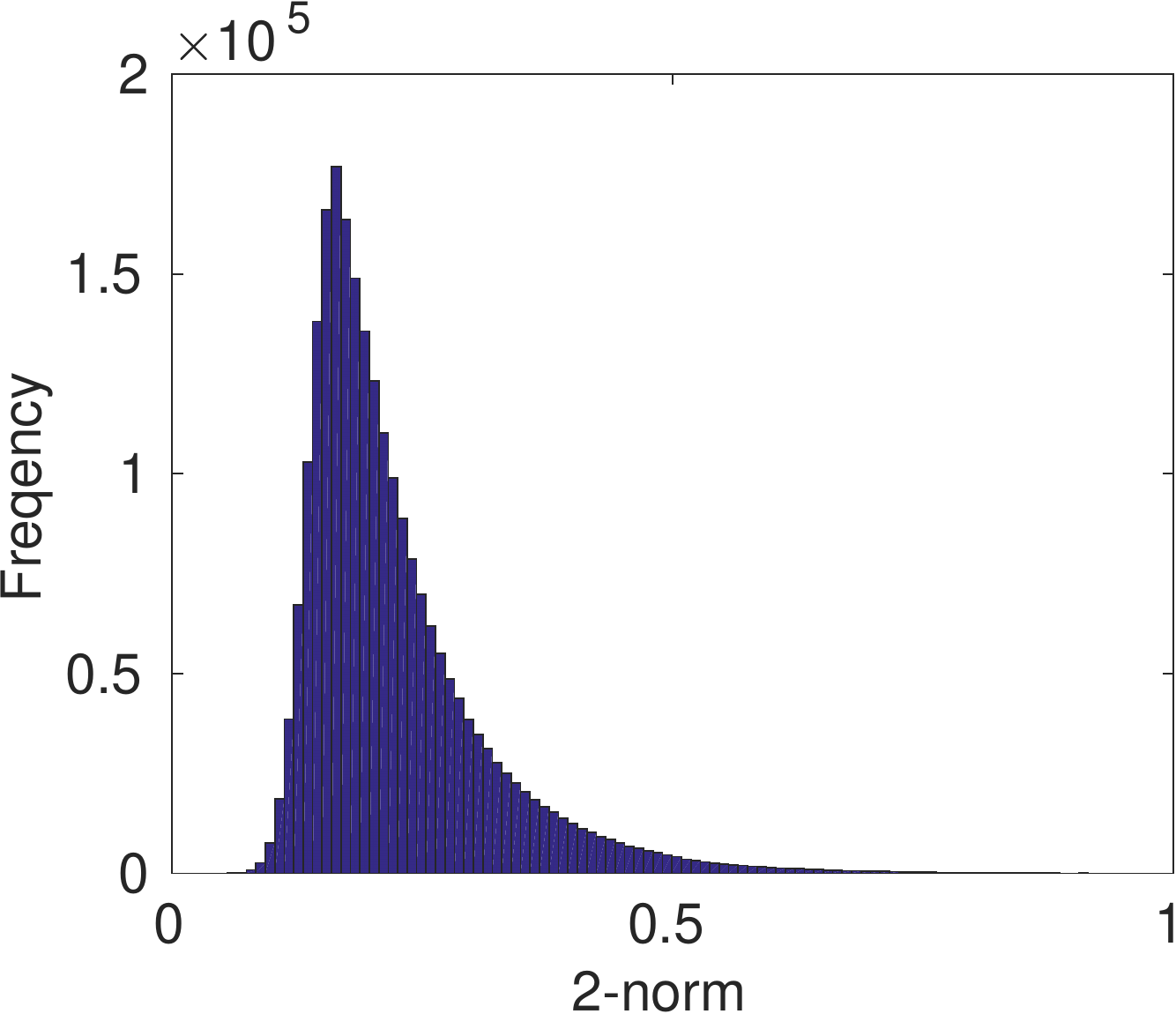}
	\caption{Norm distributions of the item embeddings of the Netflix dataset (left) and Yahoo!Music dataset (middle) and the SIFT descriptors of the ImageNet dataset (right), maximum norm scaled to 1. } 
	\label{fig:norm distribution}
\end{figure}

We show the norm distributions of the datasets used in the experiments in Figure~\ref{fig:norm distribution}. The figure shows that real datasets have large different in norm among the items, which motivates the norm-partition technique. For the three datasets, only one sub-dataset (the one that contains items with the largest norm) has $M_j=M$ after partition, which shows norm-range partition is effective in reducing the normalization factor. Moreover, the three distributions take different shapes, but norm-range partition works well on all of them, which shows empirically that norm-range partition is robust to different norm distribution.    

\section{Proof of Lemma~\ref{lemma:rho}}

Denote the hash quality $\rho$ as a function of normalization factor $M$, Lemma~\ref{lemma:rho} requires to prove that $\rho'(M)>0$ for any $0<S\le M$ and $0<c<1$.

\textbf{Simple-LSH}. For Simple-LSH, we have 
\begin{equation}
\rho(M)=\frac{\ln \left(1-\frac{\mathrm{cos}^{-1}(\frac{S}{M})}{\pi}\right)}{\ln \left(1-\frac{\mathrm{cos}^{-1}(\frac{cS}{M})}{\pi}\right)}.
\end{equation}   

Define $x=S/M$, it suffices to prove $\rho'(x)<0$ for $x\in\left[0, 1\right]$. Let $f(x)=\ln \left(1-\frac{\mathrm{cos}^{-1}(x)}{\pi}\right)$, we have $\rho(x)=\frac{f(x)}{f(cx)}$, therefore the sign of $\rho'(x)$ is decided by $f'(x)f(cx)-cf'(cx)f(x)$. With further deviation, we know that the sign of $\rho'(x)$ is decided by,

\begin{equation}
g(x)=\sqrt{1-c^2x^2} \left[\pi-\mathrm{cos}^{-1}(cx) \right] \ln (1-\frac{\mathrm{cos}^{-1}(cx)}{\pi})-\sqrt{1-x^2} \left[\pi-\mathrm{cos}^{-1}(x) \right] \ln (1-\frac{\mathrm{cos}^{-1}(x)}{\pi}).
\end{equation}

Define variables $t=1-\frac{\mathrm{cos}^{-1}(x)}{\pi}$ and $r=1-\frac{\mathrm{cos}^{-1}(cx)}{\pi}$, we have $t,r\in \left[0.5, 1\right]$ and $t>r$ as $0<c<1$. We can re-express $g(x)$ as

\begin{equation}
g(x)=r\sin(\pi r) \ln r-ct\sin(\pi t) \ln t,
\end{equation}
As $t\sin(\pi t) \ln t<0$ and $0<c<1$, $g(x)<r\sin(\pi r) \ln r-t\sin(\pi t) \ln t$, therefore, it suffices to prove that $B(t)=t\sin(\pi t) \ln t$ in an increasing function for $t\in \left[0.5, 1\right]$. We have $B'(t)=\sin(\pi t) (1+\ln t)+t \ln t \cos(\pi t)$, in which both terms are positive for $t\in \left[0.5, 1\right]$. Thus, we have proved $\rho'(M)>0$ for Simple-LSH.

\textbf{Sign-ALSH}. For Sign-ALSH, assuming the $\Vert \frac{Ux}{M} \Vert^{2^{m+1}}$ term in~\eqref{equ:Sign-ALSH} is small enough to be ignored, its hash quality $\rho$ can be expressed as,       

\begin{equation}\label{equ:sign_rho}
\rho=\frac{\log \left(1-\frac{\mathrm{cos}^{-1}(\frac{2US}{M\sqrt{m}})}{\pi}\right)}{\log \left(1-\frac{\mathrm{cos}^{-1}(\frac{2cUS}{M\sqrt{m}})}{\pi}\right)}
\end{equation} 

which is takes similar form as the $\rho$ of Simple-LSH. Define $S'=2S/\sqrt{m}$ and $0<S'\le M$ (needed to make the LSH valid), the $\rho$ of Sign-ALSH can be expressed as 

\begin{equation}
\rho(M)=\frac{\ln \left(1-\frac{\mathrm{cos}^{-1}(\frac{S'}{M})}{\pi}\right)}{\ln \left(1-\frac{\mathrm{cos}^{-1}(\frac{cS'}{M})}{\pi}\right)}.
\end{equation}    

Therefore, $\rho'(M)>0$ can be proved following the same procedure as Simple-LSH.

\textbf{Cross-LSH}. Taking derivative of the $\rho$ in~\eqref{equ:cross rho} shows the sign of $\rho'(M)$ is decided by

\begin{equation}
f(M)=2S(1-c)(M^2-cS^2),
\end{equation}   

which is positive as $S<M$ and $c<1$.

\textbf{$L_2$-ALSH}. For $L_2$-LSH, assuming the $\Vert \frac{Ux}{M} \Vert^{2^{m+1}}$ term  is small enough to be ignored, its hash quality $\rho$ can be expressed as:
 
\begin{equation}
\rho=\frac{\log F_r\left(\sqrt{1+\frac{m}{4}-2\frac{US}{M}}\right)}{\log F_r\left(\sqrt{1+\frac{m}{4}-2\frac{cUS}{M}}\right)}.
\end{equation}  
The transformation of $L_2$-ALSH maps a $(S, c)$-MIPS problem to a $(d,\alpha)$-Euclidean similarity search problem, with $d=\sqrt{1+\frac{m}{4}-2\frac{US}{M}}$ and $\alpha=\frac{\sqrt{1+\frac{m}{4}-2\frac{cUS}{M}}}{\sqrt{1+\frac{m}{4}-2\frac{US}{M}}}$. If $M_j<M$ and denote the result problem the problem transformed using $M_j$ as $(d,\alpha)$ and the problem transformed using $M$ as $(d',\alpha')$, we can prove $d<d'$ and $\alpha>\alpha'$. For the hash function in~\eqref{equ:l2lsh}, it has the property that $\rho(d,\alpha)<\rho(d',\alpha)$ if $d<d'$, and $\rho(d,\alpha)<\rho(d,\alpha')$ if $\alpha>\alpha'$. Therefore, we have $\rho(d,\alpha)<\rho(d',\alpha')$.   
 
\section{Bucket ranking metrics for $L_2$-ALSH, Sign-ALSH and Cross-LSH}
We have shown how to use the general framework in Section~\ref{subsec:rank} to rank the buckets across the buckets from the index of different sub-datasets for Simple-LSH. In this section, we apply the framework to $L_2$-ALSH, Sign-ALSH and Cross-LSH.

\textbf{$L_2$-ALSH}. Assume that the $\Vert \frac{Ux}{M} \Vert^{2^{m+1}}$ term can be ignored, the Euclidean distance between $P(x)$ and $Q(q)$ is $\Vert P(x)-Q(q) \Vert=\sqrt{1+\frac{m}{4}-2\frac{U}{M}x^{\top}q}$. For the Euclidean distance LSH in~\eqref{equ:l2lsh}, the collision probability is expressed as a function $F_r(d)$ of distance. Denote the inverse function of $F_r(d)$ as $g(y)$, and suppose a bucket has $l$ identical hashes with the query and the total number of hashes is $L$. For sub-dataset $\mathcal{S}_j$, which uses $M_j$ as normalization factor, we can get an estimate of inner product as $\hat{s}=\frac{M}{2U}\left[1+\frac{m}{4}-g(\frac{l}{L})^2\right]$.

\textbf{Sign-ALSH}. Assume that the $\Vert \frac{Ux}{M} \Vert^{2^{m+1}}$ term can be ignored, the angular similarity between $P(x)$ and $Q(q)$ can be expressed as $\frac{Q(q)^{\top}P(x)}{\Vert Q(q) \Vert \Vert P(x) \Vert }=\frac{2Uq^{\top}x}{M\sqrt{m}}$. Connect it with the collision probability in~\eqref{equ:sign_prob} and follow a procedure similar to $L_2$-ALSH, we can obtain $\hat{s}=\frac{M\sqrt{m}}{2U}\cos\left[\pi (1-\frac{l}{L})\right]$.

\textbf{Cross-LSH}. For Cross-LSH, designing a similarity metric to rank the buckets across sub-datasets is more challenging as its collision probability does not have a closed-form expression. Cross-ploytope LSH is also more complex and single hash table based multi-probe already requires a similarity metric. The authors designed a similarity metric $\Vert y_{x,v} \Vert^2$, which is vector that can deduced from the hash value of the query and the bucket under consideration. Please refer to {REF:Optimal} for the precise definition of $y_{x,v}$. We found that the joint distribution of distance $d$ and $\Vert y_{x,v} \Vert^2$ can be expressed as $p(d, \Vert y_{x,v} \Vert) \propto e^{-\Vert y_{x,v} \Vert^2(\frac{4}{4d^2-d^4}-1)}$, which means given $\Vert y_{x,v} \Vert^2$, $\frac{4}{4d^2-d^4}-1$ follows an exponential distribution and we can use its expectation $1/\Vert y_{x,v} \Vert^2$ as an estimate of it. Then we can solve the equation $\frac{4}{4d^2-d^4}-1=1/\Vert y_{x,v} \Vert^2$ for $d$. As distance between vectors on the unit sphere lies in $\left[0, 2\right]$, we only need the solution in this range. Connect $d$ with the transformation of Simple-LSH, the similarity metric is $\hat{s}=M_j(1-d^2/2)$.   
        
\section{Experiment results for top 1, top 10, top 50 MIPS (32 bit)}
%%%%%%=================================%%%%%%
\begin{figure*}[t]
		\begin{subfigure}[b]{0.36\textwidth}
			\begin{tikzpicture}
			\begin{axis}[
			height=\figurewidth\linewidth,
			width=\linewidth,
			legend style={font=\fontsize{1}{1}\selectfont},
			grid=major,
			legend pos=south east,
			change x base,
			x SI prefix=kilo,x unit=\-,
			xlabel=Probed Items , ylabel=Recall,
			xmin=0,xmax=4000,
			ymin=0,ymax=1,
			]
			
					\addplot [mark=square*, red]
					coordinates 
					{
						(1.956, 0.06)
						(3.367, 0.083)
						(5.745, 0.123)
						(10.746, 0.182)
						(19.911, 0.262)
						(38.253, 0.359)
						(82.004, 0.492)
						(158.274, 0.614)
						(339.856, 0.72)
						(634.608, 0.772)
						(1135.33, 0.834)
						(2130.34, 0.924)
						(4123.87, 0.982)
						(8201.29, 1.0)
						(16385.4, 1.0)
						(17770.0, 1.0)
					};
					\addlegendentry{Cross-LSH}
					
					\addplot  [mark=triangle*, blue]
					coordinates 
					{
						(9.294, 0.065)
						(11.16, 0.092)
						(13.082, 0.106)
						(17.92, 0.138)
						(24.575, 0.164)
						(43.384, 0.24)
						(85.428, 0.279)
						(142.613, 0.365)
						(278.485, 0.529)
						(573.04, 0.627)
						(1105.26, 0.748)
						(2204.51, 0.832)
						(4228.29, 0.927)
						(8281.92, 0.981)
						(16539.7, 1.0)
						(17770.0, 1.0)
					};
					\addlegendentry{Simple-LSH}
					\addplot 
					coordinates 
					{
						(17.786, 0.01)
						(19.472, 0.014)
						(23.312, 0.017)
						(29.185, 0.024)
						(36.983, 0.028)
						(65.372, 0.031)
						(98.869, 0.068)
						(179.253, 0.116)
						(322.414, 0.244)
						(627.332, 0.464)
						(1221.68, 0.626)
						(2248.05, 0.774)
						(4176.56, 0.923)
						(8351.56, 0.988)
						(16390.6, 1.0)
						(17770.0, 1.0)
					};
					\addlegendentry{L2-ALSH}
					\addplot 
					coordinates 
					{
						(2.552, 0.096)
						(4.077, 0.128)
						(6.804, 0.156)
						(13.211, 0.201)
						(32.038, 0.245)
						(50.098, 0.289)
						(77.63, 0.399)
						(145.34, 0.538)
						(294.815, 0.618)
						(559.7, 0.69)
						(1085.42, 0.812)
						(2117.53, 0.891)
						(4211.13, 0.944)
						(8295.86, 0.99)
						(16395.4, 1.0)
						(17770.0, 1.0)
					};
					\addlegendentry{Sign-ALSH}

			\end{axis}
			\end{tikzpicture}
	\end{subfigure}%
	\hspace{\figuregap}	
	\begin{subfigure}[b]{0.36\textwidth}
			\begin{tikzpicture}
			\begin{axis}[
			height=\figurewidth\linewidth,
			width=\linewidth,
			legend style={font=\fontsize{1}{1}\selectfont},
			grid=major,
			legend pos=south east,
			change x base,
			x SI prefix=kilo,x unit=\-,
			xlabel=Probed Items ,
			xmin=0,xmax=120000,
			ymin=0,ymax=1,
			]
			
				\addplot [mark=square*, red]
				coordinates 
				{
					(1.293, 0.031)
					(2.462, 0.056)
					(4.737, 0.088)
					(9.107, 0.146)
					(17.636, 0.231)
					(34.098, 0.299)
					(66.401, 0.394)
					(131.114, 0.487)
					(260.59, 0.59)
					(517.042, 0.681)
					(1029.51, 0.776)
					(2054.43, 0.84)
					(4102.73, 0.901)
					(8200.81, 0.947)
					(16396.9, 0.974)
					(32784.3, 0.986)
					(65570.3, 0.995)
					(131120.0, 1.0)
					(136736.0, 1.0)
				};
				\addlegendentry{Cross-LSH}
				
				\addplot  [mark=triangle*, blue]
				coordinates 
				{
					(3.101, 0.016)
					(5.208, 0.026)
					(9.009, 0.039)
					(14.446, 0.056)
					(24.113, 0.08)
					(43.358, 0.104)
					(80.267, 0.149)
					(148.319, 0.218)
					(283.795, 0.269)
					(550.859, 0.328)
					(1080.94, 0.413)
					(2119.64, 0.493)
					(4184.42, 0.572)
					(8462.12, 0.68)
					(16843.8, 0.757)
					(33567.7, 0.855)
					(66356.2, 0.935)
					(131166.0, 0.999)
					(136736.0, 1.0)
				};
				\addlegendentry{Simple-LSH}
				\addplot 
				coordinates 
				{
					(37.184, 0.009)
					(45.012, 0.014)
					(68.517, 0.017)
					(81.707, 0.023)
					(114.103, 0.038)
					(151.083, 0.062)
					(202.558, 0.089)
					(279.606, 0.117)
					(405.205, 0.142)
					(743.523, 0.199)
					(1242.04, 0.244)
					(2229.79, 0.316)
					(4300.53, 0.395)
					(8428.5, 0.512)
					(16601.8, 0.649)
					(32999.2, 0.799)
					(65753.0, 0.929)
					(131155.0, 1.0)
					(136736.0, 1.0)
				};
				\addlegendentry{L2-ALSH}
				\addplot 
				coordinates 
				{
					(2.143, 0.007)
					(3.78, 0.012)
					(6.141, 0.02)
					(10.65, 0.032)
					(19.907, 0.053)
					(36.197, 0.076)
					(69.135, 0.112)
					(134.47, 0.152)
					(263.102, 0.204)
					(519.059, 0.282)
					(1031.54, 0.398)
					(2053.86, 0.529)
					(4103.02, 0.652)
					(8198.64, 0.797)
					(16391.3, 0.908)
					(32775.1, 0.967)
					(65545.6, 0.993)
					(131096.0, 1.0)
					(136736.0, 1.0)
				};
				\addlegendentry{Sign-ALSH}

			\end{axis}
			\end{tikzpicture}
	\end{subfigure}%
	\hspace{\figuregap}	
	\begin{subfigure}[b]{0.36\textwidth}
			\begin{tikzpicture}
			\begin{axis}[
			height=\figurewidth\linewidth,
			width=\linewidth,
			legend style={font=\fontsize{1}{1}\selectfont},
			grid=major,
			legend pos=south east,
			change x base,
			x SI prefix=mega,x unit=\-,
			xlabel=Probed Items ,
			xmin=0,xmax=1000000,
			ymin=0,ymax=1,
			]
			
				\addplot  [mark=square*, red]
				coordinates 
				{
					(2.322, 0.012)
					(3.551, 0.022)
					(6.743, 0.037)
					(12.483, 0.051)
					(22.075, 0.066)
					(44.082, 0.097)
					(82.993, 0.128)
					(155.21, 0.164)
					(298.75, 0.222)
					(674.045, 0.297)
					(1262.74, 0.382)
					(2713.62, 0.447)
					(5400.41, 0.529)
					(10290.0, 0.601)
					(20632.6, 0.674)
					(41640.8, 0.726)
					(83166.0, 0.766)
					(157142.0, 0.809)
					(300941.0, 0.851)
					(581439.0, 0.9)
					(1107000.0, 0.943)
					(2124240.0, 0.993)
					(2340370.0, 1.0)
				};
				\addlegendentry{Cross-LSH}
			
				\addplot  [mark=triangle*, blue]
				coordinates 
				{
					(3.622, 0.006)
					(5.984, 0.008)
					(11.19, 0.011)
					(19.62, 0.017)
					(31.879, 0.024)
					(58.118, 0.037)
					(123.01, 0.054)
					(236.526, 0.084)
					(436.86, 0.114)
					(790.832, 0.145)
					(1452.14, 0.202)
					(2703.28, 0.257)
					(5317.71, 0.332)
					(10197.5, 0.382)
					(19090.4, 0.434)
					(38067.6, 0.541)
					(77123.5, 0.625)
					(153656.0, 0.725)
					(299082.0, 0.795)
					(574074.0, 0.881)
					(1112160.0, 0.949)
					(2119620.0, 0.994)
					(2340370.0, 1.0)
				};
				\addlegendentry{Simple-LSH}
				\addplot 
				coordinates 
				{
					(1.698, 0.0)
					(3.89, 0.001)
					(10.404, 0.001)
					(26.536, 0.001)
					(45.01, 0.003)
					(81.143, 0.008)
					(217.876, 0.012)
					(337.848, 0.018)
					(567.186, 0.029)
					(1151.04, 0.041)
					(1954.29, 0.058)
					(4568.88, 0.088)
					(8943.98, 0.14)
					(15156.9, 0.189)
					(29258.6, 0.252)
					(54284.7, 0.337)
					(107756.0, 0.428)
					(196288.0, 0.529)
					(308760.0, 0.592)
					(601008.0, 0.678)
					(1103820.0, 0.803)
					(2110970.0, 0.958)
					(2340370.0, 1.0)
				};
				\addlegendentry{L2-ALSH}
				\addplot 
				coordinates 
				{
					(1.838, 0.002)
					(3.058, 0.005)
					(5.625, 0.013)
					(11.61, 0.016)
					(24.377, 0.019)
					(45.998, 0.034)
					(86.557, 0.045)
					(173.957, 0.063)
					(340.65, 0.082)
					(685.398, 0.113)
					(1485.52, 0.156)
					(2829.99, 0.207)
					(5448.0, 0.258)
					(10701.0, 0.335)
					(21322.1, 0.386)
					(39411.7, 0.447)
					(75644.2, 0.514)
					(142955.0, 0.593)
					(275229.0, 0.684)
					(540062.0, 0.758)
					(1068430.0, 0.847)
					(2104620.0, 0.973)
					(2340370.0, 1.0)
				};
				\addlegendentry{Sign-ALSH}
				
			\end{axis}
			\end{tikzpicture}	
	\end{subfigure}%
	\caption{Probed item-recall of top-1 item comparison between Cross-LSH and existing algorithms under a code length of 32 (best viewed in color). From left to right, the datasets are Netflix, Yahoo!Music and ImageNet, respectively.} 
	\label{top1-bit32:Cross comparision}
\end{figure*}
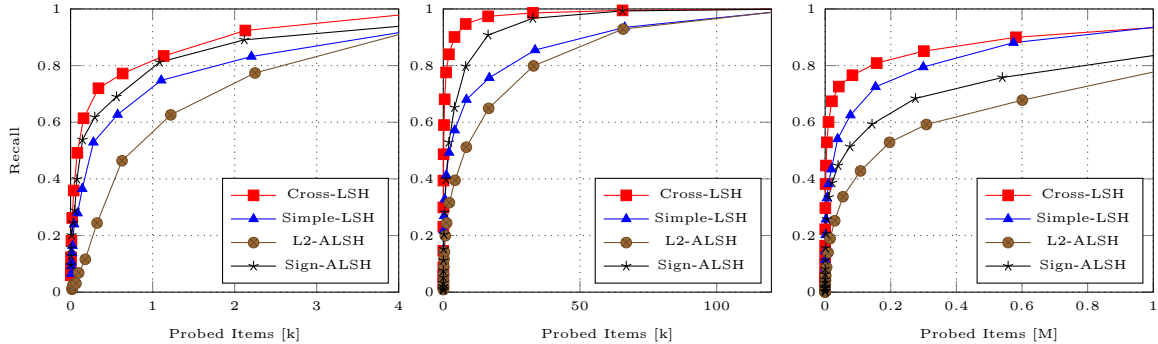  

\begin{figure*}[t]
		\begin{subfigure}[b]{\figurepercentperrow\textwidth}
			\begin{tikzpicture}
			\begin{axis}[
			height=\figurewidth\linewidth,
			width=\linewidth,
			grid=major,
			legend pos=south east,
			legend style={inner xsep=0pt, inner ysep=0pt, font=\fontsize{1}{1}\selectfont},
			change x base,
			x SI prefix=kilo,x unit=\-,
			xlabel=Probed Items , ylabel=Recall,
			xmin=0,xmax=2000,
			ymin=0.2,ymax=1,
			mark size=2.0pt,
			]

					\addplot 
					coordinates 
					{
						(1.956, 0.06)
						(3.367, 0.083)
						(5.745, 0.123)
						(10.746, 0.182)
						(19.911, 0.262)
						(38.253, 0.359)
						(82.004, 0.492)
						(158.274, 0.614)
						(339.856, 0.72)
						(634.608, 0.772)
						(1135.33, 0.834)
						(2130.34, 0.924)
						(4123.87, 0.982)
						(8201.29, 1.0)
						(16385.4, 1.0)
						(17770.0, 1.0)
					};
					\addlegendentry{Cross-LSH}
					\addplot 
					coordinates 
					{
						(2.881, 0.09)
						(5.052, 0.15)
						(9.219, 0.229)
						(14.411, 0.322)
						(23.546, 0.429)
						(45.925, 0.564)
						(77.161, 0.63)
						(139.937, 0.744)
						(261.963, 0.867)
						(516.747, 0.938)
						(1026.74, 0.996)
						(2050.04, 1.0)
						(4097.76, 1.0)
						(8193.25, 1.0)
						(16384.3, 1.0)
						(17770.0, 1.0)
					};
					\addlegendentry{Cross-LSH-NR}
			\end{axis}
			\end{tikzpicture}
	\end{subfigure}%
	\hspace{\figuregap}
	\begin{subfigure}[b]{\figurepercentperrow\textwidth}
	\begin{tikzpicture}
	\begin{axis}[
	height=\figurewidth\linewidth,
	width=\linewidth,
	legend pos=south east,
	legend style={inner xsep=0pt, inner ysep=0pt, font=\fontsize{1}{1}\selectfont},
	grid=major,
	legend pos=south east,
	change x base,
	x SI prefix=kilo,x unit=\-,
	xlabel=Probed Items ,
	xmin=0,xmax=2000,
	ymin=0.2,ymax=1,
	]
		\addplot 
		coordinates 
		{
			(17.786, 0.01)
			(19.472, 0.014)
			(23.312, 0.017)
			(29.185, 0.024)
			(36.983, 0.028)
			(65.372, 0.031)
			(98.869, 0.068)
			(179.253, 0.116)
			(322.414, 0.244)
			(627.332, 0.464)
			(1221.68, 0.626)
			(2248.05, 0.774)
			(4176.56, 0.923)
			(8351.56, 0.988)
			(16390.6, 1.0)
			(17770.0, 1.0)
		};
		\addlegendentry{L2-ALSH}
		\addplot 
		coordinates 
		{
			(9.596, 0.112)
			(11.478, 0.126)
			(13.391, 0.132)
			(17.439, 0.191)
			(25.315, 0.249)
			(41.422, 0.301)
			(74.713, 0.482)
			(139.739, 0.688)
			(267.869, 0.843)
			(533.35, 0.936)
			(1046.19, 0.963)
			(2071.32, 0.978)
			(4113.52, 0.993)
			(8201.93, 0.997)
			(16386.9, 1.0)
			(17770.0, 1.0)
		};
		\addlegendentry{L2-ALSH-NR}
	\end{axis}
	\end{tikzpicture}
	\end{subfigure}%
	\hspace{\figuregap}
	\begin{subfigure}[b]{\figurepercentperrow\textwidth}
	\begin{tikzpicture}
	\begin{axis}[
	height=\figurewidth\linewidth,
	width=\linewidth,
	legend pos=south east,
	legend style={inner xsep=0pt, inner ysep=0pt, font=\fontsize{1}{1}\selectfont},
	grid=major,
	change x base,
	x SI prefix=kilo,x unit=\-,
	xlabel=Probed Items,
	xmin=0,xmax=2000,
	ymin=0.2,ymax=1,
	]
	
		\addplot 
		coordinates 
		{
			(9.294, 0.065)
			(11.16, 0.092)
			(13.082, 0.106)
			(17.92, 0.138)
			(24.575, 0.164)
			(43.384, 0.24)
			(85.428, 0.279)
			(142.613, 0.365)
			(278.485, 0.529)
			(573.04, 0.627)
			(1105.26, 0.748)
			(2204.51, 0.832)
			(4228.29, 0.927)
			(8281.92, 0.981)
			(16539.7, 1.0)
			(17770.0, 1.0)
		};
		\addlegendentry{Simple-LSH}
		\addplot 
		coordinates 
		{
			(2.874, 0.115)
			(3.98, 0.131)
			(6.786, 0.171)
			(11.176, 0.271)
			(20.718, 0.376)
			(35.11, 0.505)
			(67.016, 0.667)
			(131.15, 0.804)
			(259.089, 0.927)
			(515.774, 0.982)
			(1028.72, 0.998)
			(2050.89, 1.0)
			(4099.11, 1.0)
			(8195.06, 1.0)
			(16386.4, 1.0)
			(17770.0, 1.0)
		};
		\addlegendentry{Simple-LSH-NR}
		
	\end{axis}
	\end{tikzpicture}
	\end{subfigure}%	
	\hspace{\figuregap}	
	\begin{subfigure}[b]{\figurepercentperrow\textwidth}
		\begin{tikzpicture}
		\begin{axis}[
		height=\figurewidth\linewidth,
		width=\linewidth,
		legend pos=south east,
		legend style={inner xsep=0pt, inner ysep=0pt, font=\fontsize{1}{1}\selectfont},
		grid=major,
		legend pos=south east,
		change x base,
		x SI prefix=kilo,x unit=\-,
		xlabel=Probed Items ,
		xmin=0,xmax=2000,
		ymin=0.2,ymax=1,
		]
			\addplot 
			coordinates 
			{
				(2.552, 0.096)
				(4.077, 0.128)
				(6.804, 0.156)
				(13.211, 0.201)
				(32.038, 0.245)
				(50.098, 0.289)
				(77.63, 0.399)
				(145.34, 0.538)
				(294.815, 0.618)
				(559.7, 0.69)
				(1085.42, 0.812)
				(2117.53, 0.891)
				(4211.13, 0.944)
				(8295.86, 0.99)
				(16395.4, 1.0)
				(17770.0, 1.0)
			};
			\addlegendentry{Sign-ALSH}
			\addplot 
			coordinates 
			{
				(12.254, 0.109)
				(14.642, 0.134)
				(19.105, 0.176)
				(24.129, 0.223)
				(31.656, 0.291)
				(53.499, 0.476)
				(91.277, 0.708)
				(137.005, 0.805)
				(259.858, 0.908)
				(513.986, 0.974)
				(1025.12, 0.994)
				(2048.71, 0.998)
				(4096.84, 1.0)
				(8192.53, 1.0)
				(16384.5, 1.0)
				(17770.0, 1.0)
				
			};
			\addlegendentry{Sign-ALSH-NR}
			
		\end{axis}
		\end{tikzpicture}
	\end{subfigure}%
	
		\begin{subfigure}[b]{\figurepercentperrow\textwidth}
			\begin{tikzpicture}
			\begin{axis}[
			height=\figurewidth\linewidth,
			width=\linewidth,
			grid=major,
			legend pos=south east,
			legend style={inner xsep=0pt, inner ysep=0pt, font=\fontsize{1}{1}\selectfont},
			change x base,
			x SI prefix=kilo,x unit=\-,
			xmin=0,xmax=20000,
			ymin=0.2,ymax=1,
			xlabel=Probed Items , ylabel=Recall,
			]
				\addplot 
				coordinates 
				{
					(1.293, 0.031)
					(2.462, 0.056)
					(4.737, 0.088)
					(9.107, 0.146)
					(17.636, 0.231)
					(34.098, 0.299)
					(66.401, 0.394)
					(131.114, 0.487)
					(260.59, 0.59)
					(517.042, 0.681)
					(1029.51, 0.776)
					(2054.43, 0.84)
					(4102.73, 0.901)
					(8200.81, 0.947)
					(16396.9, 0.974)
					(32784.3, 0.986)
					(65570.3, 0.995)
					(131120.0, 1.0)
					(136736.0, 1.0)
				};
				\addlegendentry{Cross-LSH}
				\addplot 
				coordinates 
				{
					(1.537, 0.028)
					(2.765, 0.043)
					(5.181, 0.078)
					(9.373, 0.123)
					(17.5, 0.176)
					(33.692, 0.269)
					(66.263, 0.382)
					(131.39, 0.514)
					(260.313, 0.641)
					(516.49, 0.766)
					(1029.41, 0.878)
					(2052.97, 0.932)
					(4099.2, 0.976)
					(8193.2, 0.994)
					(16385.0, 1.0)
					(32768.9, 1.0)
					(65536.8, 1.0)
					(131083.0, 1.0)
					(136736.0, 1.0)
				};
				\addlegendentry{Cross-LSH-NR}
			\end{axis}
			\end{tikzpicture}
	\end{subfigure}%
	\hspace{\figuregap}
	\begin{subfigure}[b]{\figurepercentperrow\textwidth}
	\begin{tikzpicture}
	\begin{axis}[
	height=\figurewidth\linewidth,
	width=\linewidth,
	legend pos=south east,
	legend style={inner xsep=0pt, inner ysep=0pt, font=\fontsize{1}{1}\selectfont},
	grid=major,
	change x base,
	x SI prefix=kilo,x unit=\-,
	xlabel=Probed Items,
	xmin=0,xmax=20000,
	ymin=0.2,ymax=1,
	]
			\addplot 
			coordinates 
			{
				(37.184, 0.009)
				(45.012, 0.014)
				(68.517, 0.017)
				(81.707, 0.023)
				(114.103, 0.038)
				(151.083, 0.062)
				(202.558, 0.089)
				(279.606, 0.117)
				(405.205, 0.142)
				(743.523, 0.199)
				(1242.04, 0.244)
				(2229.79, 0.316)
				(4300.53, 0.395)
				(8428.5, 0.512)
				(16601.8, 0.649)
				(32999.2, 0.799)
				(65753.0, 0.929)
				(131155.0, 1.0)
				(136736.0, 1.0)
			};
			\addlegendentry{L2-ALSH}
			\addplot 
			coordinates 
			{
				(3.086, 0.015)
				(4.719, 0.027)
				(7.329, 0.038)
				(12.796, 0.055)
				(22.726, 0.091)
				(39.009, 0.146)
				(73.836, 0.212)
				(143.679, 0.291)
				(268.473, 0.376)
				(525.072, 0.512)
				(1034.06, 0.648)
				(2056.99, 0.778)
				(4101.15, 0.865)
				(8195.94, 0.922)
				(16387.7, 0.955)
				(32771.8, 0.975)
				(65542.6, 0.993)
				(131076.0, 1.0)
				(136736.0, 1.0)
			};
			\addlegendentry{L2-ALSH-NR}

	\end{axis}
	\end{tikzpicture}
\end{subfigure}%
	\hspace{\figuregap}
	\begin{subfigure}[b]{\figurepercentperrow\textwidth}
	\begin{tikzpicture}
	\begin{axis}[
	height=\figurewidth\linewidth,
	width=\linewidth,
	legend pos=south east,
	legend style={inner xsep=0pt, inner ysep=0pt, font=\fontsize{1}{1}\selectfont},
	grid=major,
	change x base,
	x SI prefix=kilo,x unit=\-,
	xlabel=Probed Items , 
	xmin=0,xmax=20000,
	ymin=0.2,ymax=1,
	]
	
			\addplot 
			coordinates 
			{
				(3.101, 0.016)
				(5.208, 0.026)
				(9.009, 0.039)
				(14.446, 0.056)
				(24.113, 0.08)
				(43.358, 0.104)
				(80.267, 0.149)
				(148.319, 0.218)
				(283.795, 0.269)
				(550.859, 0.328)
				(1080.94, 0.413)
				(2119.64, 0.493)
				(4184.42, 0.572)
				(8462.12, 0.68)
				(16843.8, 0.757)
				(33567.7, 0.855)
				(66356.2, 0.935)
				(131166.0, 0.999)
				(136736.0, 1.0)
			};
			\addlegendentry{Simple-LSH}
			\addplot 
			coordinates 
			{
				(1.835, 0.036)
				(3.07, 0.052)
				(5.575, 0.084)
				(9.826, 0.124)
				(18.151, 0.184)
				(34.665, 0.242)
				(68.004, 0.336)
				(132.085, 0.443)
				(261.107, 0.564)
				(517.067, 0.699)
				(1028.87, 0.846)
				(2052.39, 0.936)
				(4098.48, 0.974)
				(8193.31, 0.99)
				(16385.2, 0.993)
				(32769.0, 0.998)
				(65536.7, 0.999)
				(131079.0, 1.0)
				(136736.0, 1.0)
			};
			\addlegendentry{Simple-LSH-NR}		
	\end{axis}
	\end{tikzpicture}
\end{subfigure}%
	\hspace{\figuregap}
	\begin{subfigure}[b]{\figurepercentperrow\textwidth}
		\begin{tikzpicture}
		\begin{axis}[
		height=\figurewidth\linewidth,
		width=\linewidth,
		legend pos=south east,
		legend style={inner xsep=0pt, inner ysep=0pt, font=\fontsize{1}{1}\selectfont},
		grid=major,
		change x base,
		x SI prefix=kilo,x unit=\-,
		xlabel=Probed Items ,
		xmin=0,xmax=20000,
		ymin=0.2,ymax=1,
		]
			\addplot 
			coordinates 
			{
				(2.143, 0.007)
				(3.78, 0.012)
				(6.141, 0.02)
				(10.65, 0.032)
				(19.907, 0.053)
				(36.197, 0.076)
				(69.135, 0.112)
				(134.47, 0.152)
				(263.102, 0.204)
				(519.059, 0.282)
				(1031.54, 0.398)
				(2053.86, 0.529)
				(4103.02, 0.652)
				(8198.64, 0.797)
				(16391.3, 0.908)
				(32775.1, 0.967)
				(65545.6, 0.993)
				(131096.0, 1.0)
				(136736.0, 1.0)
			};
			\addlegendentry{Sign-ALSH}
			\addplot 
			coordinates 
			{
				(2.666, 0.021)
				(4.087, 0.034)
				(7.052, 0.051)
				(11.258, 0.075)
				(20.025, 0.106)
				(38.831, 0.157)
				(72.707, 0.215)
				(138.358, 0.3)
				(263.477, 0.42)
				(518.857, 0.543)
				(1029.27, 0.694)
				(2051.8, 0.856)
				(4098.14, 0.961)
				(8193.54, 0.988)
				(16385.1, 0.997)
				(32769.1, 0.999)
				(65537.0, 1.0)
				(131084.0, 1.0)
				(136736.0, 1.0)
			};
			\addlegendentry{Sign-ALSH-NR}
		\end{axis}
		\end{tikzpicture}
	\end{subfigure}%
	
		\begin{subfigure}[b]{\figurepercentperrow\textwidth}
			\begin{tikzpicture}
			\begin{axis}[
				height=\figurewidth\linewidth,
				width=\linewidth,
				grid=major,
				legend pos=south east,
				legend style={inner xsep=0pt, inner ysep=0pt, font=\fontsize{1}{1}\selectfont},
				change x base,
				x SI prefix=mega,x unit=\-,
				xmin=0,xmax=400000,
				ymin=0.2,ymax=1,
				xlabel=Probed Items , ylabel=Recall,
			]
			
				\addplot 
				coordinates 
				{
					(2.322, 0.012)
					(3.551, 0.022)
					(6.743, 0.037)
					(12.483, 0.051)
					(22.075, 0.066)
					(44.082, 0.097)
					(82.993, 0.128)
					(155.21, 0.164)
					(298.75, 0.222)
					(674.045, 0.297)
					(1262.74, 0.382)
					(2713.62, 0.447)
					(5400.41, 0.529)
					(10290.0, 0.601)
					(20632.6, 0.674)
					(41640.8, 0.726)
					(83166.0, 0.766)
					(157142.0, 0.809)
					(300941.0, 0.851)
					(581439.0, 0.9)
					(1107000.0, 0.943)
					(2124240.0, 0.993)
					(2340370.0, 1.0)
				};
				\addlegendentry{Cross-LSH}
				\addplot 
				coordinates 
				{
					(1.603, 0.008)
					(3.171, 0.013)
					(5.933, 0.021)
					(10.55, 0.036)
					(20.033, 0.063)
					(37.074, 0.089)
					(72.621, 0.134)
					(141.37, 0.188)
					(274.065, 0.253)
					(535.697, 0.32)
					(1058.8, 0.427)
					(2087.64, 0.561)
					(4127.42, 0.68)
					(8213.85, 0.791)
					(16402.4, 0.864)
					(32796.1, 0.902)
					(131073.0, 0.997)
					(262145.0, 0.999)
					(524288.0, 1.0)
					(1048580.0, 1.0)
					(2097150.0, 1.0)
					(2340370.0, 1.0)
				};
				\addlegendentry{Cross-LSH-NR}

			\end{axis}
			\end{tikzpicture}	
	\end{subfigure}%
	\hspace{\figuregap}
	\begin{subfigure}[b]{\figurepercentperrow\textwidth}
		\begin{tikzpicture}
		\begin{axis}[
		height=\figurewidth\linewidth,
		width=\linewidth,
		grid=major,
		legend pos=south east,
		legend style={inner xsep=0pt, inner ysep=0pt, font=\fontsize{1}{1}\selectfont},
		change x base,
		x SI prefix=mega,x unit=\-,
		xlabel=Probed Items ,
		xmin=0,xmax=400000,
		ymin=0.2,ymax=1,
		]
		
			\addplot 
			coordinates 
			{
				(1.698, 0.0)
				(3.89, 0.001)
				(10.404, 0.001)
				(26.536, 0.001)
				(45.01, 0.003)
				(81.143, 0.008)
				(217.876, 0.012)
				(337.848, 0.018)
				(567.186, 0.029)
				(1151.04, 0.041)
				(1954.29, 0.058)
				(4568.88, 0.088)
				(8943.98, 0.14)
				(15156.9, 0.189)
				(29258.6, 0.252)
				(54284.7, 0.337)
				(107756.0, 0.428)
				(196288.0, 0.529)
				(308760.0, 0.592)
				(601008.0, 0.678)
				(1103820.0, 0.803)
				(2110970.0, 0.958)
				(2340370.0, 1.0)
			};
			\addlegendentry{L2-ALSH}
			\addplot 
			coordinates 
			{
				(2.72, 0.005)
				(4.605, 0.01)
				(7.125, 0.01)
				(12.01, 0.02)
				(19.845, 0.025)
				(37.605, 0.045)
				(72.71, 0.065)
				(139.66, 0.09)
				(272.875, 0.105)
				(529.6, 0.17)
				(1047.91, 0.23)
				(2101.01, 0.345)
				(4145.02, 0.41)
				(8296.3, 0.545)
				(16469.7, 0.68)
				(32928.7, 0.815)
				(65556.0, 0.9)
				(131081.0, 0.965)
				(262150.0, 0.97)
				(524290.0, 0.97)
				(1048580.0, 0.98)
			};
			\addlegendentry{L2-ALSH-NR}
			
		\end{axis}
		\end{tikzpicture}	
	\end{subfigure}%
	\hspace{\figuregap}	
	\begin{subfigure}[b]{\figurepercentperrow\textwidth}
	\begin{tikzpicture}
	\begin{axis}[
		height=\figurewidth\linewidth,
		width=\linewidth,
		grid=major,
		legend pos=south east,
		legend style={inner xsep=0pt, inner ysep=0pt, font=\fontsize{1}{1}\selectfont},
		change x base,
		x SI prefix=mega,x unit=\-,
		xlabel=Probed Items,
		xmin=0,xmax=400000,
		ymin=0.2,ymax=1,
	]
			\addplot 
			coordinates 
			{
				(3.622, 0.006)
				(5.984, 0.008)
				(11.19, 0.011)
				(19.62, 0.017)
				(31.879, 0.024)
				(58.118, 0.037)
				(123.01, 0.054)
				(236.526, 0.084)
				(436.86, 0.114)
				(790.832, 0.145)
				(1452.14, 0.202)
				(2703.28, 0.257)
				(5317.71, 0.332)
				(10197.5, 0.382)
				(19090.4, 0.434)
				(38067.6, 0.541)
				(77123.5, 0.625)
				(153656.0, 0.725)
				(299082.0, 0.795)
				(574074.0, 0.881)
				(1112160.0, 0.949)
				(2119620.0, 0.994)
				(2340370.0, 1.0)
			};
			\addlegendentry{Simple-LSH}
			\addplot 
			coordinates 
			{
				(1.593, 0.007)
				(2.909, 0.007)
				(5.503, 0.011)
				(10.321, 0.016)
				(19.587, 0.027)
				(37.364, 0.034)
				(70.64, 0.049)
				(140.299, 0.073)
				(270.518, 0.103)
				(530.901, 0.156)
				(1051.48, 0.223)
				(2080.38, 0.32)
				(4132.92, 0.447)
				(8222.22, 0.599)
				(16407.7, 0.75)
				(32789.8, 0.865)
				(65561.1, 0.921)
				(131073.0, 0.989)
				(262144.0, 0.997)
				(524289.0, 0.999)
				(1048580.0, 1.0)
				(2097160.0, 1.0)
				(2340370.0, 1.0)
			};
			\addlegendentry{Simple-LSH-NR}
	\end{axis}
	\end{tikzpicture}	
	\end{subfigure}%
	\hspace{\figuregap}	
	\begin{subfigure}[b]{\figurepercentperrow\textwidth}
			\begin{tikzpicture}
			\begin{axis}[
			height=\figurewidth\linewidth,
			width=\linewidth,
			legend pos=south east,
			legend style={inner xsep=0pt, inner ysep=0pt, font=\fontsize{1}{1}\selectfont},
			grid=major,
			change x base,
			x SI prefix=mega,x unit=\-,
			xlabel=Probed Items ,
			xmin=0,xmax=400000,
			ymin=0.2,ymax=1,
			]
					\addplot 
					coordinates 
					{
						(1.838, 0.002)
						(3.058, 0.005)
						(5.625, 0.013)
						(11.61, 0.016)
						(24.377, 0.019)
						(45.998, 0.034)
						(86.557, 0.045)
						(173.957, 0.063)
						(340.65, 0.082)
						(685.398, 0.113)
						(1485.52, 0.156)
						(2829.99, 0.207)
						(5448.0, 0.258)
						(10701.0, 0.335)
						(21322.1, 0.386)
						(39411.7, 0.447)
						(75644.2, 0.514)
						(142955.0, 0.593)
						(275229.0, 0.684)
						(540062.0, 0.758)
						(1068430.0, 0.847)
						(2104620.0, 0.973)
						(2340370.0, 1.0)
					};
					\addlegendentry{Sign-ALSH}
					\addplot 
					coordinates 
					{
						(1.638, 0.001)
						(3.031, 0.003)
						(5.574, 0.007)
						(10.232, 0.011)
						(19.918, 0.018)
						(39.653, 0.028)
						(76.367, 0.053)
						(144.581, 0.097)
						(278.829, 0.136)
						(546.643, 0.2)
						(1072.3, 0.307)
						(2114.44, 0.428)
						(4164.42, 0.552)
						(8262.58, 0.696)
						(16460.4, 0.807)
						(32829.2, 0.889)
						(65580.5, 0.932)
						(131077.0, 0.988)
						(262146.0, 0.996)
						(524289.0, 0.999)
						(1048580.0, 1.0)
						(2097150.0, 1.0)
						(2340370.0, 1.0)
					};
					\addlegendentry{Sign-ALSH-NR}
			\end{axis}
			\end{tikzpicture}	
		\end{subfigure}%
	
	\caption{Probed item-recall of top-1 item comparison between the original meta algorithms and their norm-range versions under a code length of 32. From top row to bottom row, the datasets are Netflix, Yahoo!Music and ImageNet, respectively.} 
	\label{top1-bit32:Norm-Range comparision}
\end{figure*}
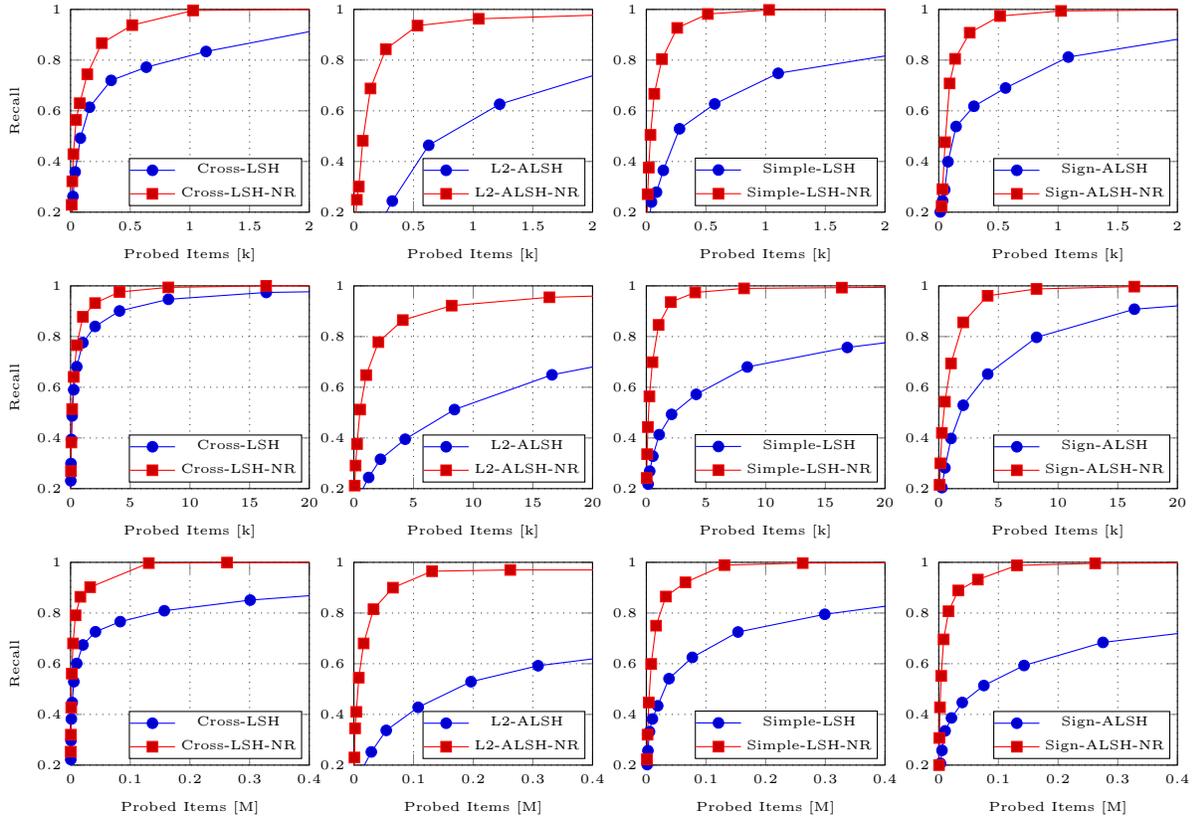

\begin{figure*}[t]
		\begin{subfigure}[b]{0.36\textwidth}
			\begin{tikzpicture}
			\begin{axis}[
			height=\figurewidth\linewidth,
			width=\linewidth,
			legend style={font=\fontsize{1}{1}\selectfont},
			grid=major,
			legend pos=south east,
			change x base,
			x SI prefix=kilo,x unit=\-,
			xlabel=Probed Items , ylabel=Recall,
			xmin=0,xmax=4000,
			ymin=0,ymax=1,
			]
					
				\addplot [mark=square*, red]
				coordinates 
				{
					(1.956, 0.0273)
					(3.367, 0.0435)
					(5.745, 0.0699998)
					(10.746, 0.1077)
					(19.911, 0.1596)
					(38.253, 0.2359)
					(82.004, 0.3382)
					(158.274, 0.449701)
					(339.856, 0.5641)
					(634.608, 0.6435)
					(1135.33, 0.743901)
					(2130.34, 0.881201)
					(4123.87, 0.976401)
					(8201.29, 0.9994)
					(16385.4, 1.0)
					(17770.0, 1.0)
				};
				\addlegendentry{Cross-LSH}
				
				\addplot  [mark=triangle*, blue]
				coordinates 
				{
					(8.79, 0.0253)
					(10.323, 0.029)
					(12.904, 0.0392)
					(23.688, 0.0544999)
					(42.275, 0.0823999)
					(63.908, 0.1051)
					(111.513, 0.18)
					(163.909, 0.2152)
					(322.068, 0.3164)
					(564.366, 0.4197)
					(1065.0, 0.639)
					(2101.27, 0.821401)
					(4163.45, 0.924501)
					(8253.57, 0.9865)
					(16405.5, 0.9995)
					(17770.0, 1.0)
				};
				\addlegendentry{Simple-LSH}
				
				\addplot 
				coordinates 
				{
					(17.786, 0.0132)
					(19.472, 0.015)
					(23.312, 0.0169)
					(29.185, 0.0203)
					(36.983, 0.0219)
					(65.372, 0.0236)
					(98.869, 0.0462)
					(179.253, 0.1284)
					(322.414, 0.2469)
					(627.332, 0.4507)
					(1221.68, 0.595899)
					(2248.05, 0.7394)
					(4176.56, 0.875101)
					(8351.56, 0.9701)
					(16390.6, 0.9999)
					(17770.0, 1.0)
				};
				\addlegendentry{L2-ALSH}

				\addplot 
				coordinates 
				{
					(8.265, 0.0429)
					(10.006, 0.0729)
					(17.613, 0.1086)
					(28.966, 0.1368)
					(42.946, 0.1926)
					(61.329, 0.2395)
					(99.26, 0.3102)
					(174.417, 0.4132)
					(323.457, 0.5277)
					(563.655, 0.6152)
					(1077.82, 0.7147)
					(2098.84, 0.842301)
					(4152.46, 0.928301)
					(8246.23, 0.9865)
					(16402.4, 1.0)
					(17770.0, 1.0)
				};
				\addlegendentry{Sign-ALSH}

			\end{axis}
			\end{tikzpicture}
	\end{subfigure}%
	\hspace{\figuregap}	
	\begin{subfigure}[b]{0.36\textwidth}
			\begin{tikzpicture}
			\begin{axis}[
			height=\figurewidth\linewidth,
			width=\linewidth,
			legend style={font=\fontsize{1}{1}\selectfont},
			grid=major,
			legend pos=south east,
			change x base,
			x SI prefix=kilo,x unit=\-,
			xlabel=Probed Items ,
			xmin=0,xmax=120000,
			ymin=0,ymax=1,
			]
			
				\addplot [mark=square*, red]
				coordinates 
				{
					(1.293, 0.0186)
					(2.462, 0.0327001)
					(4.737, 0.0527999)
					(9.107, 0.0881998)
					(17.636, 0.1395)
					(34.098, 0.1965)
					(66.401, 0.2758)
					(131.114, 0.3734)
					(260.59, 0.4728)
					(517.042, 0.5806)
					(1029.51, 0.6851)
					(2054.43, 0.7787)
					(4102.73, 0.860801)
					(8200.81, 0.921301)
					(16396.9, 0.962201)
					(32784.3, 0.9842)
					(65570.3, 0.9955)
					(131120.0, 0.9999)
					(136736.0, 1.0)
				};
				\addlegendentry{Cross-LSH}
				
				\addplot  [mark=triangle*, blue]
				coordinates 
				{
					(1.601, 0.0082)
					(3.024, 0.0143)
					(6.312, 0.0219)
					(11.367, 0.0322)
					(20.84, 0.0488)
					(39.239, 0.0690999)
					(73.361, 0.0940998)
					(148.204, 0.1409)
					(288.701, 0.1973)
					(557.718, 0.2676)
					(1083.85, 0.3476)
					(2126.98, 0.4415)
					(4197.96, 0.544)
					(8317.42, 0.657299)
					(16518.1, 0.7591)
					(32923.3, 0.871501)
					(65685.6, 0.960701)
					(131149.0, 0.9995)
					(136736.0, 1.0)
				};
				\addlegendentry{Simple-LSH}
				
				\addplot 
				coordinates 
				{
					(37.184, 0.008)
					(45.012, 0.0132)
					(68.517, 0.0175)
					(81.707, 0.0235)
					(114.103, 0.0353)
					(151.083, 0.0590999)
					(202.558, 0.0823999)
					(279.606, 0.1053)
					(405.205, 0.1321)
					(743.523, 0.1882)
					(1242.04, 0.2357)
					(2229.79, 0.3057)
					(4300.53, 0.4041)
					(8428.5, 0.5214)
					(16601.8, 0.647)
					(32999.2, 0.791401)
					(65753.0, 0.924801)
					(131155.0, 0.9987)
					(136736.0, 1.0)
				};
				\addlegendentry{L2-ALSH}

				\addplot 
				coordinates 
				{
					(2.659, 0.0082)
					(4.077, 0.014)
					(7.567, 0.0212)
					(12.545, 0.0322)
					(21.04, 0.0485999)
					(37.739, 0.0751999)
					(72.54, 0.1114)
					(137.86, 0.1584)
					(268.414, 0.2326)
					(529.784, 0.317701)
					(1053.23, 0.4221)
					(2081.19, 0.5372)
					(4133.63, 0.6628)
					(8238.43, 0.773901)
					(16445.3, 0.866501)
					(32845.8, 0.938301)
					(65613.7, 0.981701)
					(131111.0, 0.9998)
					(136736.0, 1.0)
				};
				\addlegendentry{Sign-ALSH}
			
			\end{axis}
			\end{tikzpicture}
	\end{subfigure}%
	\hspace{\figuregap}	
	\begin{subfigure}[b]{0.36\textwidth}
			\begin{tikzpicture}
			\begin{axis}[
			height=\figurewidth\linewidth,
			width=\linewidth,
			legend style={font=\fontsize{1}{1}\selectfont},
			grid=major,
			legend pos=south east,
			change x base,
			x SI prefix=mega,x unit=\-,
			xlabel=Probed Items ,
			xmin=0,xmax=1000000,
			ymin=0,ymax=1,
			]
			
				\addplot [mark=square*, red]
				coordinates 
				{
					(2.322, 0.004)
					(3.551, 0.0082)
					(6.743, 0.0147)
					(12.483, 0.0221)
					(22.075, 0.0363)
					(44.082, 0.0598999)
					(82.993, 0.0902998)
					(155.21, 0.1269)
					(298.75, 0.1793)
					(674.045, 0.2448)
					(1262.74, 0.316)
					(2713.62, 0.3989)
					(5400.41, 0.4892)
					(10290.0, 0.5682)
					(20632.6, 0.642901)
					(41640.8, 0.710401)
					(83166.0, 0.7642)
					(157142.0, 0.8063)
					(300941.0, 0.846801)
					(581439.0, 0.8892)
					(1107000.0, 0.9391)
					(2124240.0, 0.9903)
					(2340370.0, 1.0)
				};
				\addlegendentry{Cross-LSH}
				
				\addplot  [mark=triangle*, blue]
				coordinates 
				{
					(1.248, 0.0014)
					(2.626, 0.0031)
					(7.65, 0.0052)
					(18.419, 0.0113)
					(34.384, 0.0179)
					(59.331, 0.0283)
					(134.612, 0.0488)
					(224.409, 0.0724999)
					(393.673, 0.1048)
					(836.196, 0.1386)
					(1736.51, 0.1868)
					(4192.8, 0.2368)
					(6827.95, 0.281601)
					(11745.3, 0.348)
					(24404.1, 0.4427)
					(50077.5, 0.5212)
					(85367.0, 0.5799)
					(180067.0, 0.6641)
					(348969.0, 0.767601)
					(750185.0, 0.840402)
					(1283770.0, 0.897301)
					(2137650.0, 0.977801)
					(2340370.0, 1.0)
				};
				\addlegendentry{Simple-LSH}
				
				\addplot 
				coordinates 
				{
					(1.698, 0.0006)
					(3.89, 0.0009)
					(10.404, 0.0012)
					(26.536, 0.0025)
					(45.01, 0.0045)
					(81.143, 0.0085)
					(217.876, 0.0122)
					(337.848, 0.0175)
					(567.186, 0.0266)
					(1151.04, 0.04)
					(1954.29, 0.0564999)
					(4568.88, 0.0848998)
					(8943.98, 0.1236)
					(15156.9, 0.1678)
					(29258.6, 0.2266)
					(54284.7, 0.3101)
					(107756.0, 0.4023)
					(196288.0, 0.4918)
					(308760.0, 0.5684)
					(601008.0, 0.662)
					(1103820.0, 0.7923)
					(2110970.0, 0.958601)
					(2340370.0, 1.0)
				};
				\addlegendentry{L2-ALSH}

				\addplot 
				coordinates 
				{
					(1.305, 0.0017)
					(2.433, 0.0032)
					(4.604, 0.0049)
					(9.198, 0.009)
					(19.078, 0.015)
					(36.799, 0.0246)
					(75.222, 0.0391)
					(159.582, 0.0603999)
					(313.99, 0.0896998)
					(661.117, 0.1361)
					(1344.11, 0.1874)
					(3077.46, 0.255)
					(5459.62, 0.3189)
					(12122.8, 0.3893)
					(27268.7, 0.4792)
					(63670.9, 0.5682)
					(111247.0, 0.641999)
					(241682.0, 0.6907)
					(406006.0, 0.7448)
					(714642.0, 0.8243)
					(1224640.0, 0.895201)
					(2150270.0, 0.970601)
					(2340370.0, 1.0)
				};
				\addlegendentry{Sign-ALSH}

			\end{axis}
			\end{tikzpicture}	
	\end{subfigure}%
	\caption{Probed item-recall of top-10 items comparison between Cross-LSH and existing algorithms under a code length of 32 (best viewed in color). From left to right, the datasets are Netflix, Yahoo!Music and ImageNet, respectively.} 
	\label{top10-bit32:Cross comparision}
\end{figure*}
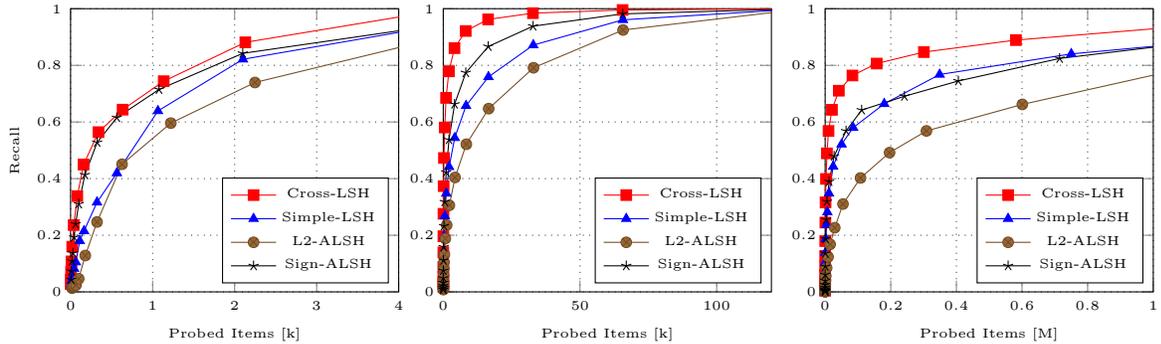  

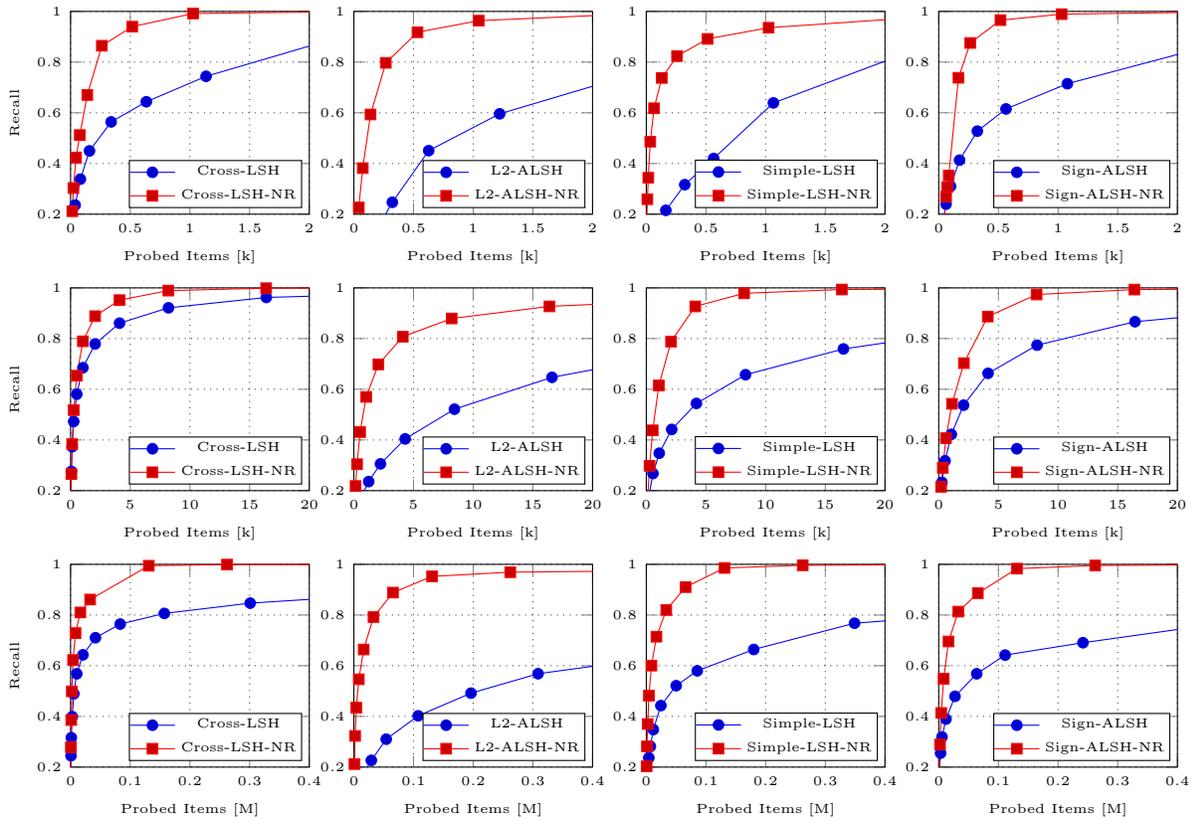
\begin{figure*}[t]
		\begin{subfigure}[b]{\figurepercentperrow\textwidth}
			\begin{tikzpicture}
			\begin{axis}[
			height=\figurewidth\linewidth,
			width=\linewidth,
			grid=major,
			legend pos=south east,
			legend style={inner xsep=0pt, inner ysep=0pt, font=\fontsize{1}{1}\selectfont},
			change x base,
			x SI prefix=kilo,x unit=\-,
			xlabel=Probed Items , ylabel=Recall,
			xmin=0,xmax=2000,
			ymin=0.2,ymax=1,
			mark size=2.0pt,
			]

					\addplot 
					coordinates 
					{
						(1.956, 0.0273)
						(3.367, 0.0435)
						(5.745, 0.0699998)
						(10.746, 0.1077)
						(19.911, 0.1596)
						(38.253, 0.2359)
						(82.004, 0.3382)
						(158.274, 0.449701)
						(339.856, 0.5641)
						(634.608, 0.6435)
						(1135.33, 0.743901)
						(2130.34, 0.881201)
						(4123.87, 0.976401)
						(8201.29, 0.9994)
						(16385.4, 1.0)
						(17770.0, 1.0)
					};
					\addlegendentry{Cross-LSH}
					\addplot 
					coordinates 
					{
						(2.881, 0.0545999)
						(5.052, 0.0872998)
						(9.219, 0.1457)
						(14.411, 0.2117)
						(23.546, 0.303)
						(45.925, 0.4236)
						(77.161, 0.5122)
						(139.937, 0.6702)
						(261.963, 0.864601)
						(516.747, 0.939801)
						(1026.74, 0.9917)
						(2050.04, 0.9977)
						(4097.76, 0.9985)
						(8193.25, 0.9999)
						(16384.3, 1.0)
						(17770.0, 1.0)
					};
					\addlegendentry{Cross-LSH-NR}
			\end{axis}
			\end{tikzpicture}
	\end{subfigure}%
	\hspace{\figuregap}
	\begin{subfigure}[b]{\figurepercentperrow\textwidth}
	\begin{tikzpicture}
	\begin{axis}[
	height=\figurewidth\linewidth,
	width=\linewidth,
	legend pos=south east,
	legend style={inner xsep=0pt, inner ysep=0pt, font=\fontsize{1}{1}\selectfont},
	grid=major,
	legend pos=south east,
	change x base,
	x SI prefix=kilo,x unit=\-,
	xlabel=Probed Items ,
	xmin=0,xmax=2000,
	ymin=0.2,ymax=1,
	]
		\addplot 
		coordinates 
		{
			(17.786, 0.0132)
			(19.472, 0.015)
			(23.312, 0.0169)
			(29.185, 0.0203)
			(36.983, 0.0219)
			(65.372, 0.0236)
			(98.869, 0.0462)
			(179.253, 0.1284)
			(322.414, 0.2469)
			(627.332, 0.4507)
			(1221.68, 0.595899)
			(2248.05, 0.7394)
			(4176.56, 0.875101)
			(8351.56, 0.9701)
			(16390.6, 0.9999)
			(17770.0, 1.0)
		};
		\addlegendentry{L2-ALSH}
		\addplot 
		coordinates 
		{
			(9.596, 0.075)
			(11.478, 0.0841999)
			(13.391, 0.0894999)
			(17.439, 0.1242)
			(25.315, 0.1609)
			(41.422, 0.2267)
			(74.713, 0.3817)
			(139.739, 0.5934)
			(267.869, 0.7971)
			(533.35, 0.917801)
			(1046.19, 0.963401)
			(2071.32, 0.984801)
			(4113.52, 0.9954)
			(8201.93, 0.999)
			(16386.9, 1.0)
			(17770.0, 1.0)
		};
		\addlegendentry{L2-ALSH-NR}
		
	\end{axis}
	\end{tikzpicture}
	\end{subfigure}%
	\hspace{\figuregap}
	\begin{subfigure}[b]{\figurepercentperrow\textwidth}
	\begin{tikzpicture}
	\begin{axis}[
	height=\figurewidth\linewidth,
	width=\linewidth,
	legend pos=south east,
	legend style={inner xsep=0pt, inner ysep=0pt, font=\fontsize{1}{1}\selectfont},
	grid=major,
	change x base,
	x SI prefix=kilo,x unit=\-,
	xlabel=Probed Items,
	xmin=0,xmax=2000,
	ymin=0.2,ymax=1,
	]
		\addplot 
		coordinates 
		{
			(8.79, 0.0253)
			(10.323, 0.029)
			(12.904, 0.0392)
			(23.688, 0.0544999)
			(42.275, 0.0823999)
			(63.908, 0.1051)
			(111.513, 0.18)
			(163.909, 0.2152)
			(322.068, 0.3164)
			(564.366, 0.4197)
			(1065.0, 0.639)
			(2101.27, 0.821401)
			(4163.45, 0.924501)
			(8253.57, 0.9865)
			(16405.5, 0.9995)
			(17770.0, 1.0)
		};
		\addlegendentry{Simple-LSH}
		\addplot 
		coordinates 
		{
			(2.326, 0.1314)
			(3.025, 0.152401)
			(4.823, 0.191401)
			(8.906, 0.2591)
			(16.676, 0.343801)
			(33.395, 0.4861)
			(65.218, 0.6179)
			(129.655, 0.7375)
			(257.53, 0.8236)
			(513.212, 0.891501)
			(1024.85, 0.935501)
			(2048.93, 0.969001)
			(4097.39, 0.9894)
			(8193.05, 0.9988)
			(16385.1, 1.0)
			(17770.0, 1.0)
		};
		\addlegendentry{Simple-LSH-NR}

	\end{axis}
	\end{tikzpicture}
	\end{subfigure}%	
	\hspace{\figuregap}	
	\begin{subfigure}[b]{\figurepercentperrow\textwidth}
		\begin{tikzpicture}
		\begin{axis}[
		height=\figurewidth\linewidth,
		width=\linewidth,
		legend pos=south east,
		legend style={inner xsep=0pt, inner ysep=0pt, font=\fontsize{1}{1}\selectfont},
		grid=major,
		legend pos=south east,
		change x base,
		x SI prefix=kilo,x unit=\-,
		xlabel=Probed Items ,
		xmin=0,xmax=2000,
		ymin=0.2,ymax=1,
		]
				\addplot 
				coordinates 
				{
					(8.265, 0.0429)
					(10.006, 0.0729)
					(17.613, 0.1086)
					(28.966, 0.1368)
					(42.946, 0.1926)
					(61.329, 0.2395)
					(99.26, 0.3102)
					(174.417, 0.4132)
					(323.457, 0.5277)
					(563.655, 0.6152)
					(1077.82, 0.7147)
					(2098.84, 0.842301)
					(4152.46, 0.928301)
					(8246.23, 0.9865)
					(16402.4, 1.0)
					(17770.0, 1.0)
				};
				\addlegendentry{Sign-ALSH}
				\addplot 
				coordinates 
				{
					(6.506, 0.0364)
					(20.284, 0.0738999)
					(38.905, 0.1309)
					(47.497, 0.1869)
					(61.5, 0.2695)
					(71.187, 0.307)
					(84.014, 0.3528)
					(165.523, 0.738201)
					(264.149, 0.875501)
					(517.51, 0.965501)
					(1029.17, 0.9892)
					(2052.35, 0.9959)
					(4100.4, 0.9991)
					(8194.82, 1.0)
					(16384.7, 1.0)
					(17770.0, 1.0)
				};
				\addlegendentry{Sign-ALSH-NR}
				
		\end{axis}
		\end{tikzpicture}
	\end{subfigure}%
	
		\begin{subfigure}[b]{\figurepercentperrow\textwidth}
			\begin{tikzpicture}
			\begin{axis}[
			height=\figurewidth\linewidth,
			width=\linewidth,
			grid=major,
			legend pos=south east,
			legend style={inner xsep=0pt, inner ysep=0pt, font=\fontsize{1}{1}\selectfont},
			change x base,
			x SI prefix=kilo,x unit=\-,
			xmin=0,xmax=20000,
			ymin=0.2,ymax=1,
			xlabel=Probed Items , ylabel=Recall,
			]
				\addplot 
				coordinates 
				{
					(1.293, 0.0186)
					(2.462, 0.0327001)
					(4.737, 0.0527999)
					(9.107, 0.0881998)
					(17.636, 0.1395)
					(34.098, 0.1965)
					(66.401, 0.2758)
					(131.114, 0.3734)
					(260.59, 0.4728)
					(517.042, 0.5806)
					(1029.51, 0.6851)
					(2054.43, 0.7787)
					(4102.73, 0.860801)
					(8200.81, 0.921301)
					(16396.9, 0.962201)
					(32784.3, 0.9842)
					(65570.3, 0.9955)
					(131120.0, 0.9999)
					(136736.0, 1.0)
				};
				\addlegendentry{Cross-LSH}
				\addplot 
				coordinates 
				{
					(1.537, 0.017)
					(2.765, 0.0277001)
					(5.181, 0.0478)
					(9.373, 0.0767999)
					(17.5, 0.1162)
					(33.692, 0.1789)
					(66.263, 0.2654)
					(131.39, 0.3836)
					(260.313, 0.5177)
					(516.49, 0.6535)
					(1029.41, 0.789201)
					(2052.97, 0.888401)
					(4099.2, 0.951701)
					(8193.2, 0.9892)
					(16385.0, 0.9987)
					(32768.9, 1.0)
					(65536.8, 1.0)
					(131083.0, 1.0)
					(136736.0, 1.0)
				};
				\addlegendentry{Cross-LSH-NR}
			\end{axis}
			\end{tikzpicture}
	\end{subfigure}%
	\hspace{\figuregap}
	\begin{subfigure}[b]{\figurepercentperrow\textwidth}
	\begin{tikzpicture}
	\begin{axis}[
	height=\figurewidth\linewidth,
	width=\linewidth,
	legend pos=south east,
	legend style={inner xsep=0pt, inner ysep=0pt, font=\fontsize{1}{1}\selectfont},
	grid=major,
	change x base,
	x SI prefix=kilo,x unit=\-,
	xlabel=Probed Items,
	xmin=0,xmax=20000,
	ymin=0.2,ymax=1,
	]
	
			\addplot 
			coordinates 
			{
				(37.184, 0.008)
				(45.012, 0.0132)
				(68.517, 0.0175)
				(81.707, 0.0235)
				(114.103, 0.0353)
				(151.083, 0.0590999)
				(202.558, 0.0823999)
				(279.606, 0.1053)
				(405.205, 0.1321)
				(743.523, 0.1882)
				(1242.04, 0.2357)
				(2229.79, 0.3057)
				(4300.53, 0.4041)
				(8428.5, 0.5214)
				(16601.8, 0.647)
				(32999.2, 0.791401)
				(65753.0, 0.924801)
				(131155.0, 0.9987)
				(136736.0, 1.0)
			};
			\addlegendentry{L2-ALSH}
			\addplot 
			coordinates 
			{
				(3.086, 0.0088)
				(4.719, 0.0156)
				(7.329, 0.025)
				(12.796, 0.0386)
				(22.726, 0.0614999)
				(39.009, 0.0935997)
				(73.836, 0.1467)
				(143.679, 0.2182)
				(268.473, 0.304401)
				(525.072, 0.4312)
				(1034.06, 0.5699)
				(2056.99, 0.6979)
				(4101.15, 0.807301)
				(8195.94, 0.879301)
				(16387.7, 0.926901)
				(32771.8, 0.962701)
				(65542.6, 0.99)
				(131076.0, 0.9998)
				(136736.0, 1.0)
			};
			\addlegendentry{L2-ALSH-NR}
	
	\end{axis}
	\end{tikzpicture}
\end{subfigure}%
	\hspace{\figuregap}
	\begin{subfigure}[b]{\figurepercentperrow\textwidth}
	\begin{tikzpicture}
	\begin{axis}[
	height=\figurewidth\linewidth,
	width=\linewidth,
	legend pos=south east,
	legend style={inner xsep=0pt, inner ysep=0pt, font=\fontsize{1}{1}\selectfont},
	grid=major,
	change x base,
	x SI prefix=kilo,x unit=\-,
	xlabel=Probed Items , 
	xmin=0,xmax=20000,
	ymin=0.2,ymax=1,
	]
			\addplot 
			coordinates 
			{
				(1.601, 0.0082)
				(3.024, 0.0143)
				(6.312, 0.0219)
				(11.367, 0.0322)
				(20.84, 0.0488)
				(39.239, 0.0690999)
				(73.361, 0.0940998)
				(148.204, 0.1409)
				(288.701, 0.1973)
				(557.718, 0.2676)
				(1083.85, 0.3476)
				(2126.98, 0.4415)
				(4197.96, 0.544)
				(8317.42, 0.657299)
				(16518.1, 0.7591)
				(32923.3, 0.871501)
				(65685.6, 0.960701)
				(131149.0, 0.9995)
				(136736.0, 1.0)
			};
			\addlegendentry{Simple-LSH}
			\addplot 
			coordinates 
			{
				(4.49, 0.0118)
				(7.271, 0.0174)
				(10.216, 0.0239)
				(14.716, 0.0374)
				(27.313, 0.0559999)
				(43.684, 0.0825998)
				(79.315, 0.1257)
				(143.826, 0.1929)
				(273.842, 0.2979)
				(537.503, 0.4379)
				(1050.59, 0.614899)
				(2073.41, 0.788)
				(4101.71, 0.926901)
				(8194.89, 0.978601)
				(16385.9, 0.9936)
				(32769.5, 0.9979)
				(65537.9, 0.9997)
				(131105.0, 1.0)
				(136736.0, 1.0)
			};
			\addlegendentry{Simple-LSH-NR}
					
	\end{axis}
	\end{tikzpicture}
\end{subfigure}%
	\hspace{\figuregap}
	\begin{subfigure}[b]{\figurepercentperrow\textwidth}
		\begin{tikzpicture}
		\begin{axis}[
		height=\figurewidth\linewidth,
		width=\linewidth,
		legend pos=south east,
		legend style={inner xsep=0pt, inner ysep=0pt, font=\fontsize{1}{1}\selectfont},
		grid=major,
		change x base,
		x SI prefix=kilo,x unit=\-,
		xlabel=Probed Items ,
		xmin=0,xmax=20000,
		ymin=0.2,ymax=1,
		]

			\addplot 
			coordinates 
			{
				(2.659, 0.0082)
				(4.077, 0.014)
				(7.567, 0.0212)
				(12.545, 0.0322)
				(21.04, 0.0485999)
				(37.739, 0.0751999)
				(72.54, 0.1114)
				(137.86, 0.1584)
				(268.414, 0.2326)
				(529.784, 0.317701)
				(1053.23, 0.4221)
				(2081.19, 0.5372)
				(4133.63, 0.6628)
				(8238.43, 0.773901)
				(16445.3, 0.866501)
				(32845.8, 0.938301)
				(65613.7, 0.981701)
				(131111.0, 0.9998)
				(136736.0, 1.0)
			};
			\addlegendentry{Sign-ALSH}
			\addplot 
			coordinates 
			{
				(8.494, 0.0194)
				(11.487, 0.0251)
				(16.967, 0.0381)
				(25.14, 0.0509999)
				(37.51, 0.0668999)
				(59.629, 0.0985998)
				(124.075, 0.1434)
				(198.693, 0.2151)
				(338.995, 0.2899)
				(601.502, 0.407101)
				(1111.74, 0.5425)
				(2109.76, 0.703)
				(4108.74, 0.8868)
				(8198.9, 0.9739)
				(16390.0, 0.9933)
				(32773.5, 0.9986)
				(65540.6, 0.9999)
				(131091.0, 1.0)
				(136736.0, 1.0)
			};
			\addlegendentry{Sign-ALSH-NR}
		\end{axis}
		\end{tikzpicture}
	\end{subfigure}%
	
		\begin{subfigure}[b]{\figurepercentperrow\textwidth}
			\begin{tikzpicture}
			\begin{axis}[
				height=\figurewidth\linewidth,
				width=\linewidth,
				grid=major,
				legend pos=south east,
				legend style={inner xsep=0pt, inner ysep=0pt, font=\fontsize{1}{1}\selectfont},
				change x base,
				x SI prefix=mega,x unit=\-,
				xmin=0,xmax=400000,
				ymin=0.2,ymax=1,
				xlabel=Probed Items , ylabel=Recall,
			]

					\addplot 
					coordinates 
					{
						(2.322, 0.004)
						(3.551, 0.0082)
						(6.743, 0.0147)
						(12.483, 0.0221)
						(22.075, 0.0363)
						(44.082, 0.0598999)
						(82.993, 0.0902998)
						(155.21, 0.1269)
						(298.75, 0.1793)
						(674.045, 0.2448)
						(1262.74, 0.316)
						(2713.62, 0.3989)
						(5400.41, 0.4892)
						(10290.0, 0.5682)
						(20632.6, 0.642901)
						(41640.8, 0.710401)
						(83166.0, 0.7642)
						(157142.0, 0.8063)
						(300941.0, 0.846801)
						(581439.0, 0.8892)
						(1107000.0, 0.9391)
						(2124240.0, 0.9903)
						(2340370.0, 1.0)
					};
					\addlegendentry{Cross-LSH}
					\addplot 
					coordinates 
					{
						(1.603, 0.0057)
						(3.171, 0.0091)
						(5.933, 0.0142)
						(10.55, 0.0233)
						(20.033, 0.0389)
						(37.074, 0.0602999)
						(72.621, 0.0932998)
						(141.37, 0.1386)
						(274.065, 0.1969)
						(535.697, 0.2783)
						(1058.8, 0.3857)
						(2087.64, 0.4989)
						(4127.42, 0.6225)
						(8213.85, 0.7287)
						(16402.4, 0.8101)
						(32796.1, 0.861201)
						(131073.0, 0.9945)
						(262145.0, 0.999)
						(524288.0, 1.0)
						(1048580.0, 1.0)
						(2097150.0, 1.0)
						(2340370.0, 1.0)
					};
					\addlegendentry{Cross-LSH-NR}
				
			\end{axis}
			\end{tikzpicture}	
	\end{subfigure}%
	\hspace{\figuregap}
	\begin{subfigure}[b]{\figurepercentperrow\textwidth}
		\begin{tikzpicture}
		\begin{axis}[
		height=\figurewidth\linewidth,
		width=\linewidth,
		grid=major,
		legend pos=south east,
		legend style={inner xsep=0pt, inner ysep=0pt, font=\fontsize{1}{1}\selectfont},
		change x base,
		x SI prefix=mega,x unit=\-,
		xlabel=Probed Items ,
		xmin=0,xmax=400000,
		ymin=0.2,ymax=1,
		]
		
				\addplot 
				coordinates 
				{
					(1.698, 0.0006)
					(3.89, 0.0009)
					(10.404, 0.0012)
					(26.536, 0.0025)
					(45.01, 0.0045)
					(81.143, 0.0085)
					(217.876, 0.0122)
					(337.848, 0.0175)
					(567.186, 0.0266)
					(1151.04, 0.04)
					(1954.29, 0.0564999)
					(4568.88, 0.0848998)
					(8943.98, 0.1236)
					(15156.9, 0.1678)
					(29258.6, 0.2266)
					(54284.7, 0.3101)
					(107756.0, 0.4023)
					(196288.0, 0.4918)
					(308760.0, 0.5684)
					(601008.0, 0.662)
					(1103820.0, 0.7923)
					(2110970.0, 0.958601)
					(2340370.0, 1.0)
				};
				\addlegendentry{L2-ALSH}
				\addplot 
				coordinates 
				{
					(2.72, 0.0015)
					(4.605, 0.0025)
					(7.125, 0.004)
					(12.01, 0.009)
					(19.845, 0.011)
					(37.605, 0.021)
					(72.71, 0.0365)
					(139.66, 0.0545)
					(272.875, 0.076)
					(529.6, 0.1355)
					(1047.91, 0.212)
					(2101.01, 0.323)
					(4145.02, 0.4345)
					(8296.3, 0.5465)
					(16469.7, 0.664)
					(32928.7, 0.792)
					(65556.0, 0.8885)
					(131081.0, 0.9525)
					(262150.0, 0.969)
					(524290.0, 0.975)
					(1048580.0, 0.978)
					(2097170.0, 0.9925)
					(2340370.0, 1.0)
				};
				\addlegendentry{L2-ALSH-NR}
		\end{axis}
		\end{tikzpicture}	
	\end{subfigure}%
	\hspace{\figuregap}	
	\begin{subfigure}[b]{\figurepercentperrow\textwidth}
	\begin{tikzpicture}
	\begin{axis}[
		height=\figurewidth\linewidth,
		width=\linewidth,
		grid=major,
		legend pos=south east,
		legend style={inner xsep=0pt, inner ysep=0pt, font=\fontsize{1}{1}\selectfont},
		change x base,
		x SI prefix=mega,x unit=\-,
		xlabel=Probed Items,
		xmin=0,xmax=400000,
		ymin=0.2,ymax=1,
	]
			\addplot 
			coordinates 
			{
				(1.248, 0.0014)
				(2.626, 0.0031)
				(7.65, 0.0052)
				(18.419, 0.0113)
				(34.384, 0.0179)
				(59.331, 0.0283)
				(134.612, 0.0488)
				(224.409, 0.0724999)
				(393.673, 0.1048)
				(836.196, 0.1386)
				(1736.51, 0.1868)
				(4192.8, 0.2368)
				(6827.95, 0.281601)
				(11745.3, 0.348)
				(24404.1, 0.4427)
				(50077.5, 0.5212)
				(85367.0, 0.5799)
				(180067.0, 0.6641)
				(348969.0, 0.767601)
				(750185.0, 0.840402)
				(1283770.0, 0.897301)
				(2137650.0, 0.977801)
				(2340370.0, 1.0)
			};
			\addlegendentry{Simple-LSH}
			\addplot 
			coordinates 
			{
				(1.418, 0.0018)
				(3.04, 0.0039)
				(5.701, 0.0082)
				(10.877, 0.0141)
				(21.333, 0.0243)
				(41.66, 0.0398)
				(82.78, 0.0641)
				(154.235, 0.0965998)
				(295.125, 0.141)
				(576.238, 0.2031)
				(1137.11, 0.2815)
				(2252.51, 0.37)
				(4408.22, 0.482)
				(8630.36, 0.600401)
				(16994.4, 0.714501)
				(33467.6, 0.819701)
				(65824.6, 0.9101)
				(131083.0, 0.9857)
				(262151.0, 0.9962)
				(524289.0, 0.9992)
				(1048580.0, 1.0)
			};
			\addlegendentry{Simple-LSH-NR}
	\end{axis}
	\end{tikzpicture}	
	\end{subfigure}%
	\hspace{\figuregap}	
	\begin{subfigure}[b]{\figurepercentperrow\textwidth}
			\begin{tikzpicture}
			\begin{axis}[
			height=\figurewidth\linewidth,
			width=\linewidth,
			legend pos=south east,
			legend style={inner xsep=0pt, inner ysep=0pt, font=\fontsize{1}{1}\selectfont},
			grid=major,
			change x base,
			x SI prefix=mega,x unit=\-,
			xlabel=Probed Items ,
			xmin=0,xmax=400000,
			ymin=0.2,ymax=1,
			]
				\addplot 
				coordinates 
				{
					(1.305, 0.0017)
					(2.433, 0.0032)
					(4.604, 0.0049)
					(9.198, 0.009)
					(19.078, 0.015)
					(36.799, 0.0246)
					(75.222, 0.0391)
					(159.582, 0.0603999)
					(313.99, 0.0896998)
					(661.117, 0.1361)
					(1344.11, 0.1874)
					(3077.46, 0.255)
					(5459.62, 0.3189)
					(12122.8, 0.3893)
					(27268.7, 0.4792)
					(63670.9, 0.5682)
					(111247.0, 0.641999)
					(241682.0, 0.6907)
					(406006.0, 0.7448)
					(714642.0, 0.8243)
					(1224640.0, 0.895201)
					(2150270.0, 0.970601)
					(2340370.0, 1.0)
				};
				\addlegendentry{Sign-ALSH}
				\addplot 
				coordinates 
				{
					(1.331, 0.0011)
					(2.446, 0.0021)
					(4.765, 0.0034)
					(10.03, 0.0072)
					(18.768, 0.0116)
					(37.191, 0.0199)
					(71.495, 0.035)
					(142.352, 0.0555)
					(272.276, 0.0840999)
					(542.329, 0.1355)
					(1060.97, 0.1983)
					(2099.53, 0.2891)
					(4155.56, 0.4136)
					(8251.94, 0.5486)
					(16449.3, 0.6957)
					(32829.7, 0.8135)
					(65585.2, 0.886201)
					(131076.0, 0.9831)
					(262145.0, 0.9954)
					(524290.0, 0.9988)
					(1048580.0, 1.0)
					(2097150.0, 1.0)
					(2340370.0, 1.0)
				};
				\addlegendentry{Sign-ALSH-NR}
			\end{axis}
			\end{tikzpicture}	
		\end{subfigure}%
	
	\caption{Probed item-recall of top-10 items comparison between the original meta algorithms and their norm-range versions under a code length of 32. From top row to bottom row, the datasets are Netflix, Yahoo!Music and ImageNet, respectively.} 
	\label{top10-bit32:Norm-Range comparision}
\end{figure*}

\begin{figure*}[t]
		\begin{subfigure}[b]{0.36\textwidth}
			\begin{tikzpicture}
			\begin{axis}[
			height=\figurewidth\linewidth,
			width=\linewidth,
			legend style={font=\fontsize{1}{1}\selectfont},
			grid=major,
			legend pos=south east,
			change x base,
			x SI prefix=kilo,x unit=\-,
			xlabel=Probed Items , ylabel=Recall,
			xmin=0,xmax=4000,
			ymin=0,ymax=1,
			]

				\addplot [mark=square*, red]
				coordinates 
				{
					(1.956, 0.01026)
					(3.367, 0.0190001)
					(5.745, 0.0313202)
					(10.746, 0.0517203)
					(19.911, 0.0810801)
					(38.253, 0.12748)
					(82.004, 0.20196)
					(158.274, 0.28794)
					(339.856, 0.39032)
					(634.608, 0.4782)
					(1135.33, 0.60514)
					(2130.34, 0.7959)
					(4123.87, 0.9418)
					(8201.29, 0.99672)
					(16385.4, 1.0)
					(17770.0, 1.0)
				};
				\addlegendentry{Cross-LSH}
				
				\addplot   [mark=triangle*, blue]
				coordinates 
				{
					(4.728, 0.00564)
					(7.521, 0.0075)
					(12.312, 0.00988001)
					(23.511, 0.01616)
					(62.02, 0.0607001)
					(83.006, 0.06972)
					(141.78, 0.15626)
					(197.928, 0.19746)
					(344.981, 0.25696)
					(675.162, 0.3584)
					(1186.94, 0.4774)
					(2136.47, 0.67668)
					(4255.7, 0.858781)
					(8432.3, 0.9634)
					(16436.0, 0.99878)
					(17770.0, 1.0)
				};
				\addlegendentry{Simple-LSH}
				
				\addplot 
				coordinates 
				{
					(17.786, 0.0093)
					(19.472, 0.01074)
					(23.312, 0.0138)
					(29.185, 0.01642)
					(36.983, 0.01802)
					(65.372, 0.02162)
					(98.869, 0.0396601)
					(179.253, 0.0992)
					(322.414, 0.20654)
					(627.332, 0.35082)
					(1221.68, 0.47586)
					(2248.05, 0.64672)
					(4176.56, 0.793761)
					(8351.56, 0.9474)
					(16390.6, 0.99972)
					(17770.0, 1.0)
				};
				\addlegendentry{L2-ALSH}

				\addplot 
				coordinates 
				{
					(5.605, 0.01468)
					(7.666, 0.0185001)
					(11.31, 0.0238401)
					(14.581, 0.0303201)
					(26.67, 0.0433802)
					(41.263, 0.0573202)
					(73.628, 0.09034)
					(138.968, 0.14856)
					(268.192, 0.24044)
					(525.745, 0.35294)
					(1036.56, 0.5239)
					(2062.23, 0.7028)
					(4117.93, 0.858861)
					(8210.89, 0.95978)
					(16393.3, 0.99936)
					(17770.0, 1.0)
				};
				\addlegendentry{Sign-ALSH}
			\end{axis}
			\end{tikzpicture}
	\end{subfigure}%
	\hspace{\figuregap}	
	\begin{subfigure}[b]{0.36\textwidth}
			\begin{tikzpicture}
			\begin{axis}[
			height=\figurewidth\linewidth,
			width=\linewidth,
			legend style={font=\fontsize{1}{1}\selectfont},
			grid=major,
			legend pos=south east,
			change x base,
			x SI prefix=kilo,x unit=\-,
			xlabel=Probed Items ,
			xmin=0,xmax=120000,
			ymin=0,ymax=1,
			]
			
				\addplot [mark=square*, red]
				coordinates 
				{
					(1.293, 0.00781999)
					(2.462, 0.01378)
					(4.737, 0.0242001)
					(9.107, 0.0423602)
					(17.636, 0.0716401)
					(34.098, 0.1146)
					(66.401, 0.17474)
					(131.114, 0.25826)
					(260.59, 0.35752)
					(517.042, 0.47192)
					(1029.51, 0.59092)
					(2054.43, 0.70568)
					(4102.73, 0.809381)
					(8200.81, 0.88966)
					(16396.9, 0.9441)
					(32784.3, 0.97664)
					(65570.3, 0.99416)
					(131120.0, 0.99968)
					(136736.0, 1.0)
				};
				\addlegendentry{Cross-LSH}
				
				\addplot   [mark=triangle*, blue]
				coordinates 
				{
					(1.981, 0.00648)
					(3.474, 0.0115)
					(5.962, 0.0179801)
					(10.695, 0.0292601)
					(19.668, 0.0458601)
					(36.785, 0.0729601)
					(70.551, 0.11476)
					(136.106, 0.17266)
					(266.411, 0.24542)
					(525.725, 0.3383)
					(1041.94, 0.43958)
					(2076.72, 0.55258)
					(4131.91, 0.65768)
					(8249.06, 0.76156)
					(16450.4, 0.84796)
					(32874.7, 0.91888)
					(65676.6, 0.96786)
					(131212.0, 0.99942)
					(136736.0, 1.0)
				};
				\addlegendentry{Simple-LSH}
				
				\addplot 
				coordinates 
				{
					(37.184, 0.00694)
					(45.012, 0.01138)
					(68.517, 0.0151)
					(81.707, 0.01896)
					(114.103, 0.0276801)
					(151.083, 0.0465201)
					(202.558, 0.0651001)
					(279.606, 0.08496)
					(405.205, 0.10714)
					(743.523, 0.15812)
					(1242.04, 0.20572)
					(2229.79, 0.2756)
					(4300.53, 0.37658)
					(8428.5, 0.49864)
					(16601.8, 0.63724)
					(32999.2, 0.789561)
					(65753.0, 0.92692)
					(131155.0, 0.99914)
					(136736.0, 1.0)
				};
				\addlegendentry{L2-ALSH}
				
				\addplot 
				coordinates 
				{
					(1.982, 0.004)
					(3.463, 0.00677999)
					(5.807, 0.01082)
					(10.339, 0.0177401)
					(19.306, 0.0287801)
					(39.431, 0.0456802)
					(74.244, 0.0720402)
					(141.491, 0.11032)
					(267.035, 0.165)
					(524.042, 0.244)
					(1039.61, 0.34564)
					(2065.3, 0.4676)
					(4114.56, 0.59954)
					(8213.47, 0.735481)
					(16409.0, 0.8514)
					(32798.9, 0.93812)
					(65577.9, 0.98468)
					(131116.0, 0.99988)
					(136736.0, 1.0)
				};
				\addlegendentry{Sign-ALSH}
			
			\end{axis}
			\end{tikzpicture}
	\end{subfigure}%
	\hspace{\figuregap}	
	\begin{subfigure}[b]{0.36\textwidth}
			\begin{tikzpicture}
			\begin{axis}[
			height=\figurewidth\linewidth,
			width=\linewidth,
			legend style={font=\fontsize{1}{1}\selectfont},
			grid=major,
			legend pos=south east,
			change x base,
			x SI prefix=mega,x unit=\-,
			xlabel=Probed Items ,
			xmin=0,xmax=1000000,
			ymin=0,ymax=1,
			]
			
					\addplot [mark=square*, red]
					coordinates 
					{
						(2.322, 0.00188)
						(3.551, 0.00352)
						(6.743, 0.00679999)
						(12.483, 0.01148)
						(22.075, 0.0198001)
						(44.082, 0.0323801)
						(82.993, 0.0536402)
						(155.21, 0.08372)
						(298.75, 0.12764)
						(674.045, 0.1846)
						(1262.74, 0.2569)
						(2713.62, 0.34288)
						(5400.41, 0.43276)
						(10290.0, 0.52154)
						(20632.6, 0.60438)
						(41640.8, 0.67858)
						(83166.0, 0.7368)
						(157142.0, 0.785421)
						(300941.0, 0.829699)
						(581439.0, 0.8779)
						(1107000.0, 0.93206)
						(2124240.0, 0.98888)
						(2340370.0, 1.0)
					};
					\addlegendentry{Cross-LSH}
					
					\addplot   [mark=triangle*, blue]
					coordinates 
					{
						(1.326, 0.00124)
						(2.878, 0.0018)
						(5.506, 0.00296)
						(11.817, 0.00466)
						(22.891, 0.00787999)
						(43.126, 0.01406)
						(94.608, 0.0242401)
						(182.383, 0.0383201)
						(329.976, 0.0611802)
						(627.04, 0.09448)
						(1457.21, 0.13626)
						(2637.62, 0.19248)
						(4909.55, 0.25284)
						(9502.59, 0.32468)
						(18467.5, 0.39686)
						(35509.0, 0.4776)
						(70174.0, 0.56368)
						(139202.0, 0.65324)
						(275613.0, 0.737101)
						(547027.0, 0.823061)
						(1075300.0, 0.90558)
						(2109100.0, 0.98864)
						(2340370.0, 1.0)
					};
					\addlegendentry{Simple-LSH}
					
					\addplot 
					coordinates 
					{
						(1.698, 0.00038)
						(3.89, 0.00054)
						(10.404, 0.00094)
						(26.536, 0.0018)
						(45.01, 0.00336)
						(81.143, 0.00626)
						(217.876, 0.00904)
						(337.848, 0.0143)
						(567.186, 0.0222001)
						(1151.04, 0.0346201)
						(1954.29, 0.0509201)
						(4568.88, 0.0755601)
						(8943.98, 0.11144)
						(15156.9, 0.15392)
						(29258.6, 0.20668)
						(54284.7, 0.28842)
						(107756.0, 0.37824)
						(196288.0, 0.4673)
						(308760.0, 0.54256)
						(601008.0, 0.64068)
						(1103820.0, 0.784481)
						(2110970.0, 0.95686)
						(2340370.0, 1.0)
					};
					\addlegendentry{L2-ALSH}

					\addplot 
					coordinates 
					{
						(1.905, 0.0011)
						(3.582, 0.00212)
						(7.059, 0.00374)
						(14.287, 0.00638)
						(53.252, 0.01018)
						(107.651, 0.0169)
						(375.693, 0.0283001)
						(543.844, 0.0443602)
						(852.903, 0.0656602)
						(1195.74, 0.09554)
						(2421.62, 0.13444)
						(4124.82, 0.1814)
						(6942.04, 0.23654)
						(12209.2, 0.30028)
						(22186.1, 0.36746)
						(39859.1, 0.43834)
						(75959.1, 0.52084)
						(142581.0, 0.5996)
						(277795.0, 0.6959)
						(545277.0, 0.794901)
						(1068470.0, 0.88838)
						(2114110.0, 0.98346)
						(2340370.0, 1.0)
					};
					\addlegendentry{Sign-ALSH}

			\end{axis}
			\end{tikzpicture}	
	\end{subfigure}%
	\caption{Probed item-recall of top-50 items comparison between Cross-LSH and existing algorithms under a code length of 32 (best viewed in color). From left to right, the datasets are Netflix, Yahoo!Music and ImageNet, respectively.} 
	\label{top50-bit32:Cross comparision}
\end{figure*}
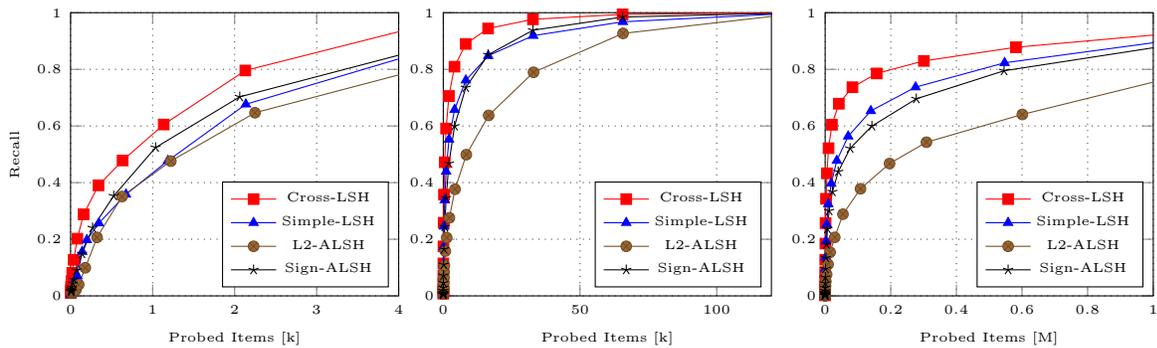  

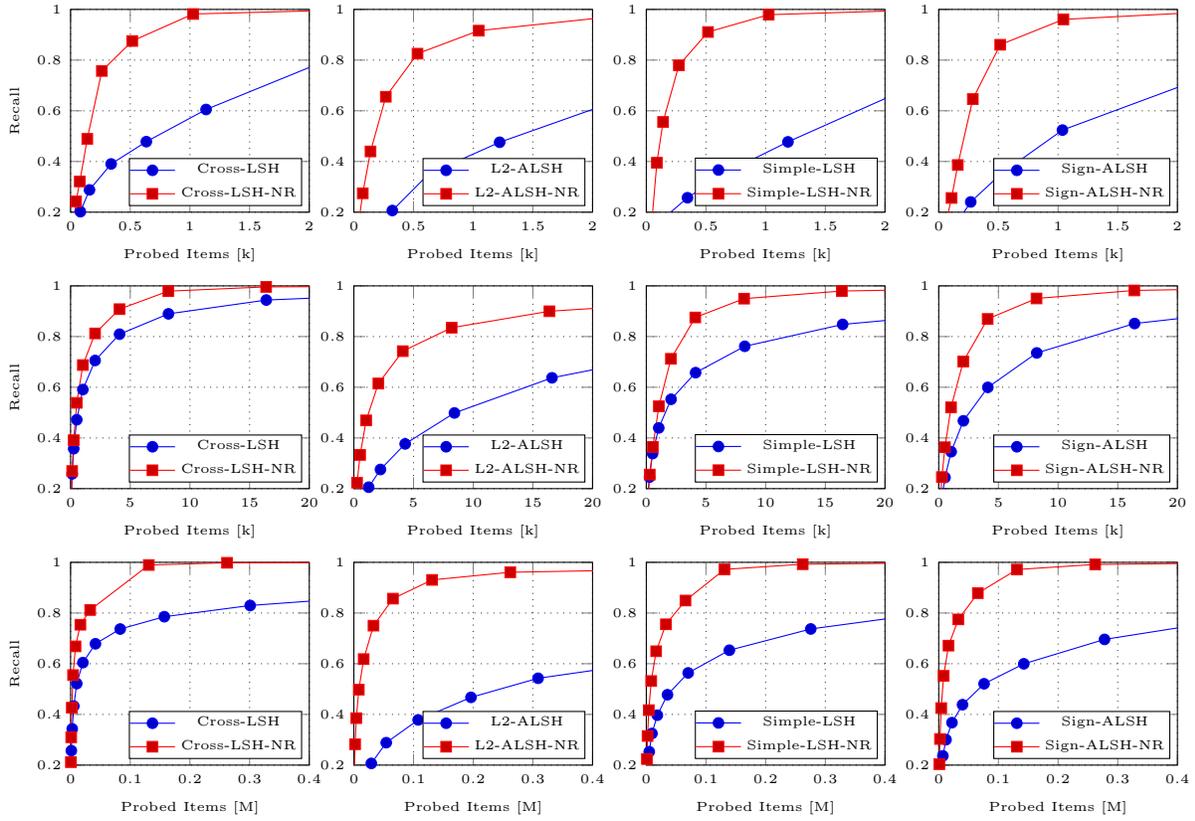
\begin{figure*}[t]
		\begin{subfigure}[b]{\figurepercentperrow\textwidth}
			\begin{tikzpicture}
			\begin{axis}[
			height=\figurewidth\linewidth,
			width=\linewidth,
			grid=major,
			legend pos=south east,
			legend style={inner xsep=0pt, inner ysep=0pt, font=\fontsize{1}{1}\selectfont},
			change x base,
			x SI prefix=kilo,x unit=\-,
			xlabel=Probed Items , ylabel=Recall,
			xmin=0,xmax=2000,
			ymin=0.2,ymax=1,
			mark size=2.0pt,
			]
				\addplot 
				coordinates 
				{
					(1.956, 0.01026)
					(3.367, 0.0190001)
					(5.745, 0.0313202)
					(10.746, 0.0517203)
					(19.911, 0.0810801)
					(38.253, 0.12748)
					(82.004, 0.20196)
					(158.274, 0.28794)
					(339.856, 0.39032)
					(634.608, 0.4782)
					(1135.33, 0.60514)
					(2130.34, 0.7959)
					(4123.87, 0.9418)
					(8201.29, 0.99672)
					(16385.4, 1.0)
					(17770.0, 1.0)
				};
				\addlegendentry{Cross-LSH}
				\addplot 
				coordinates 
				{
					(2.881, 0.0276802)
					(5.052, 0.0447603)
					(9.219, 0.0752403)
					(14.411, 0.11204)
					(23.546, 0.16292)
					(45.925, 0.24246)
					(77.161, 0.32128)
					(139.937, 0.48928)
					(261.963, 0.757201)
					(516.747, 0.875321)
					(1026.74, 0.9819)
					(2050.04, 0.99468)
					(4097.76, 0.99808)
					(8193.25, 0.99938)
					(16384.3, 1.0)
					(17770.0, 1.0)
				};
				\addlegendentry{Cross-LSH-NR}
					
			\end{axis}
			\end{tikzpicture}
	\end{subfigure}%
	\hspace{\figuregap}
	\begin{subfigure}[b]{\figurepercentperrow\textwidth}
	\begin{tikzpicture}
	\begin{axis}[
	height=\figurewidth\linewidth,
	width=\linewidth,
	legend pos=south east,
	legend style={inner xsep=0pt, inner ysep=0pt, font=\fontsize{1}{1}\selectfont},
	grid=major,
	legend pos=south east,
	change x base,
	x SI prefix=kilo,x unit=\-,
	xlabel=Probed Items ,
	xmin=0,xmax=2000,
	ymin=0.2,ymax=1,
	]
			\addplot 
			coordinates 
			{
				(17.786, 0.0093)
				(19.472, 0.01074)
				(23.312, 0.0138)
				(29.185, 0.01642)
				(36.983, 0.01802)
				(65.372, 0.02162)
				(98.869, 0.0396601)
				(179.253, 0.0992)
				(322.414, 0.20654)
				(627.332, 0.35082)
				(1221.68, 0.47586)
				(2248.05, 0.64672)
				(4176.56, 0.793761)
				(8351.56, 0.9474)
				(16390.6, 0.99972)
				(17770.0, 1.0)
			};
			\addlegendentry{L2-ALSH}
			\addplot 
			coordinates 
			{
				(9.596, 0.0451801)
				(11.478, 0.0536001)
				(13.391, 0.0585201)
				(17.439, 0.0737801)
				(25.315, 0.0988)
				(41.422, 0.16168)
				(74.713, 0.27448)
				(139.739, 0.43972)
				(267.869, 0.65538)
				(533.35, 0.82542)
				(1046.19, 0.91618)
				(2071.32, 0.9668)
				(4113.52, 0.99032)
				(8201.93, 0.99802)
				(16386.9, 0.99984)
				(17770.0, 1.0)
			};
			\addlegendentry{L2-ALSH-NR}
	\end{axis}
	\end{tikzpicture}
	\end{subfigure}%
	\hspace{\figuregap}
	\begin{subfigure}[b]{\figurepercentperrow\textwidth}
	\begin{tikzpicture}
	\begin{axis}[
	height=\figurewidth\linewidth,
	width=\linewidth,
	legend pos=south east,
	legend style={inner xsep=0pt, inner ysep=0pt, font=\fontsize{1}{1}\selectfont},
	grid=major,
	change x base,
	x SI prefix=kilo,x unit=\-,
	xlabel=Probed Items,
	xmin=0,xmax=2000,
	ymin=0.2,ymax=1,
	]
			\addplot 
			coordinates 
			{
				(4.728, 0.00564)
				(7.521, 0.0075)
				(12.312, 0.00988001)
				(23.511, 0.01616)
				(62.02, 0.0607001)
				(83.006, 0.06972)
				(141.78, 0.15626)
				(197.928, 0.19746)
				(344.981, 0.25696)
				(675.162, 0.3584)
				(1186.94, 0.4774)
				(2136.47, 0.67668)
				(4255.7, 0.858781)
				(8432.3, 0.9634)
				(16436.0, 0.99878)
				(17770.0, 1.0)
			};
			\addlegendentry{Simple-LSH}
			\addplot 
			coordinates 
			{
				(3.043, 0.00824)
				(4.068, 0.01018)
				(8.864, 0.0220401)
				(14.813, 0.0337401)
				(19.956, 0.0519201)
				(38.233, 0.11844)
				(88.487, 0.39506)
				(140.512, 0.55604)
				(272.809, 0.77956)
				(517.121, 0.91084)
				(1025.67, 0.97918)
				(2051.05, 0.99408)
				(4097.98, 0.99828)
				(8193.53, 0.9995)
				(16386.7, 1.0)
				(17770.0, 1.0)
			};
			\addlegendentry{Simple-LSH-NR}

	\end{axis}
	\end{tikzpicture}
	\end{subfigure}%	
	\hspace{\figuregap}	
	\begin{subfigure}[b]{\figurepercentperrow\textwidth}
		\begin{tikzpicture}
		\begin{axis}[
		height=\figurewidth\linewidth,
		width=\linewidth,
		legend pos=south east,
		legend style={inner xsep=0pt, inner ysep=0pt, font=\fontsize{1}{1}\selectfont},
		grid=major,
		legend pos=south east,
		change x base,
		x SI prefix=kilo,x unit=\-,
		xlabel=Probed Items ,
		xmin=0,xmax=2000,
		ymin=0.2,ymax=1,
		]
			\addplot 
			coordinates 
			{
				(5.605, 0.01468)
				(7.666, 0.0185001)
				(11.31, 0.0238401)
				(14.581, 0.0303201)
				(26.67, 0.0433802)
				(41.263, 0.0573202)
				(73.628, 0.09034)
				(138.968, 0.14856)
				(268.192, 0.24044)
				(525.745, 0.35294)
				(1036.56, 0.5239)
				(2062.23, 0.7028)
				(4117.93, 0.858861)
				(8210.89, 0.95978)
				(16393.3, 0.99936)
				(17770.0, 1.0)
			};
			\addlegendentry{Sign-ALSH}
			\addplot 
			coordinates 
			{
				(25.214, 0.0781201)
				(27.953, 0.0841)
				(30.651, 0.09256)
				(31.785, 0.09614)
				(46.884, 0.11582)
				(61.387, 0.16196)
				(107.463, 0.256)
				(159.608, 0.3862)
				(285.514, 0.64626)
				(515.132, 0.86058)
				(1044.65, 0.9606)
				(2068.63, 0.98552)
				(4114.14, 0.99614)
				(8205.83, 0.99978)
				(16386.8, 1.0)
				(17770.0, 1.0)
			};
			\addlegendentry{Sign-ALSH-NR}
		\end{axis}
		\end{tikzpicture}
	\end{subfigure}%
	
		\begin{subfigure}[b]{\figurepercentperrow\textwidth}
			\begin{tikzpicture}
			\begin{axis}[
			height=\figurewidth\linewidth,
			width=\linewidth,
			grid=major,
			legend pos=south east,
			legend style={inner xsep=0pt, inner ysep=0pt, font=\fontsize{1}{1}\selectfont},
			change x base,
			x SI prefix=kilo,x unit=\-,
			xmin=0,xmax=20000,
			ymin=0.2,ymax=1,
			xlabel=Probed Items , ylabel=Recall,
			]
						\addplot 
				coordinates 
				{
					(1.293, 0.00781999)
					(2.462, 0.01378)
					(4.737, 0.0242001)
					(9.107, 0.0423602)
					(17.636, 0.0716401)
					(34.098, 0.1146)
					(66.401, 0.17474)
					(131.114, 0.25826)
					(260.59, 0.35752)
					(517.042, 0.47192)
					(1029.51, 0.59092)
					(2054.43, 0.70568)
					(4102.73, 0.809381)
					(8200.81, 0.88966)
					(16396.9, 0.9441)
					(32784.3, 0.97664)
					(65570.3, 0.99416)
					(131120.0, 0.99968)
					(136736.0, 1.0)
				};
				\addlegendentry{Cross-LSH}
				\addplot 
				coordinates 
				{
					(1.537, 0.00781999)
					(2.765, 0.01344)
					(5.181, 0.0238601)
					(9.373, 0.0414802)
					(17.5, 0.0677002)
					(33.692, 0.10962)
					(66.263, 0.17382)
					(131.39, 0.26952)
					(260.313, 0.39118)
					(516.49, 0.5385)
					(1029.41, 0.68766)
					(2052.97, 0.81242)
					(4099.2, 0.90822)
					(8193.2, 0.97918)
					(16385.0, 0.99592)
					(32768.9, 0.99934)
					(65536.8, 1.0)
					(131083.0, 1.0)
					(136736.0, 1.0)
				};
				\addlegendentry{Cross-LSH-NR}
			\end{axis}
			\end{tikzpicture}
	\end{subfigure}%
	\hspace{\figuregap}
	\begin{subfigure}[b]{\figurepercentperrow\textwidth}
	\begin{tikzpicture}
	\begin{axis}[
	height=\figurewidth\linewidth,
	width=\linewidth,
	legend pos=south east,
	legend style={inner xsep=0pt, inner ysep=0pt, font=\fontsize{1}{1}\selectfont},
	grid=major,
	change x base,
	x SI prefix=kilo,x unit=\-,
	xlabel=Probed Items,
	xmin=0,xmax=20000,
	ymin=0.2,ymax=1,
	]

			\addplot 
			coordinates 
			{
				(37.184, 0.00694)
				(45.012, 0.01138)
				(68.517, 0.0151)
				(81.707, 0.01896)
				(114.103, 0.0276801)
				(151.083, 0.0465201)
				(202.558, 0.0651001)
				(279.606, 0.08496)
				(405.205, 0.10714)
				(743.523, 0.15812)
				(1242.04, 0.20572)
				(2229.79, 0.2756)
				(4300.53, 0.37658)
				(8428.5, 0.49864)
				(16601.8, 0.63724)
				(32999.2, 0.789561)
				(65753.0, 0.92692)
				(131155.0, 0.99914)
				(136736.0, 1.0)
			};
			\addlegendentry{L2-ALSH}
			\addplot 
			coordinates 
			{
				(3.086, 0.0051)
				(4.719, 0.00833999)
				(7.329, 0.01344)
				(12.796, 0.0226601)
				(22.726, 0.0375001)
				(39.009, 0.0579802)
				(73.836, 0.09518)
				(143.679, 0.15146)
				(268.473, 0.22364)
				(525.072, 0.33302)
				(1034.06, 0.46932)
				(2056.99, 0.61492)
				(4101.15, 0.74248)
				(8195.94, 0.835)
				(16387.7, 0.89988)
				(32771.8, 0.949399)
				(65542.6, 0.98326)
				(131076.0, 0.99946)
				(136736.0, 1.0)
			};
			\addlegendentry{L2-ALSH-NR}
	\end{axis}
	\end{tikzpicture}
\end{subfigure}%
	\hspace{\figuregap}
	\begin{subfigure}[b]{\figurepercentperrow\textwidth}
	\begin{tikzpicture}
	\begin{axis}[
	height=\figurewidth\linewidth,
	width=\linewidth,
	legend pos=south east,
	legend style={inner xsep=0pt, inner ysep=0pt, font=\fontsize{1}{1}\selectfont},
	grid=major,
	change x base,
	x SI prefix=kilo,x unit=\-,
	xlabel=Probed Items , 
	xmin=0,xmax=20000,
	ymin=0.2,ymax=1,
	]
			\addplot 
			coordinates 
			{
				(1.981, 0.00648)
				(3.474, 0.0115)
				(5.962, 0.0179801)
				(10.695, 0.0292601)
				(19.668, 0.0458601)
				(36.785, 0.0729601)
				(70.551, 0.11476)
				(136.106, 0.17266)
				(266.411, 0.24542)
				(525.725, 0.3383)
				(1041.94, 0.43958)
				(2076.72, 0.55258)
				(4131.91, 0.65768)
				(8249.06, 0.76156)
				(16450.4, 0.84796)
				(32874.7, 0.91888)
				(65676.6, 0.96786)
				(131212.0, 0.99942)
				(136736.0, 1.0)
			};
			\addlegendentry{Simple-LSH}
			\addplot 
			coordinates 
			{
				(3.051, 0.00844)
				(5.639, 0.0150801)
				(9.09, 0.0211601)
				(13.907, 0.0317001)
				(24.919, 0.0477602)
				(42.285, 0.0703201)
				(76.394, 0.1099)
				(148.104, 0.1717)
				(290.241, 0.2555)
				(547.999, 0.36462)
				(1052.61, 0.52502)
				(2064.22, 0.71224)
				(4103.43, 0.875361)
				(8194.96, 0.94964)
				(16385.4, 0.97968)
				(32769.2, 0.99324)
				(65537.1, 0.99866)
				(131080.0, 0.99998)
				(136736.0, 1.0)
			};
			\addlegendentry{Simple-LSH-NR}
					
	\end{axis}
	\end{tikzpicture}
\end{subfigure}%
	\hspace{\figuregap}
	\begin{subfigure}[b]{\figurepercentperrow\textwidth}
		\begin{tikzpicture}
		\begin{axis}[
		height=\figurewidth\linewidth,
		width=\linewidth,
		legend pos=south east,
		legend style={inner xsep=0pt, inner ysep=0pt, font=\fontsize{1}{1}\selectfont},
		grid=major,
		change x base,
		x SI prefix=kilo,x unit=\-,
		xlabel=Probed Items ,
		xmin=0,xmax=20000,
		ymin=0.2,ymax=1,
		]
		
				\addplot 
				coordinates 
				{
					(1.982, 0.004)
					(3.463, 0.00677999)
					(5.807, 0.01082)
					(10.339, 0.0177401)
					(19.306, 0.0287801)
					(39.431, 0.0456802)
					(74.244, 0.0720402)
					(141.491, 0.11032)
					(267.035, 0.165)
					(524.042, 0.244)
					(1039.61, 0.34564)
					(2065.3, 0.4676)
					(4114.56, 0.59954)
					(8213.47, 0.735481)
					(16409.0, 0.8514)
					(32798.9, 0.93812)
					(65577.9, 0.98468)
					(131116.0, 0.99988)
					(136736.0, 1.0)
				};
				\addlegendentry{Sign-ALSH}
				\addplot 
				coordinates 
				{
					(1.924, 0.00388)
					(3.398, 0.00698)
					(5.953, 0.01144)
					(10.425, 0.0192401)
					(19.251, 0.0320401)
					(35.802, 0.0540602)
					(68.913, 0.093)
					(133.114, 0.15294)
					(262.845, 0.2456)
					(519.366, 0.36342)
					(1031.66, 0.52084)
					(2054.88, 0.70156)
					(4099.23, 0.869501)
					(8193.96, 0.95038)
					(16385.6, 0.98198)
					(32769.6, 0.99534)
					(65537.9, 0.99946)
					(131086.0, 1.0)
					(136736.0, 1.0)
				};
				\addlegendentry{Sign-ALSH-NR}
		
		\end{axis}
		\end{tikzpicture}
	\end{subfigure}%
	
		\begin{subfigure}[b]{\figurepercentperrow\textwidth}
			\begin{tikzpicture}
			\begin{axis}[
				height=\figurewidth\linewidth,
				width=\linewidth,
				grid=major,
				legend pos=south east,
				legend style={inner xsep=0pt, inner ysep=0pt, font=\fontsize{1}{1}\selectfont},
				change x base,
				x SI prefix=mega,x unit=\-,
				xmin=0,xmax=400000,
				ymin=0.2,ymax=1,
				xlabel=Probed Items , ylabel=Recall,
			]
					\addplot 
					coordinates 
					{
						(2.322, 0.00188)
						(3.551, 0.00352)
						(6.743, 0.00679999)
						(12.483, 0.01148)
						(22.075, 0.0198001)
						(44.082, 0.0323801)
						(82.993, 0.0536402)
						(155.21, 0.08372)
						(298.75, 0.12764)
						(674.045, 0.1846)
						(1262.74, 0.2569)
						(2713.62, 0.34288)
						(5400.41, 0.43276)
						(10290.0, 0.52154)
						(20632.6, 0.60438)
						(41640.8, 0.67858)
						(83166.0, 0.7368)
						(157142.0, 0.785421)
						(300941.0, 0.829699)
						(581439.0, 0.8779)
						(1107000.0, 0.93206)
						(2124240.0, 0.98888)
						(2340370.0, 1.0)
					};
					\addlegendentry{Cross-LSH}
					\addplot 
					coordinates 
					{
						(1.603, 0.00226)
						(3.171, 0.00372)
						(5.933, 0.00691999)
						(10.55, 0.01192)
						(20.033, 0.0206801)
						(37.074, 0.0340601)
						(72.621, 0.0566602)
						(141.37, 0.08972)
						(274.065, 0.14024)
						(535.697, 0.21166)
						(1058.8, 0.3097)
						(2087.64, 0.42654)
						(4127.42, 0.555201)
						(8213.85, 0.66844)
						(16402.4, 0.75378)
						(32796.1, 0.81196)
						(131073.0, 0.98942)
						(262145.0, 0.99822)
						(524288.0, 0.99984)
						(1048580.0, 1.0)
						(2097150.0, 1.0)
						(2340370.0, 1.0)
					};
					\addlegendentry{Cross-LSH-NR}

			\end{axis}
			\end{tikzpicture}	
	\end{subfigure}%
	\hspace{\figuregap}
	\begin{subfigure}[b]{\figurepercentperrow\textwidth}
		\begin{tikzpicture}
		\begin{axis}[
		height=\figurewidth\linewidth,
		width=\linewidth,
		grid=major,
		legend pos=south east,
		legend style={inner xsep=0pt, inner ysep=0pt, font=\fontsize{1}{1}\selectfont},
		change x base,
		x SI prefix=mega,x unit=\-,
		xlabel=Probed Items ,
		xmin=0,xmax=400000,
		ymin=0.2,ymax=1,
		]
					\addplot 
					coordinates 
					{
						(1.698, 0.00038)
						(3.89, 0.00054)
						(10.404, 0.00094)
						(26.536, 0.0018)
						(45.01, 0.00336)
						(81.143, 0.00626)
						(217.876, 0.00904)
						(337.848, 0.0143)
						(567.186, 0.0222001)
						(1151.04, 0.0346201)
						(1954.29, 0.0509201)
						(4568.88, 0.0755601)
						(8943.98, 0.11144)
						(15156.9, 0.15392)
						(29258.6, 0.20668)
						(54284.7, 0.28842)
						(107756.0, 0.37824)
						(196288.0, 0.4673)
						(308760.0, 0.54256)
						(601008.0, 0.64068)
						(1103820.0, 0.784481)
						(2110970.0, 0.95686)
						(2340370.0, 1.0)
					};
					\addlegendentry{L2-ALSH}
					\addplot 
					coordinates 
					{
						(2.72, 0.002)
						(4.605, 0.0029)
						(7.125, 0.0037)
						(12.01, 0.0056)
						(19.845, 0.008)
						(37.605, 0.0123)
						(72.71, 0.0255)
						(139.66, 0.0431)
						(272.875, 0.0640001)
						(529.6, 0.1109)
						(1047.91, 0.1791)
						(2101.01, 0.2822)
						(4145.02, 0.3853)
						(8296.3, 0.4977)
						(16469.7, 0.6182)
						(32928.7, 0.7501)
						(65556.0, 0.8567)
						(131081.0, 0.9307)
						(262150.0, 0.9608)
						(524290.0, 0.972)
					};
					\addlegendentry{L2-ALSH-NR}

		\end{axis}
		\end{tikzpicture}	
	\end{subfigure}%
	\hspace{\figuregap}	
	\begin{subfigure}[b]{\figurepercentperrow\textwidth}
	\begin{tikzpicture}
	\begin{axis}[
		height=\figurewidth\linewidth,
		width=\linewidth,
		grid=major,
		legend pos=south east,
		legend style={inner xsep=0pt, inner ysep=0pt, font=\fontsize{1}{1}\selectfont},
		change x base,
		x SI prefix=mega,x unit=\-,
		xlabel=Probed Items,
		xmin=0,xmax=400000,
		ymin=0.2,ymax=1,
	]
			\addplot 
			coordinates 
			{
				(1.326, 0.00124)
				(2.878, 0.0018)
				(5.506, 0.00296)
				(11.817, 0.00466)
				(22.891, 0.00787999)
				(43.126, 0.01406)
				(94.608, 0.0242401)
				(182.383, 0.0383201)
				(329.976, 0.0611802)
				(627.04, 0.09448)
				(1457.21, 0.13626)
				(2637.62, 0.19248)
				(4909.55, 0.25284)
				(9502.59, 0.32468)
				(18467.5, 0.39686)
				(35509.0, 0.4776)
				(70174.0, 0.56368)
				(139202.0, 0.65324)
				(275613.0, 0.737101)
				(547027.0, 0.823061)
				(1075300.0, 0.90558)
				(2109100.0, 0.98864)
				(2340370.0, 1.0)
			};
			\addlegendentry{Simple-LSH}
			\addplot 
			coordinates 
			{
				(1.475, 0.00136)
				(2.645, 0.00264)
				(5.067, 0.00458)
				(9.445, 0.00759999)
				(18.566, 0.01312)
				(36.567, 0.0227201)
				(70.495, 0.0380601)
				(137.223, 0.0621402)
				(268.503, 0.10214)
				(533.737, 0.15332)
				(1053.28, 0.22386)
				(2094.42, 0.31484)
				(4156.38, 0.41692)
				(8276.37, 0.53168)
				(16491.9, 0.64922)
				(32891.9, 0.755521)
				(65606.2, 0.84922)
				(131075.0, 0.97216)
				(262145.0, 0.99258)
				(524289.0, 0.99896)
				(1048580.0, 0.99992)
				(2097150.0, 1.0)
				(2340370.0, 1.0)
			};
			\addlegendentry{Simple-LSH-NR}
	\end{axis}
	\end{tikzpicture}	
	\end{subfigure}%
	\hspace{\figuregap}	
	\begin{subfigure}[b]{\figurepercentperrow\textwidth}
			\begin{tikzpicture}
			\begin{axis}[
			height=\figurewidth\linewidth,
			width=\linewidth,
			legend pos=south east,
			legend style={inner xsep=0pt, inner ysep=0pt, font=\fontsize{1}{1}\selectfont},
			grid=major,
			change x base,
			x SI prefix=mega,x unit=\-,
			xlabel=Probed Items ,
			xmin=0,xmax=400000,
			ymin=0.2,ymax=1,
			]
				\addplot 
				coordinates 
				{
					(1.905, 0.0011)
					(3.582, 0.00212)
					(7.059, 0.00374)
					(14.287, 0.00638)
					(53.252, 0.01018)
					(107.651, 0.0169)
					(375.693, 0.0283001)
					(543.844, 0.0443602)
					(852.903, 0.0656602)
					(1195.74, 0.09554)
					(2421.62, 0.13444)
					(4124.82, 0.1814)
					(6942.04, 0.23654)
					(12209.2, 0.30028)
					(22186.1, 0.36746)
					(39859.1, 0.43834)
					(75959.1, 0.52084)
					(142581.0, 0.5996)
					(277795.0, 0.6959)
					(545277.0, 0.794901)
					(1068470.0, 0.88838)
					(2114110.0, 0.98346)
					(2340370.0, 1.0)
				};
				\addlegendentry{Sign-ALSH}
				\addplot 
				coordinates 
				{
					(5.614, 0.00282)
					(8.15, 0.00386)
					(10.784, 0.00496)
					(15.921, 0.0075)
					(26.732, 0.01218)
					(45.73, 0.0179401)
					(80.322, 0.0285801)
					(151.61, 0.0502402)
					(285.075, 0.083)
					(539.527, 0.12964)
					(1056.74, 0.2036)
					(2082.75, 0.30294)
					(4132.04, 0.425)
					(8223.75, 0.55212)
					(16426.8, 0.67132)
					(32818.2, 0.77492)
					(65566.8, 0.87838)
					(131077.0, 0.97164)
					(262146.0, 0.99174)
					(524289.0, 0.99818)
					(1048580.0, 0.99972)
				};
				\addlegendentry{Sign-ALSH-NR}
			\end{axis}
			\end{tikzpicture}	
		\end{subfigure}%
	
	\caption{Probed item-recall of top-50 items comparison between the original meta algorithms and their norm-range versions under a code length of 32. From top row to bottom row, the datasets are Netflix, Yahoo!Music and ImageNet, respectively.} 
	\label{top50-bit32:Norm-Range comparision}
\end{figure*}

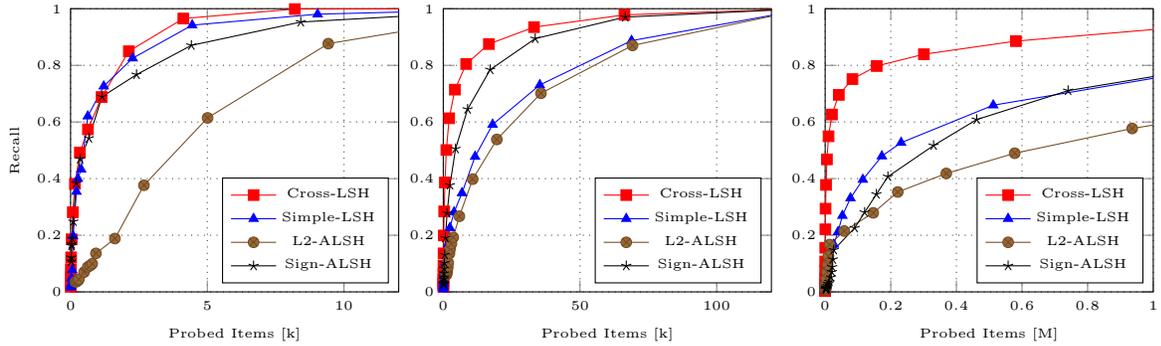
\begin{figure*}[t]
		\begin{subfigure}[b]{0.36\textwidth}
			\begin{tikzpicture}
			\begin{axis}[
			height=\figurewidth\linewidth,
			width=\linewidth,
			legend style={font=\fontsize{1}{1}\selectfont},
			grid=major,
			legend pos=south east,
			change x base,
			x SI prefix=kilo,x unit=\-,
			xlabel=Probed Items , ylabel=Recall,
			xmin=0,xmax=12000,
			ymin=0,ymax=1,
			]
					\addplot [mark=square*, red]
					coordinates 
					{
						(1.956, 0.01805)
						(3.367, 0.0320999)
						(5.745, 0.0514498)
						(10.746, 0.08005)
						(19.911, 0.122)
						(38.253, 0.1857)
						(82.004, 0.2811)
						(158.274, 0.38125)
						(339.856, 0.49175)
						(634.608, 0.57395)
						(1135.33, 0.688001)
						(2130.34, 0.849702)
						(4123.87, 0.965251)
						(8201.29, 0.99875)
						(16385.4, 1.0)
						(17770.0, 1.0)
					};
					\addlegendentry{Cross-LSH}
					
					\addplot   [mark=triangle*, blue]
					coordinates 
					{
						(15.513, 0.0164)
						(18.668, 0.0217)
						(21.116, 0.02855)
						(31.15, 0.04785)
						(60.97, 0.07755)
						(94.347, 0.19705)
						(213.601, 0.3543)
						(267.564, 0.3998)
						(386.977, 0.43265)
						(629.497, 0.6203)
						(1212.48, 0.72625)
						(2270.57, 0.825151)
						(4447.86, 0.941951)
						(9035.15, 0.98065)
						(16831.0, 0.99985)
						(17770.0, 1.0)
					};
					\addlegendentry{Simple-LSH}
					
					\addplot 
					coordinates 
					{
						(173.155, 0.03565)
						(216.145, 0.03615)
						(266.201, 0.04)
						(285.825, 0.04275)
						(301.2, 0.0445)
						(495.033, 0.0697)
						(630.641, 0.08805)
						(717.695, 0.0931)
						(790.817, 0.09935)
						(927.454, 0.1359)
						(1621.28, 0.18835)
						(2682.08, 0.3765)
						(5010.2, 0.61435)
						(9424.08, 0.876601)
						(16798.8, 0.9944)
						(17770.0, 1.0)
					};
					\addlegendentry{L2-ALSH}

					\addplot 
					coordinates 
					{
						(23.498, 0.1073)
						(25.979, 0.12045)
						(26.811, 0.12135)
						(36.265, 0.16215)
						(39.49, 0.168)
						(57.714, 0.1875)
						(101.628, 0.24845)
						(206.079, 0.38355)
						(348.819, 0.46925)
						(668.16, 0.54265)
						(1147.64, 0.689101)
						(2411.45, 0.767201)
						(4417.08, 0.870501)
						(8418.7, 0.952901)
						(16487.3, 0.9972)
						(17770.0, 1.0)
					};
					\addlegendentry{Sign-ALSH}

			\end{axis}
			\end{tikzpicture}
	\end{subfigure}%
	\hspace{\figuregap}	
	\begin{subfigure}[b]{0.36\textwidth}
			\begin{tikzpicture}
			\begin{axis}[
			height=\figurewidth\linewidth,
			width=\linewidth,
			legend style={font=\fontsize{1}{1}\selectfont},
			grid=major,
			legend pos=south east,
			change x base,
			x SI prefix=kilo,x unit=\-,
			xlabel=Probed Items ,
			xmin=0,xmax=120000,
			ymin=0,ymax=1,
			]
				\addplot [mark=square*, red]
				coordinates 
				{
					(10.313, 0.0183)
					(12.6, 0.0242)
					(17.367, 0.03075)
					(24.514, 0.0443999)
					(35.813, 0.0642999)
					(56.911, 0.0929501)
					(95.79, 0.1359)
					(171.866, 0.19925)
					(324.802, 0.2841)
					(590.489, 0.38615)
					(1127.55, 0.501)
					(2156.75, 0.6131)
					(4243.88, 0.714151)
					(8366.23, 0.804751)
					(16685.9, 0.875151)
					(33191.9, 0.934901)
					(66264.7, 0.97815)
					(131335.0, 0.99995)
					(136736.0, 1.0)
				};
				\addlegendentry{Cross-LSH}
				
				\addplot   [mark=triangle*, blue]
				coordinates 
				{
					(32.855, 0.00935001)
					(44.528, 0.0159)
					(57.224, 0.02145)
					(83.074, 0.02785)
					(127.262, 0.03115)
					(181.652, 0.0415999)
					(341.817, 0.0568999)
					(422.161, 0.0721999)
					(847.777, 0.10395)
					(1798.92, 0.17015)
					(2409.31, 0.22605)
					(3946.3, 0.28135)
					(6844.82, 0.3492)
					(11626.1, 0.4778)
					(18036.4, 0.590901)
					(35302.1, 0.730651)
					(68860.8, 0.887151)
					(132410.0, 0.99805)
					(136736.0, 1.0)
				};
				\addlegendentry{Simple-LSH}
				
				\addplot 
				coordinates 
				{
					(1088.27, 0.0605)
					(1147.36, 0.0619499)
					(1183.56, 0.0641499)
					(1339.79, 0.071)
					(1411.06, 0.0751)
					(1501.89, 0.08295)
					(1599.19, 0.0871)
					(1817.7, 0.10265)
					(2252.74, 0.13545)
					(2460.2, 0.15085)
					(2975.95, 0.16985)
					(3633.33, 0.1923)
					(5887.09, 0.267)
					(10766.5, 0.398)
					(19597.1, 0.53825)
					(35730.6, 0.701301)
					(69198.2, 0.870201)
					(132503.0, 0.99885)
					(136736.0, 1.0)
				};
				\addlegendentry{L2-ALSH}

				\addplot 
				coordinates 
				{
					(158.715, 0.02535)
					(163.964, 0.028)
					(170.62, 0.0317)
					(206.118, 0.0449999)
					(219.67, 0.0515999)
					(247.039, 0.0652999)
					(294.157, 0.08215)
					(404.002, 0.1021)
					(524.98, 0.131)
					(778.893, 0.19055)
					(1396.94, 0.2778)
					(2439.14, 0.3769)
					(4490.64, 0.5052)
					(8881.75, 0.644951)
					(17146.1, 0.784601)
					(33612.5, 0.894351)
					(66555.2, 0.970351)
					(131533.0, 0.9996)
					(136736.0, 1.0)
				};
				\addlegendentry{Sign-ALSH}

			\end{axis}
			\end{tikzpicture}
	\end{subfigure}%
	\hspace{\figuregap}	
	\begin{subfigure}[b]{0.36\textwidth}
			\begin{tikzpicture}
			\begin{axis}[
			height=\figurewidth\linewidth,
			width=\linewidth,
			legend style={font=\fontsize{1}{1}\selectfont},
			grid=major,
			legend pos=south east,
			change x base,
			x SI prefix=mega,x unit=\-,
			xlabel=Probed Items ,
			xmin=0,xmax=1000000,
			ymin=0,ymax=1,
			]
			
				\addplot [mark=square*, red]
				coordinates 
				{
					(2.322, 0.00275)
					(3.551, 0.0057)
					(6.743, 0.01085)
					(12.483, 0.01665)
					(22.075, 0.02865)
					(44.082, 0.0471499)
					(82.993, 0.0737)
					(155.21, 0.1075)
					(298.75, 0.1557)
					(674.045, 0.22095)
					(1262.74, 0.29375)
					(2713.62, 0.37775)
					(5400.41, 0.4679)
					(10290.0, 0.54945)
					(20632.6, 0.62635)
					(41640.8, 0.69575)
					(83166.0, 0.751401)
					(157142.0, 0.79755)
					(300941.0, 0.8392)
					(581439.0, 0.885501)
					(1107000.0, 0.936851)
				};
				\addlegendentry{Cross-LSH}
				
				\addplot   [mark=triangle*, blue]
				coordinates 
				{
					(118.499, 0.00485)
					(131.758, 0.0065)
					(152.118, 0.01125)
					(321.233, 0.0144)
					(354.683, 0.0191)
					(449.145, 0.02425)
					(963.481, 0.0371)
					(2292.92, 0.0502999)
					(4571.21, 0.0685999)
					(8494.07, 0.09225)
					(13546.1, 0.12405)
					(26213.0, 0.16735)
					(39204.5, 0.21155)
					(52537.0, 0.26915)
					(77151.2, 0.3311)
					(114839.0, 0.39655)
					(172804.0, 0.47935)
					(231570.0, 0.527)
					(512385.0, 0.659201)
					(1050260.0, 0.763851)
					(1608640.0, 0.798601)
					(2154750.0, 0.958901)
					(2340370.0, 1.0)
				};
				\addlegendentry{Simple-LSH}
				
				\addplot 
				coordinates 
				{
					(714.924, 0.0068)
					(800.768, 0.01035)
					(853.377, 0.01095)
					(967.017, 0.0121)
					(1021.38, 0.0132)
					(1268.98, 0.0188)
					(1861.18, 0.0239)
					(3569.72, 0.0303)
					(4042.46, 0.0466999)
					(5380.81, 0.0696499)
					(7590.95, 0.09605)
					(9859.17, 0.1265)
					(13582.6, 0.16705)
					(58068.6, 0.2149)
					(146516.0, 0.279)
					(221358.0, 0.3527)
					(369174.0, 0.4182)
					(577747.0, 0.48965)
					(935814.0, 0.57705)
					(1185970.0, 0.6238)
					(1441520.0, 0.6534)
					(2138950.0, 0.868852)
					(2340370.0, 1.0)
				};
				\addlegendentry{L2-ALSH}
				
				\addplot 
				coordinates 
				{
					(2537.5, 0.0089)
					(2554.65, 0.0095)
					(2728.69, 0.01225)
					(2963.13, 0.01395)
					(3554.03, 0.0159)
					(6373.82, 0.0208)
					(9874.68, 0.028)
					(12628.7, 0.0353999)
					(17910.5, 0.0512999)
					(20426.8, 0.07)
					(21519.7, 0.0875501)
					(23280.3, 0.1147)
					(26085.5, 0.148)
					(89933.9, 0.22575)
					(120251.0, 0.2808)
					(156148.0, 0.3448)
					(191407.0, 0.407)
					(330493.0, 0.51695)
					(461998.0, 0.608501)
					(740518.0, 0.710701)
					(1234490.0, 0.805201)
					(2154130.0, 0.978301)
					(2340370.0, 1.0)
				};
				\addlegendentry{Sign-ALSH}

			\end{axis}
			\end{tikzpicture}	
	\end{subfigure}%
	\caption{Probed item-recall of top-20 items comparison between Cross-LSH and existing algorithms under a code length of 16 (best viewed in color). From left to right, the datasets are Netflix, Yahoo!Music and ImageNet, respectively.} 
	\label{top20-bit16:Cross comparision}
\end{figure*}  

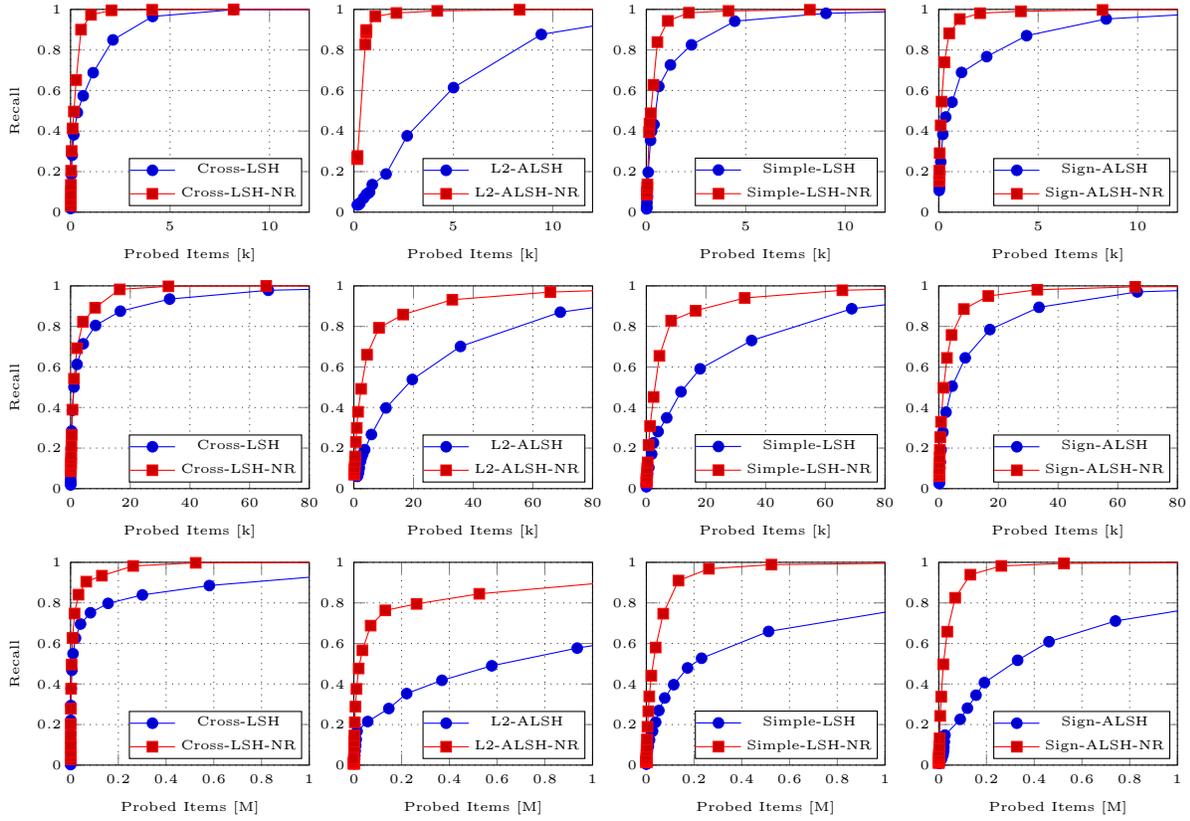
\begin{figure*}[t]
		\begin{subfigure}[b]{\figurepercentperrow\textwidth}
			\begin{tikzpicture}
			\begin{axis}[
			height=\figurewidth\linewidth,
			width=\linewidth,
			grid=major,
			legend pos=south east,
			legend style={inner xsep=0pt, inner ysep=0pt, font=\fontsize{1}{1}\selectfont},
			change x base,
			x SI prefix=kilo,x unit=\-,
			xlabel=Probed Items , ylabel=Recall,
			xmin=0,xmax=12000,
			ymin=0,ymax=1,
			mark size=2.0pt,
			]

				\addplot 
				coordinates 
				{
					(1.956, 0.01805)
					(3.367, 0.0320999)
					(5.745, 0.0514498)
					(10.746, 0.08005)
					(19.911, 0.122)
					(38.253, 0.1857)
					(82.004, 0.2811)
					(158.274, 0.38125)
					(339.856, 0.49175)
					(634.608, 0.57395)
					(1135.33, 0.688001)
					(2130.34, 0.849702)
					(4123.87, 0.965251)
					(8201.29, 0.99875)
					(16385.4, 1.0)
					(17770.0, 1.0)
				};
				\addlegendentry{Cross-LSH}
				\addplot 
				coordinates 
				{
					(3.024, 0.0316999)
					(6.207, 0.0588498)
					(10.52, 0.0880501)
					(16.946, 0.13485)
					(27.22, 0.2043)
					(51.886, 0.3017)
					(103.883, 0.413)
					(167.395, 0.4966)
					(280.491, 0.651801)
					(525.52, 0.900101)
					(1026.7, 0.973151)
					(2056.44, 0.99555)
					(4101.34, 0.9982)
					(8196.02, 0.9998)
					(16384.7, 1.0)
					(17770.0, 1.0)
				};
				\addlegendentry{Cross-LSH-NR}
					
			\end{axis}
			\end{tikzpicture}
	\end{subfigure}%
	\hspace{\figuregap}
	\begin{subfigure}[b]{\figurepercentperrow\textwidth}
	\begin{tikzpicture}
	\begin{axis}[
	height=\figurewidth\linewidth,
	width=\linewidth,
	legend pos=south east,
	legend style={inner xsep=0pt, inner ysep=0pt, font=\fontsize{1}{1}\selectfont},
	grid=major,
	legend pos=south east,
	change x base,
	x SI prefix=kilo,x unit=\-,
	xlabel=Probed Items ,
	xmin=0,xmax=12000,
	ymin=0,ymax=1,
	]
		\addplot 
		coordinates 
		{
			(173.155, 0.03565)
			(216.145, 0.03615)
			(266.201, 0.04)
			(285.825, 0.04275)
			(301.2, 0.0445)
			(495.033, 0.0697)
			(630.641, 0.08805)
			(717.695, 0.0931)
			(790.817, 0.09935)
			(927.454, 0.1359)
			(1621.28, 0.18835)
			(2682.08, 0.3765)
			(5010.2, 0.61435)
			(9424.08, 0.876601)
			(16798.8, 0.9944)
			(17770.0, 1.0)
		};
		\addlegendentry{L2-ALSH}
		\addplot 
		coordinates 
		{
			(174.257, 0.26165)
			(177.811, 0.26555)
			(180.762, 0.26825)
			(181.077, 0.2685)
			(181.242, 0.26885)
			(187.181, 0.2771)
			(575.934, 0.827151)
			(621.659, 0.886902)
			(632.032, 0.895752)
			(636.977, 0.898602)
			(1082.75, 0.965851)
			(2140.87, 0.9824)
			(4192.03, 0.9926)
			(8329.45, 0.99845)
			(16466.9, 1.0)
			(17770.0, 1.0)
		};
		\addlegendentry{L2-ALSH-NR}
	\end{axis}
	\end{tikzpicture}
	\end{subfigure}%
	\hspace{\figuregap}
	\begin{subfigure}[b]{\figurepercentperrow\textwidth}
	\begin{tikzpicture}
	\begin{axis}[
	height=\figurewidth\linewidth,
	width=\linewidth,
	legend pos=south east,
	legend style={inner xsep=0pt, inner ysep=0pt, font=\fontsize{1}{1}\selectfont},
	grid=major,
	change x base,
	x SI prefix=kilo,x unit=\-,
	xlabel=Probed Items,
	xmin=0,xmax=12000,
	ymin=0,ymax=1,
	]
		\addplot 
		coordinates 
		{
			(15.513, 0.0164)
			(18.668, 0.0217)
			(21.116, 0.02855)
			(31.15, 0.04785)
			(60.97, 0.07755)
			(94.347, 0.19705)
			(213.601, 0.3543)
			(267.564, 0.3998)
			(386.977, 0.43265)
			(629.497, 0.6203)
			(1212.48, 0.72625)
			(2270.57, 0.825151)
			(4447.86, 0.941951)
			(9035.15, 0.98065)
			(16831.0, 0.99985)
			(17770.0, 1.0)
		};
		\addlegendentry{Simple-LSH}
		\addplot 
		coordinates 
		{
			(23.583, 0.0866)
			(25.701, 0.0915001)
			(35.793, 0.10025)
			(50.01, 0.12215)
			(54.875, 0.1368)
			(138.589, 0.39455)
			(161.029, 0.437)
			(225.354, 0.48785)
			(355.136, 0.62795)
			(558.187, 0.838651)
			(1083.92, 0.943301)
			(2139.73, 0.983551)
			(4139.72, 0.99245)
			(8223.63, 0.99815)
			(16407.0, 0.99995)
			(17770.0, 1.0)
		};
		\addlegendentry{Simple-LSH-NR}
	
	\end{axis}
	\end{tikzpicture}
	\end{subfigure}%	
	\hspace{\figuregap}	
	\begin{subfigure}[b]{\figurepercentperrow\textwidth}
		\begin{tikzpicture}
		\begin{axis}[
		height=\figurewidth\linewidth,
		width=\linewidth,
		legend pos=south east,
		legend style={inner xsep=0pt, inner ysep=0pt, font=\fontsize{1}{1}\selectfont},
		grid=major,
		legend pos=south east,
		change x base,
		x SI prefix=kilo,x unit=\-,
		xlabel=Probed Items ,
		xmin=0,xmax=12000,
		ymin=0,ymax=1,
		]
			\addplot 
			coordinates 
			{
				(23.498, 0.1073)
				(25.979, 0.12045)
				(26.811, 0.12135)
				(36.265, 0.16215)
				(39.49, 0.168)
				(57.714, 0.1875)
				(101.628, 0.24845)
				(206.079, 0.38355)
				(348.819, 0.46925)
				(668.16, 0.54265)
				(1147.64, 0.689101)
				(2411.45, 0.767201)
				(4417.08, 0.870501)
				(8418.7, 0.952901)
				(16487.3, 0.9972)
				(17770.0, 1.0)
			};
			\addlegendentry{Sign-ALSH}
			\addplot 
			coordinates 
			{
				(21.784, 0.15305)
				(22.282, 0.15585)
				(26.058, 0.1845)
				(26.487, 0.18685)
				(32.184, 0.20475)
				(47.963, 0.2911)
				(98.545, 0.42815)
				(143.799, 0.5453)
				(283.166, 0.739151)
				(525.79, 0.882001)
				(1058.78, 0.952502)
				(2080.53, 0.981351)
				(4131.58, 0.9907)
				(8231.52, 0.9974)
				(16441.3, 1.0)
				(17770.0, 1.0)
			};
			\addlegendentry{Sign-ALSH-NR}
		\end{axis}
		\end{tikzpicture}
	\end{subfigure}%
	
		\begin{subfigure}[b]{\figurepercentperrow\textwidth}
			\begin{tikzpicture}
			\begin{axis}[
			height=\figurewidth\linewidth,
			width=\linewidth,
			grid=major,
			legend pos=south east,
			legend style={inner xsep=0pt, inner ysep=0pt, font=\fontsize{1}{1}\selectfont},
			change x base,
			x SI prefix=kilo,x unit=\-,
			xmin=0,xmax=80000,
			ymin=0,ymax=1,
			xlabel=Probed Items , ylabel=Recall,
			]
				\addplot 
				coordinates 
				{
					(10.313, 0.0183)
					(12.6, 0.0242)
					(17.367, 0.03075)
					(24.514, 0.0443999)
					(35.813, 0.0642999)
					(56.911, 0.0929501)
					(95.79, 0.1359)
					(171.866, 0.19925)
					(324.802, 0.2841)
					(590.489, 0.38615)
					(1127.55, 0.501)
					(2156.75, 0.6131)
					(4243.88, 0.714151)
					(8366.23, 0.804751)
					(16685.9, 0.875151)
					(33191.9, 0.934901)
					(66264.7, 0.97815)
					(131335.0, 0.99995)
					(136736.0, 1.0)
				};
				\addlegendentry{Cross-LSH}
				\addplot 
				coordinates 
				{
					(95.733, 0.08585)
					(103.78, 0.0942001)
					(118.671, 0.10695)
					(135.472, 0.11815)
					(146.69, 0.1267)
					(170.121, 0.14865)
					(198.236, 0.16825)
					(266.859, 0.20995)
					(371.964, 0.2645)
					(616.947, 0.38945)
					(1145.93, 0.541901)
					(2146.61, 0.692651)
					(4153.44, 0.822951)
					(8209.48, 0.892301)
					(16467.8, 0.9828)
					(32808.7, 0.99765)
					(65565.3, 0.99975)
					(132161.0, 1.0)
					(136736.0, 1.0)
				};
				\addlegendentry{Cross-LSH-NR}
			\end{axis}
			\end{tikzpicture}
	\end{subfigure}%
	\hspace{\figuregap}
	\begin{subfigure}[b]{\figurepercentperrow\textwidth}
	\begin{tikzpicture}
	\begin{axis}[
	height=\figurewidth\linewidth,
	width=\linewidth,
	legend pos=south east,
	legend style={inner xsep=0pt, inner ysep=0pt, font=\fontsize{1}{1}\selectfont},
	grid=major,
	change x base,
	x SI prefix=kilo,x unit=\-,
	xlabel=Probed Items,
	xmin=0,xmax=80000,
	ymin=0,ymax=1,
	]

			\addplot 
			coordinates 
			{
				(1088.27, 0.0605)
				(1147.36, 0.0619499)
				(1183.56, 0.0641499)
				(1339.79, 0.071)
				(1411.06, 0.0751)
				(1501.89, 0.08295)
				(1599.19, 0.0871)
				(1817.7, 0.10265)
				(2252.74, 0.13545)
				(2460.2, 0.15085)
				(2975.95, 0.16985)
				(3633.33, 0.1923)
				(5887.09, 0.267)
				(10766.5, 0.398)
				(19597.1, 0.53825)
				(35730.6, 0.701301)
				(69198.2, 0.870201)
				(132503.0, 0.99885)
				(136736.0, 1.0)
			};
			\addlegendentry{L2-ALSH}
			\addplot 
			coordinates 
			{
				(184.721, 0.0687999)
				(190.335, 0.0709499)
				(197.547, 0.0742999)
				(215.898, 0.0845999)
				(251.789, 0.09425)
				(278.085, 0.1056)
				(380.62, 0.13255)
				(442.489, 0.1588)
				(699.575, 0.2304)
				(979.785, 0.2995)
				(1407.26, 0.37885)
				(2420.77, 0.49205)
				(4421.32, 0.6604)
				(8453.78, 0.793201)
				(16528.4, 0.858651)
				(33003.6, 0.931801)
				(65878.2, 0.969501)
				(131148.0, 0.9993)
				(136736.0, 1.0)
			};
			\addlegendentry{L2-ALSH-NR}
	\end{axis}
	\end{tikzpicture}
\end{subfigure}%
	\hspace{\figuregap}
	\begin{subfigure}[b]{\figurepercentperrow\textwidth}
	\begin{tikzpicture}
	\begin{axis}[
	height=\figurewidth\linewidth,
	width=\linewidth,
	legend pos=south east,
	legend style={inner xsep=0pt, inner ysep=0pt, font=\fontsize{1}{1}\selectfont},
	grid=major,
	change x base,
	x SI prefix=kilo,x unit=\-,
	xlabel=Probed Items , 
	xmin=0,xmax=80000,
	ymin=0,ymax=1,
	]
			\addplot 
			coordinates 
			{
				(32.855, 0.00935001)
				(44.528, 0.0159)
				(57.224, 0.02145)
				(83.074, 0.02785)
				(127.262, 0.03115)
				(181.652, 0.0415999)
				(341.817, 0.0568999)
				(422.161, 0.0721999)
				(847.777, 0.10395)
				(1798.92, 0.17015)
				(2409.31, 0.22605)
				(3946.3, 0.28135)
				(6844.82, 0.3492)
				(11626.1, 0.4778)
				(18036.4, 0.590901)
				(35302.1, 0.730651)
				(68860.8, 0.887151)
				(132410.0, 0.99805)
				(136736.0, 1.0)
			};
			\addlegendentry{Simple-LSH}
			\addplot 
			coordinates 
			{
				(57.399, 0.03115)
				(64.112, 0.0351)
				(85.481, 0.0449999)
				(103.374, 0.0482999)
				(105.89, 0.0499999)
				(150.749, 0.0598999)
				(254.997, 0.0809499)
				(310.095, 0.0992001)
				(449.95, 0.12935)
				(770.635, 0.21495)
				(1270.41, 0.30855)
				(2438.89, 0.45215)
				(4400.4, 0.654751)
				(8330.88, 0.828101)
				(16533.9, 0.877451)
				(32989.3, 0.940401)
				(65721.5, 0.978351)
				(131206.0, 0.99935)
				(136736.0, 1.0)
			};
			\addlegendentry{Simple-LSH-NR}
					
	\end{axis}
	\end{tikzpicture}
\end{subfigure}%
	\hspace{\figuregap}
	\begin{subfigure}[b]{\figurepercentperrow\textwidth}
		\begin{tikzpicture}
		\begin{axis}[
		height=\figurewidth\linewidth,
		width=\linewidth,
		legend pos=south east,
		legend style={inner xsep=0pt, inner ysep=0pt, font=\fontsize{1}{1}\selectfont},
		grid=major,
		change x base,
		x SI prefix=kilo,x unit=\-,
		xlabel=Probed Items ,
		xmin=0,xmax=80000,
		ymin=0,ymax=1,
		]
		
				\addplot 
				coordinates 
				{
					(158.715, 0.02535)
					(163.964, 0.028)
					(170.62, 0.0317)
					(206.118, 0.0449999)
					(219.67, 0.0515999)
					(247.039, 0.0652999)
					(294.157, 0.08215)
					(404.002, 0.1021)
					(524.98, 0.131)
					(778.893, 0.19055)
					(1396.94, 0.2778)
					(2439.14, 0.3769)
					(4490.64, 0.5052)
					(8881.75, 0.644951)
					(17146.1, 0.784601)
					(33612.5, 0.894351)
					(66555.2, 0.970351)
					(131533.0, 0.9996)
					(136736.0, 1.0)
				};
				\addlegendentry{Sign-ALSH}
				\addplot 
				coordinates 
				{
					(144.415, 0.0621999)
					(146.012, 0.0636499)
					(197.332, 0.08025)
					(206.418, 0.0927001)
					(215.459, 0.0975001)
					(259.344, 0.1262)
					(318.312, 0.15505)
					(379.227, 0.19165)
					(524.998, 0.25385)
					(770.722, 0.32955)
					(1620.42, 0.4973)
					(2749.15, 0.644651)
					(4308.87, 0.757751)
					(8430.68, 0.885602)
					(16604.1, 0.950001)
					(33039.9, 0.980951)
					(65838.5, 0.9954)
					(131185.0, 0.99975)
					(136736.0, 1.0)
				};
				\addlegendentry{Sign-ALSH-NR}
		
		\end{axis}
		\end{tikzpicture}
	\end{subfigure}%
	
		\begin{subfigure}[b]{\figurepercentperrow\textwidth}
			\begin{tikzpicture}
			\begin{axis}[
				height=\figurewidth\linewidth,
				width=\linewidth,
				grid=major,
				legend pos=south east,
				legend style={inner xsep=0pt, inner ysep=0pt, font=\fontsize{1}{1}\selectfont},
				change x base,
				x SI prefix=mega,x unit=\-,
				xmin=0,xmax=1000000,
				ymin=0,ymax=1,
				xlabel=Probed Items , ylabel=Recall,
			]
					\addplot 
					coordinates 
					{
						(2.322, 0.00275)
						(3.551, 0.0057)
						(6.743, 0.01085)
						(12.483, 0.01665)
						(22.075, 0.02865)
						(44.082, 0.0471499)
						(82.993, 0.0737)
						(155.21, 0.1075)
						(298.75, 0.1557)
						(674.045, 0.22095)
						(1262.74, 0.29375)
						(2713.62, 0.37775)
						(5400.41, 0.4679)
						(10290.0, 0.54945)
						(20632.6, 0.62635)
						(41640.8, 0.69575)
						(83166.0, 0.751401)
						(157142.0, 0.79755)
						(300941.0, 0.8392)
						(581439.0, 0.885501)
						(1107000.0, 0.936851)
					};
					\addlegendentry{Cross-LSH}
					\addplot 
					coordinates 
					{
						(28.804, 0.02825)
						(55.705, 0.0427499)
						(96.599, 0.0638999)
						(177.157, 0.0990501)
						(327.687, 0.1444)
						(597.816, 0.2018)
						(1113.1, 0.27805)
						(2147.16, 0.37685)
						(4181.91, 0.4961)
						(8296.2, 0.627)
						(16492.7, 0.747551)
						(32887.3, 0.840601)
						(65725.2, 0.904801)
						(131162.0, 0.934051)
						(262145.0, 0.98215)
						(524288.0, 0.9975)
						(1048580.0, 0.99985)
						(2097150.0, 1.0)
						(2340370.0, 1.0)
					};
					\addlegendentry{Cross-LSH-NR}

			\end{axis}
			\end{tikzpicture}	
	\end{subfigure}%
	\hspace{\figuregap}
	\begin{subfigure}[b]{\figurepercentperrow\textwidth}
		\begin{tikzpicture}
		\begin{axis}[
		height=\figurewidth\linewidth,
		width=\linewidth,
		grid=major,
		legend pos=south east,
		legend style={inner xsep=0pt, inner ysep=0pt, font=\fontsize{1}{1}\selectfont},
		change x base,
		x SI prefix=mega,x unit=\-,
		xlabel=Probed Items ,
		xmin=0,xmax=1000000,
		ymin=0,ymax=1,
		]
		
				\addplot 
				coordinates 
				{
					(714.924, 0.0068)
					(800.768, 0.01035)
					(853.377, 0.01095)
					(967.017, 0.0121)
					(1021.38, 0.0132)
					(1268.98, 0.0188)
					(1861.18, 0.0239)
					(3569.72, 0.0303)
					(4042.46, 0.0466999)
					(5380.81, 0.0696499)
					(7590.95, 0.09605)
					(9859.17, 0.1265)
					(13582.6, 0.16705)
					(58068.6, 0.2149)
					(146516.0, 0.279)
					(221358.0, 0.3527)
					(369174.0, 0.4182)
					(577747.0, 0.48965)
					(935814.0, 0.57705)
					(1185970.0, 0.6238)
					(1441520.0, 0.6534)
					(2138950.0, 0.868852)
					(2340370.0, 1.0)
				};
				\addlegendentry{L2-ALSH}
				\addplot 
				coordinates 
				{
					(96.71, 0.0062)
					(101.938, 0.0069)
					(108.392, 0.0079)
					(124.176, 0.00905)
					(139.602, 0.0109)
					(199.73, 0.01365)
					(328.312, 0.0235)
					(471.589, 0.03305)
					(1197.15, 0.08925)
					(1573.92, 0.1159)
					(2233.66, 0.14595)
					(3911.18, 0.21125)
					(6770.96, 0.28845)
					(11111.8, 0.37545)
					(20245.2, 0.4766)
					(35980.8, 0.5663)
					(70452.8, 0.6879)
					(131917.0, 0.763101)
					(262908.0, 0.7951)
					(525475.0, 0.84475)
					(1049490.0, 0.899201)
					(2097690.0, 0.925751)
					(2340370.0, 1.0)
				};
				\addlegendentry{L2-ALSH-NR}
		\end{axis}
		\end{tikzpicture}	
	\end{subfigure}%
	\hspace{\figuregap}	
	\begin{subfigure}[b]{\figurepercentperrow\textwidth}
	\begin{tikzpicture}
	\begin{axis}[
		height=\figurewidth\linewidth,
		width=\linewidth,
		grid=major,
		legend pos=south east,
		legend style={inner xsep=0pt, inner ysep=0pt, font=\fontsize{1}{1}\selectfont},
		change x base,
		x SI prefix=mega,x unit=\-,
		xlabel=Probed Items,
		xmin=0,xmax=1000000,
		ymin=0,ymax=1,
	]
			\addplot 
			coordinates 
			{
				(118.499, 0.00485)
				(131.758, 0.0065)
				(152.118, 0.01125)
				(321.233, 0.0144)
				(354.683, 0.0191)
				(449.145, 0.02425)
				(963.481, 0.0371)
				(2292.92, 0.0502999)
				(4571.21, 0.0685999)
				(8494.07, 0.09225)
				(13546.1, 0.12405)
				(26213.0, 0.16735)
				(39204.5, 0.21155)
				(52537.0, 0.26915)
				(77151.2, 0.3311)
				(114839.0, 0.39655)
				(172804.0, 0.47935)
				(231570.0, 0.527)
				(512385.0, 0.659201)
				(1050260.0, 0.763851)
				(1608640.0, 0.798601)
				(2154750.0, 0.958901)
				(2340370.0, 1.0)
			};
			\addlegendentry{Simple-LSH}
			\addplot 
			coordinates 
			{
				(367.212, 0.0127)
				(380.829, 0.0136)
				(403.341, 0.01525)
				(421.872, 0.01865)
				(432.645, 0.02065)
				(458.147, 0.0274)
				(556.891, 0.0345)
				(789.681, 0.0496499)
				(1070.24, 0.07275)
				(1571.88, 0.09095)
				(2752.64, 0.1262)
				(4878.38, 0.1889)
				(8502.14, 0.26565)
				(12432.3, 0.3388)
				(20995.0, 0.44085)
				(39105.4, 0.579851)
				(70918.4, 0.746751)
				(134839.0, 0.910051)
				(262296.0, 0.968201)
				(524559.0, 0.988601)
				(1048890.0, 0.99555)
				(2097620.0, 0.9979)
				(2340370.0, 1.0)
			};
			\addlegendentry{Simple-LSH-NR}
	\end{axis}
	\end{tikzpicture}	
	\end{subfigure}%
	\hspace{\figuregap}	
	\begin{subfigure}[b]{\figurepercentperrow\textwidth}
			\begin{tikzpicture}
			\begin{axis}[
			height=\figurewidth\linewidth,
			width=\linewidth,
			legend pos=south east,
			legend style={inner xsep=0pt, inner ysep=0pt, font=\fontsize{1}{1}\selectfont},
			grid=major,
			change x base,
			x SI prefix=mega,x unit=\-,
			xlabel=Probed Items ,
			xmin=0,xmax=1000000,
			ymin=0,ymax=1,
			]
					\addplot 
					coordinates 
					{
						(2537.5, 0.0089)
						(2554.65, 0.0095)
						(2728.69, 0.01225)
						(2963.13, 0.01395)
						(3554.03, 0.0159)
						(6373.82, 0.0208)
						(9874.68, 0.028)
						(12628.7, 0.0353999)
						(17910.5, 0.0512999)
						(20426.8, 0.07)
						(21519.7, 0.0875501)
						(23280.3, 0.1147)
						(26085.5, 0.148)
						(89933.9, 0.22575)
						(120251.0, 0.2808)
						(156148.0, 0.3448)
						(191407.0, 0.407)
						(330493.0, 0.51695)
						(461998.0, 0.608501)
						(740518.0, 0.710701)
						(1234490.0, 0.805201)
						(2154130.0, 0.978301)
						(2340370.0, 1.0)
					};
					\addlegendentry{Sign-ALSH}
					\addplot 
					coordinates 
					{
						(151.844, 0.01075)
						(184.679, 0.01135)
						(363.529, 0.018)
						(410.037, 0.02045)
						(470.239, 0.02365)
						(535.203, 0.0273)
						(614.764, 0.0332)
						(811.138, 0.0437999)
						(1185.69, 0.0557999)
						(1704.89, 0.0714)
						(2418.14, 0.0926001)
						(3670.23, 0.13165)
						(6178.59, 0.24255)
						(10474.6, 0.33735)
						(19159.2, 0.49705)
						(35816.7, 0.658001)
						(68900.6, 0.825201)
						(132602.0, 0.939051)
						(262443.0, 0.9823)
						(524530.0, 0.9945)
						(1048890.0, 0.99895)
						(2097480.0, 0.9996)
						(2340370.0, 1.0)
					};
					\addlegendentry{Sign-ALSH-NR}
			\end{axis}
			\end{tikzpicture}	
		\end{subfigure}%
	
	\caption{Probed item-recall of top-20 items comparison between the original meta algorithms and their norm-range versions under a code length of 16. From top row to bottom row, the datasets are Netflix, Yahoo!Music and ImageNet, respectively.} 
	\label{top20-bit16:Norm-Range comparision}
\end{figure*}

\begin{figure*}[t]
		\begin{subfigure}[b]{0.36\textwidth}
			\begin{tikzpicture}
			\begin{axis}[
			height=\figurewidth\linewidth,
			width=\linewidth,
			legend style={font=\fontsize{1}{1}\selectfont},
			grid=major,
			legend pos=south east,
			change x base,
			x SI prefix=kilo,x unit=\-,
			xlabel=Probed Items , ylabel=Recall,
			xmin=0,xmax=4000,
			ymin=0,ymax=1,
			]
						 \addplot  [mark=square*, red]
						 coordinates 
						 {
						 	(1.491, 0.0546997)
						 	(2.295, 0.0765496)
						 	(4.401, 0.12775)
						 	(8.493, 0.19905)
						 	(16.584, 0.2909)
						 	(32.474, 0.4102)
						 	(65.11, 0.5486)
						 	(131.029, 0.673751)
						 	(260.075, 0.770651)
						 	(517.461, 0.873702)
						 	(1027.45, 0.950102)
						 	(2054.4, 0.983601)
						 	(4100.19, 0.99655)
						 	(8193.16, 1.0)
						 	(16384.3, 1.0)
						 	(17770.0, 1.0)
						 };
						 \addlegendentry{Cross-LSH}
						 
						 \addplot  [mark=triangle*, blue]
						 coordinates 
						 {
						 	(1.437, 0.02565)
						 	(2.62, 0.0423499)
						 	(4.873, 0.0676499)
						 	(8.817, 0.11135)
						 	(17.166, 0.17865)
						 	(33.092, 0.2726)
						 	(65.215, 0.3874)
						 	(129.49, 0.5321)
						 	(257.717, 0.686751)
						 	(514.692, 0.804551)
						 	(1027.27, 0.890602)
						 	(2052.24, 0.946101)
						 	(4099.77, 0.980901)
						 	(8196.18, 0.99685)
						 	(16386.4, 1.0)
						 	(17770.0, 1.0)
						 };
						 \addlegendentry{Simple-LSH}
						 
						 \addplot 
						 coordinates 
						 {
						 	(1.087, 0.0233999)
						 	(2.145, 0.0423498)
						 	(4.311, 0.0781)
						 	(8.578, 0.12685)
						 	(16.94, 0.20385)
						 	(33.562, 0.3134)
						 	(67.885, 0.44245)
						 	(132.704, 0.5882)
						 	(262.463, 0.7083)
						 	(531.802, 0.808451)
						 	(1048.8, 0.878802)
						 	(2133.6, 0.936651)
						 	(4216.32, 0.969701)
						 	(8263.74, 0.9957)
						 	(16399.2, 0.99995)
						 	(17770.0, 1.0)
						 };
						 \addlegendentry{L2-ALSH}

						 \addplot 
						 coordinates 
						 {
						 	(1.167, 0.0236999)
						 	(2.502, 0.0451498)
						 	(4.515, 0.0803999)
						 	(8.663, 0.15265)
						 	(16.594, 0.2404)
						 	(32.888, 0.35705)
						 	(64.897, 0.4864)
						 	(129.3, 0.6308)
						 	(257.464, 0.7655)
						 	(513.701, 0.867552)
						 	(1025.08, 0.939251)
						 	(2049.39, 0.974501)
						 	(4097.67, 0.99325)
						 	(8193.56, 0.99895)
						 	(16387.2, 0.99995)
						 	(17770.0, 1.0)
						 };
						 \addlegendentry{Sign-ALSH}

			\end{axis}
			\end{tikzpicture}
	\end{subfigure}%
	\hspace{\figuregap}	
	\begin{subfigure}[b]{0.36\textwidth}
			\begin{tikzpicture}
			\begin{axis}[
			height=\figurewidth\linewidth,
			width=\linewidth,
			legend style={font=\fontsize{1}{1}\selectfont},
			grid=major,
			legend pos=south east,
			change x base,
			x SI prefix=kilo,x unit=\-,
			xlabel=Probed Items ,
			xmin=0,xmax=4000,
			ymin=0,ymax=1,
			]
			
					\addplot  [mark=square*, red]
					coordinates 
					{
						(1.032, 0.0238999)
						(2.024, 0.0420498)
						(4.047, 0.0711499)
						(8.084, 0.1186)
						(16.092, 0.1843)
						(32.152, 0.2724)
						(64.171, 0.38)
						(128.23, 0.50545)
						(256.276, 0.63425)
						(512.3, 0.738401)
						(1024.34, 0.827151)
						(2048.28, 0.894801)
						(4096.36, 0.940651)
					};
					\addlegendentry{Cross-LSH}
					
					\addplot  [mark=triangle*, blue]
					coordinates 
					{
						(1.08, 0.0117)
						(2.229, 0.01975)
						(4.326, 0.0354499)
						(8.441, 0.0564499)
						(16.455, 0.0869501)
						(32.696, 0.13125)
						(64.761, 0.1923)
						(128.987, 0.27365)
						(256.988, 0.3709)
						(513.2, 0.479849)
						(1025.03, 0.5867)
						(2049.32, 0.69345)
						(4097.04, 0.783801)
						(8193.31, 0.859801)
						(16386.0, 0.922851)
						(32770.4, 0.964201)
						(65538.6, 0.98645)
						(131074.0, 1.0)
						(136736.0, 1.0)
					};
					\addlegendentry{Simple-LSH}
					
					\addplot 
					coordinates 
					{
						(1.265, 0.00980002)
						(2.398, 0.01565)
						(4.699, 0.02575)
						(8.962, 0.0437999)
						(17.238, 0.0653499)
						(33.281, 0.0961)
						(65.468, 0.1427)
						(130.28, 0.20615)
						(258.902, 0.28975)
						(515.12, 0.3888)
						(1027.18, 0.5083)
						(2051.45, 0.629)
						(4100.54, 0.7422)
						(8198.43, 0.836951)
						(16392.0, 0.910301)
					};
					\addlegendentry{L2-ALSH}
					
					\addplot 
					coordinates 
					{
						(1.435, 0.00995002)
						(2.562, 0.0175)
						(4.775, 0.0311499)
						(9.074, 0.0515499)
						(17.422, 0.08425)
						(33.8, 0.1329)
						(65.684, 0.1957)
						(129.808, 0.28945)
						(258.004, 0.4074)
						(513.596, 0.5384)
						(1025.69, 0.6736)
						(2049.53, 0.792901)
						(4097.79, 0.886952)
						(8193.17, 0.947701)
						(16385.2, 0.981451)
						(32769.3, 0.99505)
						(65537.9, 0.99965)
						(131077.0, 1.0)
						(136736.0, 1.0)
					};
					\addlegendentry{Sign-ALSH}
			
			\end{axis}
			\end{tikzpicture}
	\end{subfigure}%
	\hspace{\figuregap}	
	\begin{subfigure}[b]{0.36\textwidth}
			\begin{tikzpicture}
			\begin{axis}[
			height=\figurewidth\linewidth,
			width=\linewidth,
			legend style={font=\fontsize{1}{1}\selectfont},
			grid=major,
			legend pos=south east,
			change x base,
			x SI prefix=mega,x unit=\-,
			xlabel=Probed Items ,
			xmin=0,xmax=1000000,
			ymin=0,ymax=1,
			]

					\addplot  [mark=square*, red]
					coordinates 
					{
						(1.004, 0.0143)
						(2.001, 0.02655)
						(4.014, 0.0463998)
						(8.008, 0.0784999)
						(16.018, 0.1259)
						(32.022, 0.18965)
						(64.011, 0.27435)
						(128.049, 0.37045)
						(256.342, 0.4747)
						(512.347, 0.5756)
						(1024.49, 0.668)
						(2050.06, 0.750751)
						(4098.2, 0.813401)
						(8198.39, 0.860951)
						(16393.6, 0.896801)
						(32783.3, 0.924401)
						(65576.7, 0.94795)
						(131128.0, 0.966951)
						(262207.0, 0.982751)
						(524358.0, 0.993351)
						(1048630.0, 0.9984)
					};
					\addlegendentry{Cross-LSH}
					
					\addplot  [mark=triangle*, blue]
					coordinates 
					{
						(1.036, 0.0031)
						(2.059, 0.00555)
						(4.119, 0.0109)
						(8.239, 0.0196)
						(16.331, 0.0338499)
						(33.027, 0.0555499)
						(65.55, 0.08705)
						(131.357, 0.1295)
						(262.399, 0.1889)
						(523.418, 0.2561)
						(1043.76, 0.3339)
						(2095.2, 0.418)
						(4151.56, 0.50215)
						(8283.43, 0.583301)
						(16549.7, 0.65225)
						(32987.8, 0.718701)
						(65948.4, 0.782851)
						(131820.0, 0.84195)
						(263167.0, 0.891451)
						(525877.0, 0.940201)
						(1051490.0, 0.974451)
						(2098280.0, 0.99725)
						(2340370.0, 1.0)
					};
					\addlegendentry{Simple-LSH}
					
					\addplot 
					coordinates 
					{
						(1.025, 0.00145)
						(2.034, 0.00245)
						(4.068, 0.00465)
						(8.142, 0.0077)
						(16.267, 0.01415)
						(32.725, 0.0244)
						(65.326, 0.0390999)
						(129.954, 0.0620499)
						(261.795, 0.0949)
						(519.582, 0.1385)
						(1034.14, 0.19355)
						(2069.42, 0.2616)
						(4131.94, 0.33705)
						(8296.37, 0.4155)
						(16564.8, 0.5007)
						(33040.0, 0.57895)
						(66230.9, 0.657)
						(132537.0, 0.73545)
						(264618.0, 0.804901)
						(528288.0, 0.872501)
						(1053480.0, 0.936451)
						(2098040.0, 0.99425)
						(2340370.0, 1.0)
					};
					\addlegendentry{L2-ALSH}
					
					\addplot 
					coordinates 
					{
						(1.038, 0.00395)
						(2.076, 0.00645)
						(4.166, 0.0116)
						(8.237, 0.0185)
						(16.497, 0.03005)
						(32.648, 0.0461499)
						(65.663, 0.0743999)
						(133.752, 0.1142)
						(266.49, 0.1678)
						(535.231, 0.23005)
						(1053.4, 0.30225)
						(2090.16, 0.38225)
						(4165.23, 0.46545)
						(8312.95, 0.55095)
						(16770.4, 0.633901)
						(33660.0, 0.714051)
						(67845.2, 0.792551)
						(134819.0, 0.855051)
						(268005.0, 0.906201)
						(531336.0, 0.944401)
						(1053230.0, 0.977251)
						(2099520.0, 0.99885)
						(2340370.0, 1.0)
					};
					\addlegendentry{Sign-ALSH}
			
			\end{axis}
			\end{tikzpicture}	
	\end{subfigure}%
	\caption{Probed item-recall of top-20 items comparison between Cross-LSH and existing algorithms under a code length of 64 (best viewed in color). From left to right, the datasets are Netflix, Yahoo!Music and ImageNet, respectively.} 
	\label{top20-bit64:Cross comparision}
\end{figure*}
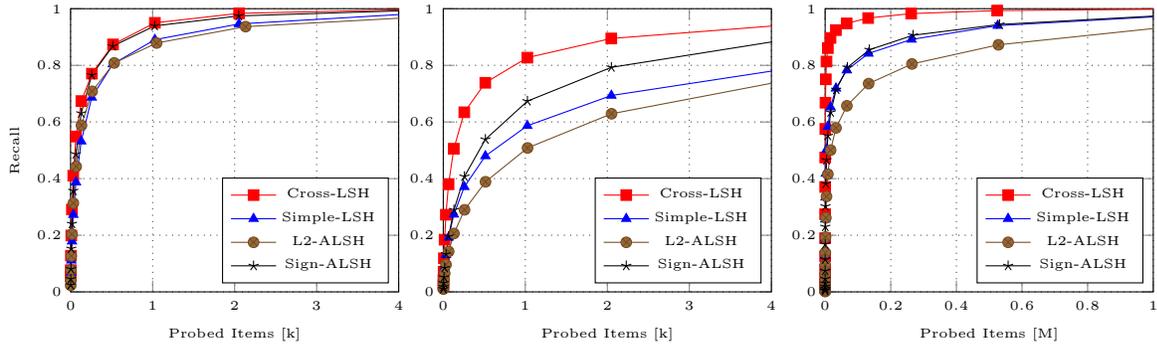  

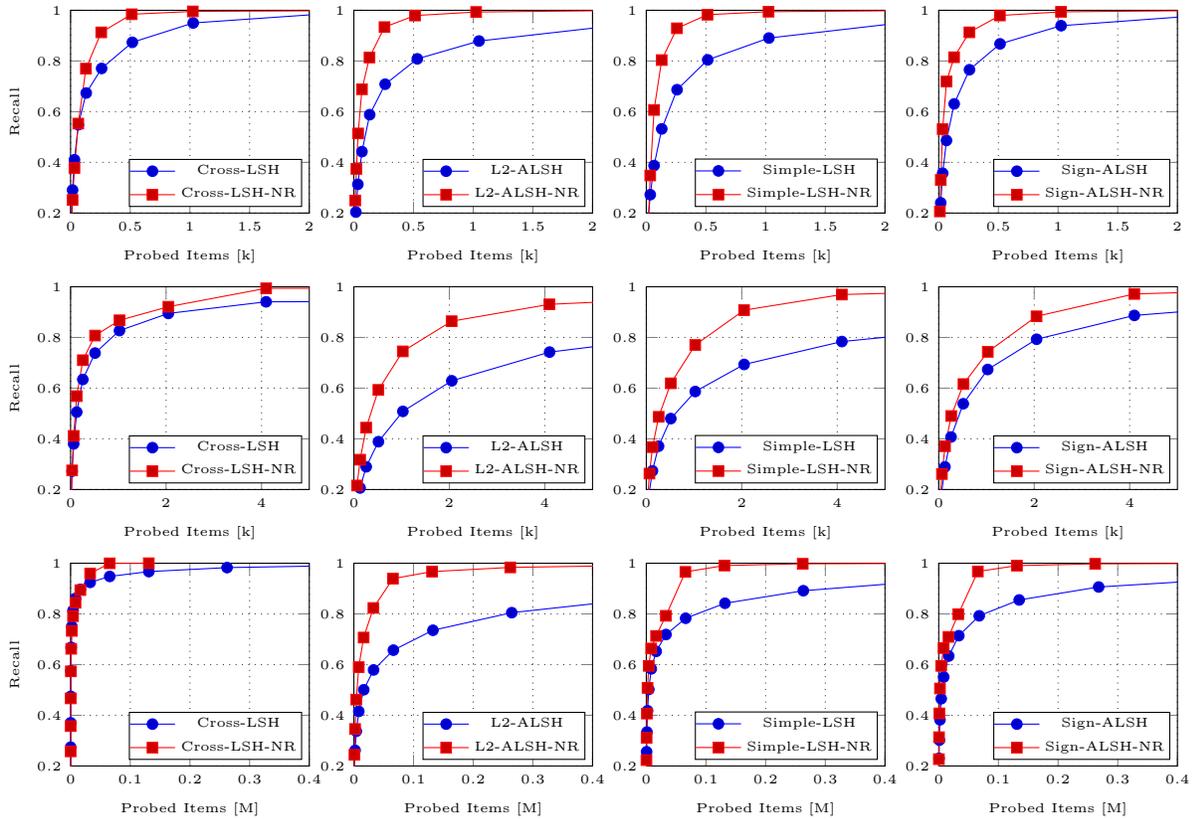
\begin{figure*}[t]
		\begin{subfigure}[b]{\figurepercentperrow\textwidth}
			\begin{tikzpicture}
			\begin{axis}[
			height=\figurewidth\linewidth,
			width=\linewidth,
			grid=major,
			legend pos=south east,
			legend style={inner xsep=0pt, inner ysep=0pt, font=\fontsize{1}{1}\selectfont},
			change x base,
			x SI prefix=kilo,x unit=\-,
			xlabel=Probed Items , ylabel=Recall,
			xmin=0,xmax=2000,
			ymin=0.2,ymax=1,
			mark size=2.0pt,
			]

					\addplot 
					coordinates 
					{
						(1.491, 0.0546997)
						(2.295, 0.0765496)
						(4.401, 0.12775)
						(8.493, 0.19905)
						(16.584, 0.2909)
						(32.474, 0.4102)
						(65.11, 0.5486)
						(131.029, 0.673751)
						(260.075, 0.770651)
						(517.461, 0.873702)
						(1027.45, 0.950102)
						(2054.4, 0.983601)
						(4100.19, 0.99655)
						(8193.16, 1.0)
						(16384.3, 1.0)
						(17770.0, 1.0)
					};
					\addlegendentry{Cross-LSH}
					\addplot 
					coordinates 
					{
						(1.247, 0.0304)
						(2.233, 0.0496998)
						(4.372, 0.0862)
						(8.407, 0.151)
						(16.44, 0.2518)
						(32.418, 0.3778)
						(64.393, 0.553151)
						(128.386, 0.7698)
						(256.773, 0.912952)
						(512.358, 0.984451)
						(1024.1, 0.99565)
						(2048.08, 0.99795)
						(4096.03, 0.99885)
						(8192.03, 0.9999)
						(16384.0, 1.0)
						(17770.0, 1.0)
					};
					\addlegendentry{Cross-LSH-NR}
			\end{axis}
			\end{tikzpicture}
	\end{subfigure}%
	\hspace{\figuregap}
	\begin{subfigure}[b]{\figurepercentperrow\textwidth}
	\begin{tikzpicture}
	\begin{axis}[
	height=\figurewidth\linewidth,
	width=\linewidth,
	legend pos=south east,
	legend style={inner xsep=0pt, inner ysep=0pt, font=\fontsize{1}{1}\selectfont},
	grid=major,
	legend pos=south east,
	change x base,
	x SI prefix=kilo,x unit=\-,
	xlabel=Probed Items ,
	xmin=0,xmax=2000,
	ymin=0.2,ymax=1,
	]
				
			\addplot 
			coordinates 
			{
				(1.087, 0.0233999)
				(2.145, 0.0423498)
				(4.311, 0.0781)
				(8.578, 0.12685)
				(16.94, 0.20385)
				(33.562, 0.3134)
				(67.885, 0.44245)
				(132.704, 0.5882)
				(262.463, 0.7083)
				(531.802, 0.808451)
				(1048.8, 0.878802)
				(2133.6, 0.936651)
				(4216.32, 0.969701)
				(8263.74, 0.9957)
				(16399.2, 0.99995)
				(17770.0, 1.0)
			};
			\addlegendentry{L2-ALSH}
			\addplot 
			coordinates 
			{
				(3.776, 0.0880002)
				(6.762, 0.1674)
				(8.553, 0.1988)
				(11.673, 0.2495)
				(21.681, 0.3744)
				(35.837, 0.5142)
				(67.949, 0.6885)
				(131.021, 0.813851)
				(256.771, 0.933802)
				(512.52, 0.979301)
				(1024.43, 0.99335)
				(2048.26, 0.9982)
				(4096.16, 0.99935)
				(8192.13, 0.99995)
				(16384.0, 1.0)
				(17770.0, 1.0)
			};
			\addlegendentry{L2-ALSH-NR}

	\end{axis}
	\end{tikzpicture}
	\end{subfigure}%
	\hspace{\figuregap}
	\begin{subfigure}[b]{\figurepercentperrow\textwidth}
	\begin{tikzpicture}
	\begin{axis}[
	height=\figurewidth\linewidth,
	width=\linewidth,
	legend pos=south east,
	legend style={inner xsep=0pt, inner ysep=0pt, font=\fontsize{1}{1}\selectfont},
	grid=major,
	change x base,
	x SI prefix=kilo,x unit=\-,
	xlabel=Probed Items,
	xmin=0,xmax=2000,
	ymin=0.2,ymax=1,
	]

			\addplot 
			coordinates 
			{
				(1.437, 0.02565)
				(2.62, 0.0423499)
				(4.873, 0.0676499)
				(8.817, 0.11135)
				(17.166, 0.17865)
				(33.092, 0.2726)
				(65.215, 0.3874)
				(129.49, 0.5321)
				(257.717, 0.686751)
				(514.692, 0.804551)
				(1027.27, 0.890602)
				(2052.24, 0.946101)
				(4099.77, 0.980901)
				(8196.18, 0.99685)
				(16386.4, 1.0)
				(17770.0, 1.0)
			};
			\addlegendentry{Simple-LSH}
			\addplot 
			coordinates 
			{
				(1.585, 0.01875)
				(2.898, 0.03015)
				(5.299, 0.0491999)
				(8.716, 0.0897)
				(16.905, 0.15705)
				(32.927, 0.34845)
				(65.001, 0.6068)
				(128.771, 0.80365)
				(256.556, 0.929101)
				(512.448, 0.982651)
				(1024.28, 0.9945)
				(2048.14, 0.9981)
				(4096.05, 0.9994)
				(8192.04, 0.9998)
				(16384.0, 1.0)
				(17770.0, 1.0)
			};
			\addlegendentry{Simple-LSH-NR}
	\end{axis}
	\end{tikzpicture}
	\end{subfigure}%	
	\hspace{\figuregap}	
	\begin{subfigure}[b]{\figurepercentperrow\textwidth}
		\begin{tikzpicture}
		\begin{axis}[
		height=\figurewidth\linewidth,
		width=\linewidth,
		legend pos=south east,
		legend style={inner xsep=0pt, inner ysep=0pt, font=\fontsize{1}{1}\selectfont},
		grid=major,
		legend pos=south east,
		change x base,
		x SI prefix=kilo,x unit=\-,
		xlabel=Probed Items ,
		xmin=0,xmax=2000,
		ymin=0.2,ymax=1,
		]
				\addplot 
				coordinates 
				{
					(1.167, 0.0236999)
					(2.502, 0.0451498)
					(4.515, 0.0803999)
					(8.663, 0.15265)
					(16.594, 0.2404)
					(32.888, 0.35705)
					(64.897, 0.4864)
					(129.3, 0.6308)
					(257.464, 0.7655)
					(513.701, 0.867552)
					(1025.08, 0.939251)
					(2049.39, 0.974501)
					(4097.67, 0.99325)
					(8193.56, 0.99895)
					(16387.2, 0.99995)
					(17770.0, 1.0)
				};
				\addlegendentry{Sign-ALSH}
				\addplot 
				coordinates 
				{
					(2.796, 0.0675498)
					(3.54, 0.0830999)
					(5.245, 0.12455)
					(9.537, 0.2064)
					(17.038, 0.33075)
					(32.784, 0.5312)
					(65.191, 0.71915)
					(128.692, 0.81485)
					(256.951, 0.913151)
					(512.519, 0.979501)
					(1024.36, 0.994)
					(2048.17, 0.99825)
					(4096.11, 0.9993)
					(8192.06, 0.99985)
					(16384.0, 1.0)
					(17770.0, 1.0)
				};
				\addlegendentry{Sign-ALSH-NR}
		\end{axis}
		\end{tikzpicture}
	\end{subfigure}%
	
		\begin{subfigure}[b]{\figurepercentperrow\textwidth}
			\begin{tikzpicture}
			\begin{axis}[
			height=\figurewidth\linewidth,
			width=\linewidth,
			grid=major,
			legend pos=south east,
			legend style={inner xsep=0pt, inner ysep=0pt, font=\fontsize{1}{1}\selectfont},
			change x base,
			x SI prefix=kilo,x unit=\-,
			xmin=0,xmax=5000,
			ymin=0.2,ymax=1,
			xlabel=Probed Items , ylabel=Recall,
			]
					\addplot 
					coordinates 
					{
						(1.032, 0.0238999)
						(2.024, 0.0420498)
						(4.047, 0.0711499)
						(8.084, 0.1186)
						(16.092, 0.1843)
						(32.152, 0.2724)
						(64.171, 0.38)
						(128.23, 0.50545)
						(256.276, 0.63425)
						(512.3, 0.738401)
						(1024.34, 0.827151)
						(2048.28, 0.894801)
						(4096.36, 0.940651)
						(136736.0, 1.0)
					};
					\addlegendentry{Cross-LSH}
					\addplot 
					coordinates 
					{
						(1.018, 0.01685)
						(2.037, 0.0316999)
						(4.066, 0.0595498)
						(8.105, 0.1027)
						(16.087, 0.17375)
						(32.122, 0.2754)
						(64.25, 0.41175)
						(128.268, 0.5685)
						(256.763, 0.710351)
						(512.642, 0.807401)
						(1025.31, 0.867751)
						(2048.86, 0.921251)
						(4096.06, 0.9939)
						(136736.0, 1.0)
					};
					\addlegendentry{Cross-LSH-NR}
			\end{axis}
			\end{tikzpicture}
	\end{subfigure}%
	\hspace{\figuregap}
	\begin{subfigure}[b]{\figurepercentperrow\textwidth}
	\begin{tikzpicture}
	\begin{axis}[
	height=\figurewidth\linewidth,
	width=\linewidth,
	legend pos=south east,
	legend style={inner xsep=0pt, inner ysep=0pt, font=\fontsize{1}{1}\selectfont},
	grid=major,
	change x base,
	x SI prefix=kilo,x unit=\-,
	xlabel=Probed Items,
	xmin=0,xmax=5000,
	ymin=0.2,ymax=1,
	]
	
				\addplot 
				coordinates 
				{
					(1.265, 0.00980002)
					(2.398, 0.01565)
					(4.699, 0.02575)
					(8.962, 0.0437999)
					(17.238, 0.0653499)
					(33.281, 0.0961)
					(65.468, 0.1427)
					(130.28, 0.20615)
					(258.902, 0.28975)
					(515.12, 0.3888)
					(1027.18, 0.5083)
					(2051.45, 0.629)
					(4100.54, 0.7422)
					(8198.43, 0.836951)
					(16392.0, 0.910301)
					(136736.0, 1.0)
				};
				\addlegendentry{L2-ALSH}
				\addplot 
				coordinates 
				{
					(1.243, 0.00945002)
					(2.463, 0.01725)
					(4.721, 0.02955)
					(9.227, 0.0514999)
					(18.086, 0.08535)
					(34.71, 0.1366)
					(66.96, 0.21595)
					(131.491, 0.31775)
					(258.673, 0.44415)
					(513.954, 0.59315)
					(1025.27, 0.745101)
					(2048.48, 0.864102)
					(4096.19, 0.930652)
					(8192.33, 0.965951)
					(16384.3, 0.987201)
					(32768.1, 0.9964)
					(136736.0, 1.0)
				};
				\addlegendentry{L2-ALSH-NR}
	
	\end{axis}
	\end{tikzpicture}
\end{subfigure}%
	\hspace{\figuregap}
	\begin{subfigure}[b]{\figurepercentperrow\textwidth}
	\begin{tikzpicture}
	\begin{axis}[
	height=\figurewidth\linewidth,
	width=\linewidth,
	legend pos=south east,
	legend style={inner xsep=0pt, inner ysep=0pt, font=\fontsize{1}{1}\selectfont},
	grid=major,
	change x base,
	x SI prefix=kilo,x unit=\-,
	xlabel=Probed Items , 
	xmin=0,xmax=5000,
	ymin=0.2,ymax=1,
	]
	
				\addplot 
				coordinates 
				{
					(1.08, 0.0117)
					(2.229, 0.01975)
					(4.326, 0.0354499)
					(8.441, 0.0564499)
					(16.455, 0.0869501)
					(32.696, 0.13125)
					(64.761, 0.1923)
					(128.987, 0.27365)
					(256.988, 0.3709)
					(513.2, 0.479849)
					(1025.03, 0.5867)
					(2049.32, 0.69345)
					(4097.04, 0.783801)
					(8193.31, 0.859801)
					(16386.0, 0.922851)
					(32770.4, 0.964201)
					(65538.6, 0.98645)
					(131074.0, 1.0)
					(136736.0, 1.0)
				};
				\addlegendentry{Simple-LSH}
				\addplot 
				coordinates 
				{
					(1.062, 0.0136)
					(2.171, 0.02495)
					(4.237, 0.0432999)
					(8.555, 0.0722999)
					(16.968, 0.11665)
					(33.648, 0.17795)
					(66.077, 0.26295)
					(130.513, 0.3673)
					(258.714, 0.48685)
					(513.881, 0.61905)
					(1025.03, 0.770301)
					(2048.47, 0.907701)
					(4096.19, 0.969251)
					(8192.09, 0.9899)
					(16384.1, 0.9972)
					(32768.1, 0.9998)
					(65536.1, 1.0)
					(131072.0, 1.0)
					(136736.0, 1.0)
				};
				\addlegendentry{Simple-LSH-NR}	
	\end{axis}
	\end{tikzpicture}
\end{subfigure}%
	\hspace{\figuregap}
	\begin{subfigure}[b]{\figurepercentperrow\textwidth}
		\begin{tikzpicture}
		\begin{axis}[
		height=\figurewidth\linewidth,
		width=\linewidth,
		legend pos=south east,
		legend style={inner xsep=0pt, inner ysep=0pt, font=\fontsize{1}{1}\selectfont},
		grid=major,
		change x base,
		x SI prefix=kilo,x unit=\-,
		xlabel=Probed Items ,
		xmin=0,xmax=5000,
		ymin=0.2,ymax=1,
		]
		
				\addplot 
				coordinates 
				{
					(1.435, 0.00995002)
					(2.562, 0.0175)
					(4.775, 0.0311499)
					(9.074, 0.0515499)
					(17.422, 0.08425)
					(33.8, 0.1329)
					(65.684, 0.1957)
					(129.808, 0.28945)
					(258.004, 0.4074)
					(513.596, 0.5384)
					(1025.69, 0.6736)
					(2049.53, 0.792901)
					(4097.79, 0.886952)
					(8193.17, 0.947701)
					(16385.2, 0.981451)
					(32769.3, 0.99505)
					(65537.9, 0.99965)
					(131077.0, 1.0)
					(136736.0, 1.0)
				};
				\addlegendentry{Sign-ALSH}
				\addplot 
				coordinates 
				{
					(1.668, 0.014)
					(2.993, 0.02415)
					(5.013, 0.0416499)
					(9.191, 0.0682499)
					(17.864, 0.1091)
					(34.448, 0.17095)
					(67.193, 0.26115)
					(132.272, 0.3709)
					(261.687, 0.49015)
					(518.516, 0.61615)
					(1027.7, 0.743201)
					(2048.83, 0.883351)
					(4096.37, 0.971601)
					(8192.27, 0.99375)
					(16384.1, 0.9991)
					(32768.1, 0.9999)
					(65536.0, 1.0)
					(131072.0, 1.0)
					(136736.0, 1.0)
				};
				\addlegendentry{Sign-ALSH-NR}
		
		\end{axis}
		\end{tikzpicture}
	\end{subfigure}%
	
		\begin{subfigure}[b]{\figurepercentperrow\textwidth}
			\begin{tikzpicture}
			\begin{axis}[
				height=\figurewidth\linewidth,
				width=\linewidth,
				grid=major,
				legend pos=south east,
				legend style={inner xsep=0pt, inner ysep=0pt, font=\fontsize{1}{1}\selectfont},
				change x base,
				x SI prefix=mega,x unit=\-,
				xmin=0,xmax=400000,
				ymin=0.2,ymax=1,
				xlabel=Probed Items , ylabel=Recall,
			]

				\addplot 
				coordinates 
				{
					(1.004, 0.0143)
					(2.001, 0.02655)
					(4.014, 0.0463998)
					(8.008, 0.0784999)
					(16.018, 0.1259)
					(32.022, 0.18965)
					(64.011, 0.27435)
					(128.049, 0.37045)
					(256.342, 0.4747)
					(512.347, 0.5756)
					(1024.49, 0.668)
					(2050.06, 0.750751)
					(4098.2, 0.813401)
					(8198.39, 0.860951)
					(16393.6, 0.896801)
					(32783.3, 0.924401)
					(65576.7, 0.94795)
					(131128.0, 0.966951)
					(262207.0, 0.982751)
					(524358.0, 0.993351)
					(1048630.0, 0.9984)
				};
				\addlegendentry{Cross-LSH}
				\addplot 
				coordinates 
				{
					(1.00125, 0.00375)
					(2.0, 0.0149375)
					(4.00125, 0.0335)
					(8.0075, 0.0649374)
					(16.0075, 0.110125)
					(32.0075, 0.172375)
					(64.0, 0.257625)
					(128.001, 0.357313)
					(256.003, 0.466687)
					(512.005, 0.573375)
					(1024.0, 0.662313)
					(2048.0, 0.732938)
					(4096.01, 0.791001)
					(8192.0, 0.844063)
					(16384.0, 0.893938)
					(32768.0, 0.959625)
					(65536.0, 0.999625)
					(131072.0, 1.0)
				};
				\addlegendentry{Cross-LSH-NR}
				
			\end{axis}
			\end{tikzpicture}	
	\end{subfigure}%
	\hspace{\figuregap}
	\begin{subfigure}[b]{\figurepercentperrow\textwidth}
		\begin{tikzpicture}
		\begin{axis}[
		height=\figurewidth\linewidth,
		width=\linewidth,
		grid=major,
		legend pos=south east,
		legend style={inner xsep=0pt, inner ysep=0pt, font=\fontsize{1}{1}\selectfont},
		change x base,
		x SI prefix=mega,x unit=\-,
		xlabel=Probed Items ,
		xmin=0,xmax=400000,
		ymin=0.2,ymax=1,
		]
			\addplot 
			coordinates 
			{
				(1.025, 0.00145)
				(2.034, 0.00245)
				(4.068, 0.00465)
				(8.142, 0.0077)
				(16.267, 0.01415)
				(32.725, 0.0244)
				(65.326, 0.0390999)
				(129.954, 0.0620499)
				(261.795, 0.0949)
				(519.582, 0.1385)
				(1034.14, 0.19355)
				(2069.42, 0.2616)
				(4131.94, 0.33705)
				(8296.37, 0.4155)
				(16564.8, 0.5007)
				(33040.0, 0.57895)
				(66230.9, 0.657)
				(132537.0, 0.73545)
				(264618.0, 0.804901)
				(528288.0, 0.872501)
				(1053480.0, 0.936451)
				(2098040.0, 0.99425)
				(2340370.0, 1.0)
			};
			\addlegendentry{L2-ALSH}
			\addplot 
			coordinates 
			{
				(1.09, 0.00125)
				(2.085, 0.003)
				(4.155, 0.00625)
				(8.36, 0.00775)
				(16.495, 0.01275)
				(32.86, 0.02475)
				(64.785, 0.03725)
				(130.005, 0.063)
				(258.21, 0.107)
				(514.865, 0.16475)
				(1027.88, 0.24425)
				(2052.23, 0.346)
				(4098.38, 0.462)
				(8195.49, 0.59)
				(16386.5, 0.70675)
				(32769.0, 0.8235)
				(65536.2, 0.93925)
				(131072.0, 0.967)
				(262144.0, 0.9835)
				(524288.0, 0.993)
				(1048580.0, 0.9965)
				(2097150.0, 0.999)
				(2340370.0, 1.0)
			};
			\addlegendentry{L2-ALSH-NR}
		
		\end{axis}
		\end{tikzpicture}	
	\end{subfigure}%
	\hspace{\figuregap}	
	\begin{subfigure}[b]{\figurepercentperrow\textwidth}
	\begin{tikzpicture}
	\begin{axis}[
		height=\figurewidth\linewidth,
		width=\linewidth,
		grid=major,
		legend pos=south east,
		legend style={inner xsep=0pt, inner ysep=0pt, font=\fontsize{1}{1}\selectfont},
		change x base,
		x SI prefix=mega,x unit=\-,
		xlabel=Probed Items,
		xmin=0,xmax=400000,
		ymin=0.2,ymax=1,
	]
			\addplot 
			coordinates 
			{
				(1.036, 0.0031)
				(2.059, 0.00555)
				(4.119, 0.0109)
				(8.239, 0.0196)
				(16.331, 0.0338499)
				(33.027, 0.0555499)
				(65.55, 0.08705)
				(131.357, 0.1295)
				(262.399, 0.1889)
				(523.418, 0.2561)
				(1043.76, 0.3339)
				(2095.2, 0.418)
				(4151.56, 0.50215)
				(8283.43, 0.583301)
				(16549.7, 0.65225)
				(32987.8, 0.718701)
				(65948.4, 0.782851)
				(131820.0, 0.84195)
				(263167.0, 0.891451)
				(525877.0, 0.940201)
				(1051490.0, 0.974451)
				(2098280.0, 0.99725)
				(2340370.0, 1.0)
			};
			\addlegendentry{Simple-LSH}
			\addplot 
			coordinates 
			{
				(1.005, 0.0035)
				(2.017, 0.0066)
				(4.021, 0.0112)
				(8.036, 0.01855)
				(16.062, 0.0341499)
				(32.095, 0.0577999)
				(64.185, 0.0951)
				(128.199, 0.14675)
				(256.336, 0.2231)
				(512.413, 0.3112)
				(1024.39, 0.406699)
				(2048.44, 0.50825)
				(4096.52, 0.59495)
				(8192.34, 0.6635)
				(16384.6, 0.7131)
				(32768.5, 0.7925)
				(65536.0, 0.9658)
				(131072.0, 0.9909)
				(262144.0, 0.9978)
				(524288.0, 0.9999)
				(1048580.0, 1.0)
				(2097150.0, 1.0)
			};
			\addlegendentry{Simple-LSH-NR}
	\end{axis}
	\end{tikzpicture}	
	\end{subfigure}%
	\hspace{\figuregap}	
	\begin{subfigure}[b]{\figurepercentperrow\textwidth}
			\begin{tikzpicture}
			\begin{axis}[
			height=\figurewidth\linewidth,
			width=\linewidth,
			legend pos=south east,
			legend style={inner xsep=0pt, inner ysep=0pt, font=\fontsize{1}{1}\selectfont},
			grid=major,
			change x base,
			x SI prefix=mega,x unit=\-,
			xlabel=Probed Items ,
			xmin=0,xmax=400000,
			ymin=0.2,ymax=1,
			]
				\addplot 
				coordinates 
				{
					(1.038, 0.00395)
					(2.076, 0.00645)
					(4.166, 0.0116)
					(8.237, 0.0185)
					(16.497, 0.03005)
					(32.648, 0.0461499)
					(65.663, 0.0743999)
					(133.752, 0.1142)
					(266.49, 0.1678)
					(535.231, 0.23005)
					(1053.4, 0.30225)
					(2090.16, 0.38225)
					(4165.23, 0.46545)
					(8312.95, 0.55095)
					(16770.4, 0.633901)
					(33660.0, 0.714051)
					(67845.2, 0.792551)
					(134819.0, 0.855051)
					(268005.0, 0.906201)
					(531336.0, 0.944401)
					(1053230.0, 0.977251)
					(2099520.0, 0.99885)
					(2340370.0, 1.0)
				};
				\addlegendentry{Sign-ALSH}
				\addplot 
				coordinates 
				{
					(1.086, 0.00415)
					(2.164, 0.00690001)
					(4.154, 0.0116)
					(8.491, 0.02055)
					(16.72, 0.035)
					(33.289, 0.0611999)
					(64.808, 0.09675)
					(129.73, 0.14935)
					(256.983, 0.22695)
					(512.797, 0.31365)
					(1024.62, 0.4075)
					(2048.45, 0.50575)
					(4096.44, 0.595)
					(8192.29, 0.6653)
					(16384.2, 0.7093)
					(32768.3, 0.798851)
					(65536.1, 0.96745)
					(131072.0, 0.9905)
					(262144.0, 0.9977)
					(524288.0, 0.9998)
					(1048580.0, 1.0)
					(2097150.0, 1.0)
				};
				\addlegendentry{Sign-ALSH-NR}
			\end{axis}
			\end{tikzpicture}	
		\end{subfigure}%
	
	\caption{Probed item-recall of top-20 items comparison between the original meta algorithms and their norm-range versions under a code length of 64. From top row to bottom row, the datasets are Netflix, Yahoo!Music and ImageNet, respectively.} 
	\label{top20-bit64:Norm-Range comparision}
\end{figure*} 

%%%%%%=================================%%%%%%

\section{Experiment results under more code lengths (only top 20)}

\end{document}